\documentclass[letterpaper]{article} 
\usepackage{aaai2027} 
\affiliations{}
\usepackage[hyphens]{url}  
\usepackage{graphicx} 
\usepackage{natbib}  
\usepackage{caption} 
\usepackage{optidef} 
\usepackage{amsmath,amsfonts,bm}

\def\eqref#1{equation~\ref{#1}}

\def\1{\bm{1}}

\DeclareMathAlphabet{\mathsfit}{\encodingdefault}{\sfdefault}{m}{sl}
\SetMathAlphabet{\mathsfit}{bold}{\encodingdefault}{\sfdefault}{bx}{n}

\usepackage{enumitem}
\usepackage{bm}
\usepackage{algorithm}
\usepackage{algorithmic}
\usepackage[T1]{fontenc}
\usepackage{xcolor}
\usepackage{array}
\usepackage{booktabs}
\usepackage{multirow}
\usepackage[labelformat=simple]{subcaption}

\usepackage{amsmath}
\usepackage{nicefrac}       
\usepackage{microtype}
\usepackage{amsthm}
\usepackage{amssymb}
\usepackage{amsfonts}
\usepackage{mathtools}
\usepackage{soul}
\usepackage{comment} 

\newcommand{\cB}{{\mathcal B}}

\newcommand{\cE}{{\mathcal E}}

\newcommand{\cP}{{\mathcal P}} 
 
\newcommand{\cR}{{\mathcal R}}

\newcommand{\cM}{{\mathcal M}}

\newcommand{\cY}{{\mathcal Y}}

\newcommand{\bma}{{\bm a}}
\newcommand{\bmb}{{\bm b}}

\newcommand{\bmg}{{\bm g}}

\newcommand{\bmy}{{\bm y}}

\newcommand{\bmQ}{{\bm Q}}

\newcommand{\bmZ}{{\bm Z}}

\newtheorem{theorem}{Theorem}
\newtheorem{definition}{Definition}
\newtheorem{lemma}{Lemma}
\newtheorem{proposition}{Proposition}
\newcommand{\CUT}[1]{}
\newcommand\blfootnote[1]{%
\begingroup
\renewcommand\thefootnote{}\footnote{#1}%
\addtocounter{footnote}{-1}%
\endgroup
}

\date{}
\begin{document}
\title{Learning Multi-Timescale Interventions under Safety and Resource Constraints}
\author{%
  David H. Mguni\textsuperscript{$\dagger$, \rm 1,$\ast$}, Wanrong Yang\textsuperscript{$\ast$, \rm 3} ,
  Jing Dong\textsuperscript{\rm 2},   Jing Peng\textsuperscript{\rm 4},
  Ziquan Liu\textsuperscript{\rm 1},
  Muhammad Salman Haleem\textsuperscript{\rm 1}, 
  Baoxiang Wang\textsuperscript{\rm 2}, Dominik Wojtczak\textsuperscript{\rm 3}}
  \affiliations {
    \textsuperscript{\rm 1}Queen Mary University London,
    \textsuperscript{\rm 2}The Chinese University of Hong Kong,
    \textsuperscript{\rm 3}University of Liverpool
    ,
    \textsuperscript{\rm 4}Amazon
}
\maketitle

\begin{abstract}
\blfootnote{$^\ast$equal contribution\\ $^\dagger$Corresponding author  \textlangle d.mguni@qmul.ac.uk\textrangle. }
Many sequential decision problems offer qualitatively different ways of influencing the environment: some interventions act immediately, whereas others induce persistent effects that continue to shape future states long after the decision that initiated them. An agent must then decide jointly \emph{when} to intervene, \emph{which temporal mode} to use and \emph{how strongly}, while accounting for residual effects and limited intervention resources. We introduce \textsc{MINT}: Multi-timescale Intervention Network Training. Persistent effects are carried by an augmented intervention state that accumulates and decays, while a structured policy separates intervention-mode selection from conditional control. Unlike temporal abstractions that extend policy execution, persistent-effect interventions remain part of the environment dynamics and may overlap with later interventions. We show that the augmented state is a sufficient statistic for the intervention history, preserving the Markov property, and establish Bellman contraction and almost-sure convergence of a tabular Q-learning instance. Across persistent-control MuJoCo locomotion, stochastic inventory management, and a physiologically grounded Type 1 Diabetes Mellitus (T1DM) simulator, \textsc{MINT} attains the best mean primary metric on two of three benchmarks while using fewer intervention activations than an identically augmented flat policy. Its return advantage is present at every binding intervention budget and narrows to parity in the unconstrained reference setting. In T1DM it achieves $90.9\pm0.9\%$ time in range with zero time below range, improving on the strongest baseline by $21.5\%$ points. Code, benchmarks and the configuration for every reported run are available at \url{https://github.com/heywanrong/mint-rl}.
\end{abstract}

\section{Introduction}
Many sequential decision problems offer multiple mechanisms for influencing a system, whose effects unfold over very different time scales. Some actions produce immediate and short-lived changes, whereas others create persistent effects that remain active long after the decision that initiated them. A controller operating in such settings must therefore reason not only about \emph{what} action to take, but also about \emph{when} intervention is worthwhile, which temporal mode should be used, and how previous interventions continue to shape the current state. These decisions become particularly important when interventions are costly, limited by a finite budget, or capable of producing unsafe outcomes~\cite{mguni2022timing,wachi2024survey,lin2023safe}.

This structure arises naturally across otherwise disparate domains. In Type 1 Diabetes Mellitus (T1DM), fast-acting bolus insulin corrects short-term glucose excursions while basal insulin produces a longer-lasting physiological effect~\cite{nwokolo2023artificial,boughton2024role}. In inventory management, emergency replenishment can address immediate shortages while standard replenishment remains in a supply pipeline and affects inventory only over subsequent periods~\cite{leluc2023marlim,khirwar2023cooperative,maggiar2025structure}. In physical control, an actuator may combine immediate control inputs with persistent actuation or residual dynamics. Although the semantics differ, these problems share a common decision structure: the effects of past interventions remain active while the agent is free to make new decisions.

Standard reinforcement learning (RL)~\cite{sutton2018reinforcement} does not
exploit this structure. A flat policy can represent both intervention types by
enlarging the state and action spaces, but must learn from data their different
temporal roles, how residual effects interact, and when to withhold
intervention. Temporal abstraction offers another route: the options
framework~\cite{klissarov2021flexible} and
option-critic~\cite{bacon2017option} extend policy execution over multiple time
steps and remain an active research area~\cite{klissarov2025discovering}. Our
setting concerns a different source of temporal structure: the decision that
initiates an intervention may terminate immediately while its \emph{physical
consequences} persist in the environment.

A simple example illustrates the distinction. In a building with an immediate
actuator and underfloor heating, the actuator alters the current temperature at
once, whereas heating the floor stores thermal energy that keeps affecting the
room after the intervention that produced it has ended, while the controller
remains free to act again say, in response to a change in external
temperature. Temporally extended \emph{behaviour} and temporally extended
\emph{consequences} are therefore distinct.

Options therefore abstract persistent \emph{behaviour},
whereas our formulation explicitly models persistent
\emph{consequences}. A related line of work adapts how long
an action is held or how often decisions are
taken~\cite{sabbioni2023simultaneously,hu2025state}, again
governing the duration of the \emph{decision} rather than the
lifetime of its effect. Although options can represent such
problems with suitable state augmentation, we hypothesise
that directly modelling residual effects and intervention
timing provides a more appropriate inductive bias when
effects overlap and resources are limited. We test this
against fixed-option and state-augmented option baselines.

To make this explicit, consider a persistent-effect intervention state $z_t$
under exponential persistence, $z_{t+1} = \rho z_t + B_P\eta_t^P$, so that
\begin{equation}
z_t = \rho^t z_0 +
\sum_{k=0}^{t-1}\rho^{t-1-k}B_P\eta_k^P.
\label{eq:superposition}
\end{equation}
Hence $z_t$ summarises the superposition of residual effects from previous
interventions, and while those effects remain active the controller may still
select new immediate or persistent-effect interventions. The problem is
therefore not simply one of temporal abstraction, but of \emph{sequential
decision making under persistent, overlapping interventions}.

\begin{figure*}[t]
    \centering
    \includegraphics[width=\textwidth]{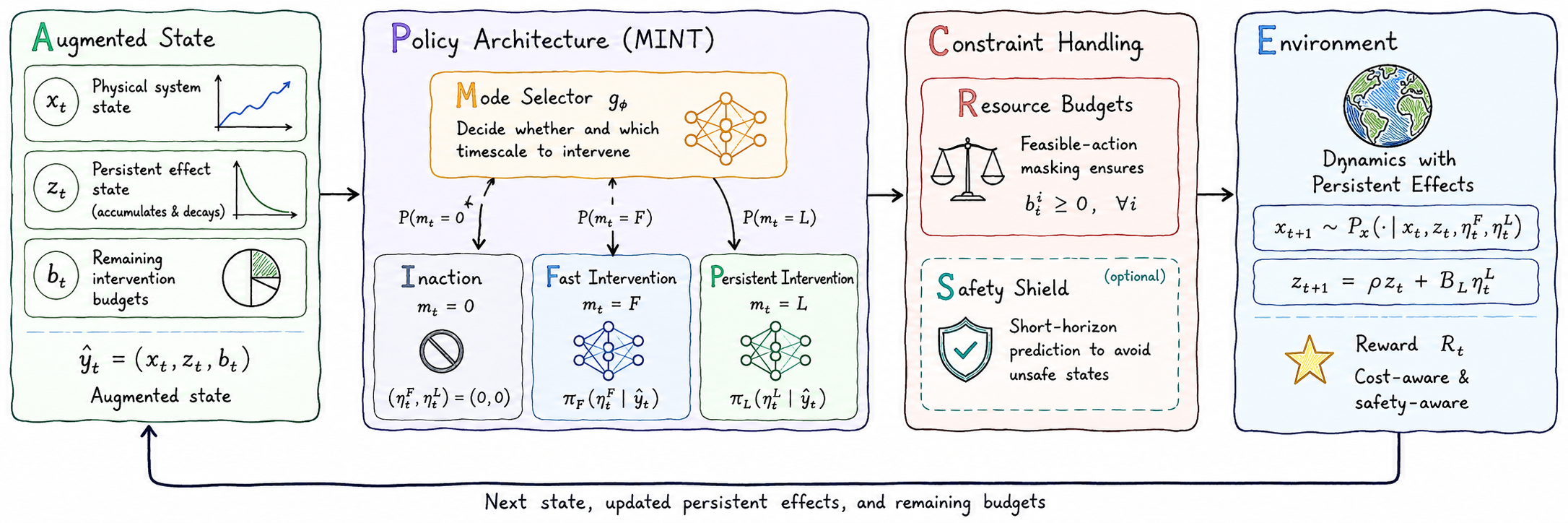}
    \caption{
Overview of \textsc{MINT}. The augmented state
$\hat y_t=(x_t,z_t,b_t)$ records the physical state, residual
persistent effects and remaining budgets. An intervention
selector chooses inaction, an immediate intervention or a
persistent-effect intervention, while conditional policies set
the control. Feasible-action masking enforces hard budgets and
an optional shield rejects actions predicted to be unsafe.
The figure names the immediate and persistent modes
\emph{fast} and \emph{long}, so $F\equiv I$, $L\equiv P$ and
$g_\phi\equiv g_\psi$ throughout.
}
    \label{fig:mint_framework}
\end{figure*}

We formalise this setting as a Multi-Timescale Intervention
Markov decision process (MTI-MDP) and introduce \textsc{MINT}
(Multi-timescale Intervention Network Training). Its augmented
state records residual effects and remaining budgets, while
its structured policy separates intervention selection from
conditional control. The selector learns whether and on
which temporal scale to intervene, and the conditional
policies determine how to act. Figure~\ref{fig:mint_framework}
summarises the framework.
Resource constraints are incorporated directly into the state following state-augmentation approaches to constrained RL~\cite{sootla2022saute,wachi2024survey}. Feasible-action masking guarantees that finite intervention budgets cannot be exceeded~\cite{varricchione2024pure}, while an optional short-horizon predictive shield~\cite{bastani2021safe,banerjee2024dynamic} is used for state-safety constraints. The resulting formulation retains the Markov structure required by standard RL despite the apparent history dependence induced by persistent-effect interventions.

We evaluate the framework on three complementary problem families. A persistent-action variant of a standard MuJoCo locomotion task~\cite{towers2024gymnasium} allows us to vary the persistence parameter $\rho$ directly and test whether structured intervention learning becomes more useful as temporal effects become longer lived. Stochastic inventory management provides a naturally occurring delayed-intervention problem in which emergency and pipeline replenishment operate over different horizons. Finally, T1DM glucose control~\cite{man2014uva,hettiarachchi2022glucoenv} provides a safety-critical application in which fast- and long-acting insulin have directly interpretable temporal roles, and in which hypoglycaemia makes over-intervention actively dangerous~\cite{cryer2010hypoglycemia}. Our experiments are organised around three questions: \emph{Does explicit multi-timescale structure become more valuable as intervention persistence increases? Does learning when to intervene become more important when intervention resources are scarce? Do these effects persist across qualitatively different domains?}

\noindent\textbf{Contributions.}
Our main contributions are:
\begin{itemize}[leftmargin=*,nosep]
    \item We formulate the MTI-MDP, a sequential decision problem with immediate and persistent-effect intervention channels, residual effects that can accumulate across decisions, and explicit intervention budgets.
\item We introduce \textsc{MINT}, a structured reinforcement learning framework that explicitly separates
\emph{whether}, \emph{when}, \emph{on which temporal scale},
and \emph{how} to intervene. Unlike temporal abstractions
that extend policy execution, \textsc{MINT} represents
persistent-effect intervention effects directly in the environment
state, allowing multiple residual interventions to coexist
while new intervention decisions continue to be made.
    \item We show that the persistent-effect and budget variables provide a sufficient Markov state for the history-dependent intervention process. We establish budget feasibility by construction and retain standard Bellman contraction and tabular Q-learning convergence guarantees.
    \item Across MuJoCo, inventory control and T1DM, \textsc{MINT} attains
the best mean primary metric on two of three benchmarks and
uses fewer reported channel activations than the flat
augmented policy in all three, most strikingly raising T1DM time in range from the flat augmented policy's $64.7\%$ to $90.9\%$ with zero time below range, the return gap being present at every binding intervention budget and narrowing to parity once interventions are unconstrained.
\end{itemize}

\section{Related Work}

Temporally extended action is usually modelled with options, which hold an
intra-option policy until a termination condition fires~\cite{sutton1999between,klissarov2021flexible,klissarov2025discovering,zadem2024reconciling,li2024when},
or by keeping the action space flat and learning how long to repeat an action or
how often to decide~\cite{sabbioni2023simultaneously,lee2024learning,hu2025state,karimi2023dynamic}.
All of these control how long a \emph{decision} lasts, whereas MTI-MDPs
represent persistence in its \emph{consequences}: residual effects become part
of the environment state, can superpose, and remain active while new decisions
are taken --- hence our fixed-option and state-augmented option baselines.
Constraints are handled in the constrained MDP literature~\cite{wachi2024survey}
by Lagrangian and trust-region methods~\cite{achiam2017constrained},
shielding~\cite{bastani2021safe,konighofer2025shields,corsi2024verification,banerjee2024dynamic}, action
masking~\cite{varricchione2024pure} and state augmentation~\cite{sootla2022saute},
with standardised benchmarks~\cite{ji2023safety} and extensions to offline budgets~\cite{lin2023safe} and feasibility-aware representations~\cite{cen2024feasibility}; we build on state augmentation
but address an orthogonal structure, namely that actions have heterogeneous
temporal footprints. Deciding \emph{whether} an action justifies its cost is the
learning counterpart of optimal stopping~\cite{mguni2022timing,pmlr-v202-mguni23a,tsitsiklis1999optimal},
and because an intervention's reward accrues long after it is issued the problem
is also one of credit assignment~\cite{pignatelli2024survey}, which
delayed-feedback methods address by reconstructing the signal from
history~\cite{wu2024boosting} rather than exposing it in the state as we do.
In our applied domains, deep RL has been used for inventory and supply-chain
control~\cite{leluc2023marlim,khirwar2023cooperative,maggiar2025structure} and for T1DM basal and bolus
dosing~\cite{zhu2023offline,emerson2023offline,jafar2024personalized,marchetti2025deep,denesfazakas2024reinforcement} and wider
clinical decision making~\cite{wu2023value,choi2024deep,kalimouttou2025optimal,luo2024position}; these
optimise a single intervention channel, or treat dosing as a flat per-step
choice, whereas we treat short- and long-horizon modes jointly under explicit
intervention budgets.

\section{Multi-Timescale Intervention MDPs}

We consider a discounted MDP in which control is available through an immediate intervention channel and a persistent-effect intervention channel. Let $x_t\in\mathcal{X}\subseteq\mathbb{R}^{d}$ denote the system state and $z_t\in\mathcal{Z}\subseteq\mathbb{R}^{m}$ the residual intervention state. At time $t$, the agent may select an immediate intervention $\eta_t^I\in\mathcal{H}_I$, a persistent-effect intervention $\eta_t^P\in\mathcal{H}_P$, or remain inactive.

\paragraph{Persistent effects.}
The controlled dynamics are
\begin{align}
x_{t+1} &\sim P_x(\cdot\mid x_t,z_t,\eta_t^I,\eta_t^P), \label{eq:x-transition}\\
z_{t+1} &= f_z(z_t,\eta_t^P). \label{eq:z-transition}
\end{align}
The central case we consider is linear decay,
\begin{equation}
z_{t+1}=\rho z_t+B_P\eta_t^P,
\qquad \rho\in[0,1),
\label{eq:linear-persistence}
\end{equation}
although the formulation only requires $z_t$ to contain the information about previous persistent-effect interventions required to determine future transitions. Equation~\ref{eq:linear-persistence} permits residual effects from different decisions to superpose. An immediate intervention can therefore be selected at time $t$ even while persistent-effect interventions selected at earlier times continue to influence the state.

\paragraph{Intervention modes.}
We introduce a mode variable
\[
m_t\in\{0,I,P\},
\]
corresponding to inaction, immediate intervention, and persistent-effect intervention. The realised control is
\[
\eta_t(m_t)=
\begin{cases}
(0,0), & m_t=0,\\
(\eta_t^I,0), & m_t=I,\\
(0,\eta_t^P), & m_t=P.
\end{cases}
\]
The action sets $\mathcal{H}_I$ and $\mathcal{H}_P$ may be discrete or continuous. For example, long-acting intervention initiation is discrete in our T1DM experiments, whereas persistent MuJoCo actuation is continuous.

\paragraph{Objective.}
The general one-step reward takes the form
\begin{equation}
R(y_t,\eta_t)
=
r(x_t,z_t)
-
c_I(\eta_t^I)
-
c_P(\eta_t^P),
\label{eq:general-reward}
\end{equation}
where $y_t=(x_t,z_t)$; below we write $R(\hat y_t,\eta_t)$ for its extension to the augmented state. The functions $c_I$ and $c_P$ encode intervention magnitude or initiation costs. Typical rewards include regulation objectives,
tracking costs, inventory costs, or locomotion rewards.
For example, target-zone regulation is recovered by
\[
r(x_t,z_t)=-(x_t-M)^2.
\]
while MuJoCo and inventory control retain their native task
rewards together with intervention costs.

\subsection{Resource and Safety Constraints}

Let $L^i(y,\eta)\geq 0$ denote the resource cost charged to
constraint $i$, and let $n_i$ be its initial budget. We maintain
the remaining budget through
\begin{equation}
b_{t+1}^i
=
b_t^i-L^i(y_t,\eta_t),
\qquad
b_0^i=n_i,
\label{eq:budget}
\end{equation}
with $b_t=(b_t^1,\ldots,b_t^J)$. The complete augmented state is
\[
\hat y_t=(x_t,z_t,b_t).
\]

The budget cost $L^i$ is defined at the environment level and need
not coincide with the diagnostic intervention count reported in the
experiments. In particular, the experimental tables report the
number of non-null channel activations, whereas the hard budget is
applied to the corresponding environment-specific intervention
cost defined in each benchmark. We therefore distinguish throughout
between \emph{budget consumption}, which determines feasibility, and
\emph{reported channel activations}, which measure intervention use.

For hard resource constraints, the admissible action set is
\begin{equation}
\mathcal{H}(\hat y_t)
=
\left\{
\eta\in\mathcal{H}:
L^i(y_t,\eta)\leq b_t^i,\ \forall i
\right\}.
\label{eq:feasible-actions}
\end{equation}
We assume that the null intervention has zero resource cost and is
feasible in every augmented state. Feasible-action masking therefore
prevents the specified hard budget from being exceeded. This is
distinct from state-safety constraints, such as collision avoidance,
unsafe temperatures, or hypoglycaemia, which depend on future
stochastic transitions. For these constraints we optionally employ
the predictive shielding mechanism described below.

The objective is
\begin{equation}
V^\pi(\hat y)
=
\mathbb{E}_\pi
\left[
\sum_{t=0}^{\infty}\gamma^t
R(\hat y_t,\eta_t)
\mid
\hat y_0=\hat y
\right],
\label{eq:objective}
\end{equation}
subject to $\eta_t\in\mathcal{H}(\hat y_t)$ for all $t$.

\begin{proposition}[Markovisation]
\label{prop:markov}
Suppose that the conditional distribution of $x_{t+1}$
depends on the past only through $(x_t,z_t,\eta_t)$, that
$z_{t+1}=f_z(z_t,\eta_t^P)$, and that resource consumption,
reward and termination depend only on the current augmented
state and action. Then $\hat y_t=(x_t,z_t,b_t)$ is Markov.
\end{proposition}

\noindent
For the linear model~\eqref{eq:linear-persistence}, $z_t$ recursively summarises the complete history of persistent-effect interventions through
\[
z_t=\rho^t z_0+\sum_{k<t}\rho^{t-1-k}B_P\eta_k^P.
\]
Budget evolution~\eqref{eq:budget} is likewise determined entirely by the current augmented state and action. A proof is provided in the supplementary material.

\section{\textsc{MINT}: Structured Intervention Learning}

A flat policy can optimise the MTI-MDP by directly selecting from the joint action space. Our hypothesis is that explicitly exposing its intervention structure provides a useful inductive bias, particularly when persistent effects are long-lived or intervention opportunities are scarce.

\paragraph{Intervention selector and conditional policies.}
\textsc{MINT} decomposes the policy into an intervention selector
\[
g_\psi(m_t\mid \hat y_t),
\qquad m_t\in\{0,I,P\},
\]
and intervention policies
\[
\pi_I(\eta_t^I\mid\hat y_t),
\qquad
\pi_P(\eta_t^P\mid\hat y_t).
\]
The intervention policies first generate candidate controls and the intervention selector determines whether either candidate should be executed. Selecting $m_t=0$ produces inaction. Thus $g_\psi$ learns \emph{where intervention is worthwhile}, while $\pi_I$ and $\pi_P$ learn \emph{how to intervene} on the corresponding temporal scale.
This decomposition is more structured than conventional temporal abstraction:
selecting $m_t=I$ does not terminate the effects of earlier persistent-effect
interventions, whose contribution remains encoded in $z_t$, and a new
persistent-effect intervention augments rather than replaces the residual
state. Interventions initiated at different times therefore coexist through the
shared persistent-effect state while the controller keeps deciding at every
time step.

Equally importantly, the intervention selector learns intervention
timing independently of intervention magnitude. Rather than
learning a single monolithic policy over a joint action space,
\textsc{MINT} partitions the state space into regions where
inaction, immediate intervention, or persistent-effect intervention
are preferable, while the conditional policies specialise in
determining the corresponding control. This separation
aligns naturally with intervention problems in which deciding
\emph{whether} to intervene is often as important as
\emph{how}.

\paragraph{Cost-aware intervention timing.}
Let $Q_0$, $Q_I$, and $Q_P$ denote the values associated with inaction and the two intervention modes. The optimal mode satisfies
\begin{equation}
m^*(\hat y)
\in
\arg\max_{m\in\{0,I,P\}} Q_m(\hat y).
\label{eq:mode}
\end{equation}
Equation~\ref{eq:mode} shows that intervention timing
emerges as an optimisation problem in its own right.
Rather than treating intervention as another continuous
action, the controller explicitly compares the expected value
of inaction, immediate intervention, and persistent
intervention. Consequently, intervention decisions are driven
not only by immediate reward but also by their future impact
through persistent state evolution and remaining resource
availability. This interpretation connects the learned mode
selector directly to the optimal intervention problem studied
in sequential decision making.

\paragraph{Constraint handling and shielding.}
Before execution, candidate actions are restricted to the feasible set~\eqref{eq:feasible-actions}, guaranteeing adherence to intervention budgets. For environments with state-safety constraints, we use a $K$-step predictive shield. For a proposed control $\eta_t$, the shield samples short-horizon trajectories and rejects the proposal if the predicted trajectory enters a prohibited state region. The resulting action is either an alternative feasible mode or inaction. We treat this shield as a safety layer rather than a defining component of \textsc{MINT}, and enable it identically for every method.

\paragraph{Optimisation.}
Our deep implementation uses Soft Actor-Critic (SAC)~\cite{haarnoja2018soft} for both the intervention selector and the conditional intervention policies, with separate entropy temperatures for the discrete and continuous components. These choices are implementation-specific: the MTI-MDP and policy decomposition are compatible with other actor--critic or value-based learners. Algorithmic details and hyperparameters are given in the supplementary material.

\section{Analysis}

We next characterise the dynamic-programming structure of the augmented problem. Full proofs are provided in the supplementary material.
For any bounded $Q$, define the Bellman optimality operator
\begin{equation}
(\mathcal{T}Q)(\hat y,\eta)
=
R(\hat y,\eta)
+
\gamma
\mathbb{E}_{\hat y'}
\left[
\max_{\eta'\in\mathcal{H}(\hat y')}
Q(\hat y',\eta')
\right].
\label{eq:bellman}
\end{equation}

The intervention structure permits the value function to be written as
\begin{equation}
V^*(\hat y)
=
\max\{
Q_0^*(\hat y),
Q_I^*(\hat y),
Q_P^*(\hat y)
\},
\label{eq:q-decomp}
\end{equation}
where
\begin{align*}
Q_0^*(\hat y)
&=Q^*(\hat y,(0,0)),\\
Q_I^*(\hat y)
&=
\sup_{\eta^I:(\eta^I,0)\in\mathcal{H}(\hat y)}
Q^*(\hat y,(\eta^I,0)),\\
Q_P^*(\hat y)
&=
\sup_{\eta^P:(0,\eta^P)\in\mathcal{H}(\hat y)}
Q^*(\hat y,(0,\eta^P)).
\end{align*}
Equation~\ref{eq:q-decomp} formalises the separation between intervention timing and intervention content used by \textsc{MINT}.

\begin{proposition}[Budget feasibility]
\label{prop:budget}
If $b_0\succeq0$ and the controller selects $\eta_t\in\mathcal{H}(\hat y_t)$ at every step, then $b_t\succeq0$ for all $t$ almost surely.
\end{proposition}

\begin{theorem}
\label{thm:contraction}
For bounded rewards and $\gamma\in[0,1)$, $\mathcal{T}$ is a $\gamma$-contraction under the supremum norm and therefore admits a unique fixed point $Q^*$.
\end{theorem}

Proposition~\ref{prop:markov} is essential here: augmenting the physical state with $z_t$ converts the apparent history dependence of persistent-effect interventions into ordinary Markov dynamics. Augmenting further with $b_t$ preserves this property because the budgets evolve deterministically according to~\eqref{eq:budget}. The contraction proof then follows from the standard discounted Bellman argument.

For finite state and action spaces, consider the update
\begin{align*}
&Q_{t+1}(\hat y_t,\eta_t)
=
Q_t(\hat y_t,\eta_t)
\\&\quad+
\alpha_t
\left[
R_t+
\gamma
\max_{\eta'\in\mathcal{H}(\hat y_{t+1})}
Q_t(\hat y_{t+1},\eta')
-
Q_t(\hat y_t,\eta_t)
\right].
\end{align*}

\begin{theorem}
\label{thm:qlearning}
If all feasible state--action pairs are sampled infinitely often, then $Q_t\rightarrow Q^*$ almost surely.
\end{theorem}

The result follows from Theorem~\ref{thm:contraction} and classical stochastic-approximation results for Q-learning~\cite{jaakkola1994convergence}. The theorem concerns the tabular instance and does not imply convergence of our SAC-based function-approximation implementation.

\section{Experiments}
\label{sec:experiments}

Our experiments address three questions. \textbf{RQ1 (Persistence):} does explicitly modelling intervention structure become more valuable as persistent effects last? \textbf{RQ2 (Resource scarcity):} does separating intervention timing from intervention magnitude improve the performance--resource trade-off when intervention budgets bind? \textbf{RQ3 (Generality):} do these effects occur across environments with qualitatively different dynamics rather than being specific to insulin control? We evaluate on persistent-action variants of MuJoCo, stochastic inventory management, and physiological glucose control.

\paragraph{Protocol.}
Every method is trained for $200$K physical environment steps, CPU-only on a single Apple M4 Pro with $24$ GB of unified memory. Hyperparameters were chosen on development seeds disjoint from those reported here and frozen; no checkpoint selection is applied, so every number is the fixed $200$K endpoint, not a best-so-far model. All methods are evaluated on the same frozen panel of environment seeds ($10$ episodes for MuJoCo and inventory, $5$ for T1DM). We report mean $\pm$ standard deviation (SD) over $5$ seeds; intervals are two-sided Student-$t$ $95\%$ confidence intervals (CIs) treating the seed, not the episode, as the statistical unit, with Benjamini--Hochberg correction within each sweep. Two limits should be stated plainly: the evaluation panel is shared with development rather than held out, and T1DM uses a single virtual patient and meal scenario, so it is not cross-validated across patients.

\subsection{Benchmarks}

\paragraph{Persistent MuJoCo control.}
We construct a persistent-control variant of \texttt{HalfCheetah-v5}. Given immediate and persistent controls $(\eta_t^I,\eta_t^P)$, the environment receives
\[
u_t=\eta_t^I+z_t,
\qquad
z_{t+1}=\rho z_t+\eta_t^P.
\]
The agent observes $z_t$, making the augmented dynamics Markov. We vary
\(
\rho\in\{0,0.5,0.9,0.98\},
\)
where $\rho=0$ removes temporal persistence, and evaluate both unconstrained control and intervention budgets
\(
n_Z\in\{\infty,50,100,200\}.
\)
The reward is the native task reward minus immediate- and persistent-control costs. Episodes last $1000$ steps, so a policy acting on both channels at every step issues $2000$ activations. Unless stated otherwise the main setting is $\rho=0.9$ with an unconstrained budget. These experiments test whether the benefit of structured intervention learning changes with persistence and with scarcity.

\paragraph{Inventory management.}
We use an OR-Gym~\cite{hubbs2020orgym} inventory environment with stochastic demand and replenishment lead times. Standard replenishment is the persistent-effect intervention, since orders remain in the pipeline before affecting stock, while costly emergency replenishment is the immediate channel. We use the default demand process over a $100$-period horizon with a hard intervention budget of $n_Z=60$.

\paragraph{T1DM glucose regulation.}
We use GlucoEnv~\cite{hettiarachchi2022glucoenv}, which implements physiological dynamics based on the UVA/Padova T1DM simulator~\cite{man2014uva}. Bolus insulin is the immediate channel and basal insulin the persistent one. We evaluate the Ambulatory Glycaemic Variability Pattern (AGVP) meal scenario on patient \texttt{adolescent\#001} over a full $288$-step ($24$-hour) horizon, with a hard intervention budget of $n_Z=40$ per episode and predictive hypoglycaemia shielding enabled for all methods. The principal metrics are time in range (TIR; $70$--$180$ mg/dL) and time below range ($<70$ mg/dL); time above range is the remainder. This tests the framework in a safety-critical domain where the two time scales have a direct physical interpretation.

\paragraph{Baselines.}
We compare \textsc{MINT} against three classes of baselines. First, PPO~\cite{schulman2017proximal} provides a standard flat on-policy baseline. More importantly, a \emph{flat augmented-action} baseline built on SAC~\cite{haarnoja2018soft} receives the same $(x_t,z_t,b_t)$ state and the same immediate and persistent action channels as \textsc{MINT}, but selects their values jointly through a single monolithic policy, controlling for the possibility that gains arise simply from state augmentation. It is simultaneously our SAC baseline, so it appears as one row of Table~\ref{tab:main-results} rather than two.

Second, temporal-abstraction baselines: the fixed-option agent commits to an intervention at an option boundary and holds it open-loop for a fixed duration, while the augmented-option baseline additionally observes $z_t$. These test whether the mode and intervention factorisation provides value beyond conventional temporal abstraction~\cite{sutton1999between,bacon2017option}, separating gains from modelling persistent effects from those obtained by temporally abstracting policy execution.

Third, we include two constrained-RL baselines that enforce the budget through a cost signal rather than through the hard contract used by the other methods: PPO with a Lagrangian penalty on the pre-shield intervention proposal, following the convention of standard safe-RL suites~\cite{ji2023safety}, and Constrained Policy Optimisation (CPO)~\cite{achiam2017constrained}. Both inherit the environment, reward, budget, shield and evaluation panel of the other runs, changing only the algorithm. HalfCheetah carries no budget under our protocol, so the multiplier is switched off and Lagrangian PPO reduces to PPO while CPO enters its unconstrained trust-region branch; only the Inventory and T1DM columns are budgeted constrained-RL results. All methods otherwise receive identical observations, budgets, shields and rewards.

\subsection{Main Results}

\paragraph{Cross-domain performance.}
Table~\ref{tab:main-results} reports the principal results. \textsc{MINT} attains the best mean primary metric on two of three benchmarks, improving inventory return over the flat augmented policy by $35.1$ and T1DM time in range by $26.3\%$ points (unpaired $95\%$ CI $[15.0, 37.6]$, $p=0.003$); on HalfCheetah the two are indistinguishable (paired difference $-25.2$, CI $[-1573.3, 1522.9]$). It issues fewer channel activations than the flat augmented policy on every benchmark, $50.0\%$, $24.7\%$ and $8.7\%$ fewer so each buys almost twice the return on HalfCheetah ($1.65$ against $0.84$).

This economy is not paid for elsewhere. \textsc{MINT} meets $100.0\%$ of inventory demand, matching the flat augmented policy while placing $24.7\%$ fewer orders, and the fixed-option agent's much lower return is a genuine stockout failure at $62.6\%$ service rather than a scoring artefact. The low PPO time in range reflects instability, not merely poor regulation: $40\%$ of its T1DM episodes end early in a severe glycaemic excursion. Since the flat augmented baseline sees the same intervention state and action channels, the difference isolates the value of separating mode selection from intervention magnitude; a component ablation that removes the selector, the persistent state and the hard budget contract one at a time is reported in the appendix. Both temporal-abstraction baselines are far weaker on the two non-clinical benchmarks even though they act at every step, so the tested temporal-abstraction baselines do not recover
MINT's performance, and against the strongest of them on T1DM, the augmented-option agent, \textsc{MINT} raises time in range by $21.5$ percentage points (unpaired $95\%$ CI $[5.9, 37.1]$). \textsc{MINT}'s HalfCheetah variance is dominated by one collapsed seed; four of five exceed the best flat augmented seed, while seed $4$ diverges to $-398.9$, spending $99.8\%$ of its activations on the persistent channel against under $1\%$ for the healthy seeds, which we retain in all aggregates.

\begin{table*}[t]
\centering
\setlength{\tabcolsep}{4pt}
\caption{Main comparison after $200$K physical environment steps, over $5$ training seeds (mean $\pm$ SD). \emph{Interv.}\ is the mean number of channel activations per evaluation episode and \emph{Ret./interv.}\ the return each buys; \emph{Service} is the fraction of demand met from stock; \emph{Fail.}\ is the percentage of evaluation episodes ended early by a severe glycaemic excursion. Time in range is measured over the full $288$-step horizon, so steps lost to an early termination are not credited as in range. No method records a budget violation, and all hold time below range at $0.0\%$ in every seed. Intervention counts are deliberately not bolded, since a low count is informative only at matched task performance. HalfCheetah is unbudgeted, so both constrained baselines degenerate there (Lagrangian PPO to PPO, CPO to its trust-region branch) and those columns are not budgeted constrained-RL results.}
\label{tab:main-results}
\resizebox{\textwidth}{!}{%
\begin{tabular}{l ccc ccc ccc}
\toprule
& \multicolumn{3}{c}{\textbf{HalfCheetah} ($\rho=0.9$, unbudgeted)} & \multicolumn{3}{c}{\textbf{Inventory} ($n_Z=60$)} & \multicolumn{3}{c}{\textbf{T1DM} ($n_Z=40$)}\\
\cmidrule(lr){2-4}\cmidrule(lr){5-7}\cmidrule(lr){8-10}
Method & Return $\uparrow$ & Interv.\ $\downarrow$ & Ret./interv.\ $\uparrow$
       & Return $\uparrow$ & Service (\%) $\uparrow$ & Interv.\ $\downarrow$
       & TIR (\%) $\uparrow$ & Fail.\ (\%) $\downarrow$ & Interv.\ $\downarrow$\\
\midrule
PPO~\cite{schulman2017proximal} & $1297 \pm 188$ & $2000$ & $0.65 \pm 0.09$ & $-294 \pm 148$ & $98.6 \pm 3.0$ & $68.2 \pm 5.9$ & $12.6 \pm 1.0$ & $40.0$ & $41.7 \pm 1.5$\\
Flat augmented SAC~\cite{haarnoja2018soft} & $\mathbf{1676 \pm 152}$ & $2000$ & $0.84 \pm 0.08$ & $-254 \pm 30$ & $\mathbf{100.0 \pm 0.0}$ & $51.6 \pm 4.3$ & $64.7 \pm 9.1$ & $0.0$ & $43.8 \pm 0.8$\\
Fixed options & $48 \pm 376$ & $2000$ & $0.02 \pm 0.19$ & $-1505 \pm 169$ & $62.6 \pm 5.5$ & $67.9 \pm 4.3$ & $65.8 \pm 9.4$ & $0.0$ & $44.4 \pm 1.9$\\
Augmented options & $-63 \pm 248$ & $2000$ & $-0.03 \pm 0.12$ & $-378 \pm 76$ & $99.3 \pm 1.1$ & $68.6 \pm 4.7$ & $69.5 \pm 12.6$ & $0.0$ & $44.6 \pm 2.9$\\
\addlinespace[1pt]
Lagrangian PPO~\cite{ji2023safety} & $1297 \pm 188$ & $2000$ & $0.65 \pm 0.09$ & $-1874 \pm 953$ & $61.5 \pm 37.1$ & $40.3 \pm 24.7$ & $11.1 \pm 0.0$ & $32.0$ & $30.7 \pm 14.3$\\
CPO~\cite{achiam2017constrained} & $170 \pm 259$ & $2000$ & $0.09 \pm 0.13$ & $-1323 \pm 131$ & $64.1 \pm 1.5$ & $75.0 \pm 0.0$ & $11.9 \pm 1.5$ & $44.0$ & $27.0 \pm 17.4$\\
\addlinespace[1pt]
\midrule
\textsc{MINT} (ours) & $1650 \pm 1146$ & $1000$ & $\mathbf{1.65 \pm 1.15}$ & $\mathbf{-219 \pm 24}$ & $\mathbf{100.0 \pm 0.0}$ & $38.9 \pm 2.4$ & $\mathbf{90.9 \pm 0.9}$ & $0.0$ & $40.0 \pm 0.0$\\
\bottomrule
\end{tabular}}
\end{table*}

\paragraph{Constrained-RL baselines.}
Neither cost-based baseline is competitive where its constraint is active: on Inventory both return below $-1300$ at under $65\%$ service, worse than every hard-contract method. On T1DM both fall to the floor, $11.1\%$ and $11.9\%$ time in range, and the failure is legible: in almost every episode they keep exactly the $32$ steps already in range at reset and never bring glucose back, settling at $255$--$375$ mg/dL. They fail by under-dosing, which is why their time below range stays at $0.0\%$ and their intervention counts are the lowest in the table: a scalar cost term discourages the very interventions the task needs, and one Lagrangian PPO seed issues $6.8$ activations across the horizon. Two readings must be separated, however. The T1DM split follows the learner family, not the constraint mechanism: all three on-policy methods land between $11$ and $13\%$ while every method with an off-policy critic reaches $64\%$ or more, and at $200$K steps a $2000$-step PPO rollout spans under seven $288$-step episodes, giving $100$ policy updates against roughly $195$K off-policy gradient updates. We therefore do not read this column as evidence that constrained RL cannot regulate glucose; the like-for-like comparison for \textsc{MINT} remains flat augmented SAC. What these baselines isolate is the constraint mechanism: because the budget enters \textsc{MINT}'s state and feasible-action set rather than a penalty term, it never trades task reward against constraint satisfaction.

\begin{figure}[t]
    \centering
    \includegraphics[width=\columnwidth]{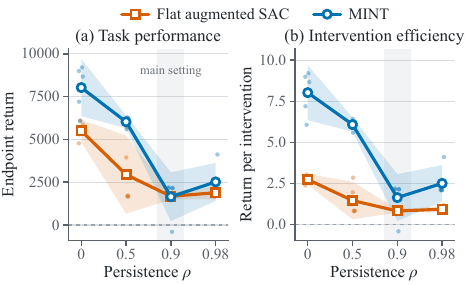}
    \caption{Effect of the persistent-action parameter $\rho$ on \texttt{HalfCheetah-v5}. (a) Endpoint return; (b) return per channel activation. Faint dots are training seeds, lines seed means, bands two-sided Student-$t$ $95\%$ CIs ($n=5$); the shaded column is the $\rho=0.9$ main setting.}
\vspace{-.25cm}\label{fig:persistence}
\end{figure}

\paragraph{Effect of persistence.}
Figure~\ref{fig:persistence} tests RQ1 by varying $\rho$ with everything else fixed. At $\rho=0$ the persistent channel becomes a second immediate actuator; even here \textsc{MINT} improves mean return by $2521.7$ (unadjusted $95\%$ CI $[438.6, 4604.9]$) while acting on half as many channels. As persistence increases both methods degrade, and the raw-return gap is \emph{non-monotonic}: it peaks at $\rho=0.5$ ($+3084.2$), vanishes at $\rho=0.9$ ($-25.2$) and partially recovers at $\rho=0.98$ ($+629.5$), no individual contrast reaching $q<0.05$ after Benjamini--Hochberg correction. We therefore do not claim the return advantage grows with persistence. Panel (b) behaves differently: reaching those returns on half as many activations earns \textsc{MINT} $+5.27$, $+4.62$, $+0.81$ and $+1.57$ more return per intervention, three of the four contrasts surviving correction. Explicitly representing intervention modes buys a better trade-off in every regime tested, not a return advantage that grows monotonically in $\rho$.

\begin{figure}[t]
    \centering
    \includegraphics[width=\columnwidth]{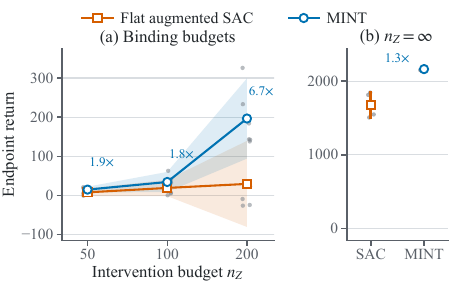}
    \caption{Performance--resource trade-off on \texttt{HalfCheetah-v5} ($\rho=0.9$) as the hard budget $n_Z$ varies. Both panels plot endpoint return and \textsc{MINT}'s return ratio over the flat policy. (a) The three binding budgets on one linear scale; (b) the unconstrained $n_Z=\infty$, excluding the one \textsc{MINT} run that collapsed onto the persistent channel ($n=5$). Faint dots are seeds, open markers seed means, bands and whiskers two-sided Student-$t$ $95\%$ CIs.}
    \vspace{-.25cm}\label{fig:budget}
\end{figure}

\paragraph{Performance under intervention scarcity.}
We next vary the intervention budget at $\rho=0.9$, holding everything else fixed (Figure~\ref{fig:budget}). Both degrade sharply as interventions become scarce, but \textsc{MINT} retains a higher mean return at every binding budget: $14.8$ versus $7.8$, $34.3$ versus $19.1$ and $196.5$ versus $29.4$ at $n_Z=50$, $100$ and $200$. The resource audit identifies the mechanism. Both exhaust
exactly $n_Z$ activations without violation, but a monolithic
continuous policy cannot abstain on one channel: the flat policy
spends two budget units per decision and so acts at $n_Z/2$ time
steps, whereas \textsc{MINT}'s selector commits to one channel and
acts at $n_Z$. Turning the same activation budget into twice as
many decision times is what separating timing from magnitude
buys; the sweep holds that budget fixed rather than the decision
count. It also reallocates its activations as the budget tightens, placing $41.4\%$ and $40.8\%$ of them on the persistent channel at $n_Z=50$ and $100$ against $19.6\%$ at $n_Z=200$: when interventions are rationed it leans on the channel whose effect outlives the decision, as Equation~\ref{eq:mode} predicts. Two caveats temper the claim: the gap is not ordered by scarcity, being largest at the loosest binding budget $n_Z=200$ ($+167.1$); and no individual contrast survives correction (smallest $q=0.24$), so it is the consistent sign across all three binding budgets that carries the result. The bands in Figure~\ref{fig:budget} are correspondingly wide: a binding budget leaves the agent unactuated for most of a $1000$-step episode, so return is small in absolute terms and sensitive to where the few interventions land. The spread is structured rather than diffuse: at $n_Z=200$ four of five flat seeds finish below $23$ while all five \textsc{MINT} seeds exceed $137$, and at $n_Z=\infty$ one \textsc{MINT} run collapses onto the persistent channel ($998$ of $1000$ activations, $-398.9$ against $2146$--$2196$). Panel (b) excludes it, leaving \textsc{MINT} ahead by $487.2$ (Welch $95\%$ CI $[299.9, 674.4]$); retaining it, as Table~\ref{tab:main-results} does, the two are indistinguishable ($-25.2$).

\paragraph{Discussion and limitations.}
\textsc{MINT}'s main advantage is selective allocation of intervention
channels, and it survives at $\rho=0$, so part of the gain comes from
abstention and mode selection rather than persistence alone. Limitations: the
formulation uses a linear decay model although the MTI-MDP permits general
Markov persistent dynamics; the convergence result is tabular; and the T1DM
evaluation is simulated, establishing no clinical safety or efficacy.

\section{Conclusion}

We introduced the MTI-MDP and \textsc{MINT} for reinforcement learning with
immediate and persistent intervention effects. Augmenting the state with
residual effects and remaining budgets, and separating intervention selection
from conditional control, supports overlapping interventions, explicit timing
and hard resource constraints. Across MuJoCo, inventory control and T1DM,
\textsc{MINT} achieves a stronger performance--intervention trade-off than
flat, temporal-abstraction and constrained-RL baselines, particularly when
budgets bind.

\makeatletter
\providecommand\stublabel[2]{\global\@namedef{r@#1}{#2}}
\makeatother
\stublabel{eq:budget}{{6}{3}{}{equation.6}{}}
\stublabel{eq:feasible-actions}{{7}{4}{}{equation.7}{}}
\stublabel{eq:mode}{{9}{4}{}{equation.9}{}}
\stublabel{prop:markov}{{1}{4}{}{proposition.1}{}}
\stublabel{thm:qlearning}{{2}{5}{}{theorem.2}{}}
\stublabel{tab:main-results}{{1}{7}{}{table.1}{}}

\bibliography{sample}

@inproceedings{bacon2017option,
  title={The option-critic architecture},
  author={Bacon, Pierre-Luc and Harb, Jean and Precup, Doina},
  booktitle={Proceedings of the AAAI conference on artificial intelligence},
  volume={31},
  year={2017}
}

@article{tsitsiklis1999optimal,
  title={Optimal stopping of Markov processes: Hilbert space theory, approximation algorithms, and an application to pricing high-dimensional financial derivatives},
  author={Tsitsiklis, John N and Van Roy, Benjamin},
  journal={IEEE Transactions on Automatic Control},
  volume={44},
  number={10},
  pages={1840--1851},
  year={1999},
  publisher={IEEE}
}

@inproceedings{jaakkola1994convergence,
  title={Convergence of stochastic iterative dynamic programming algorithms},
  author={Jaakkola, Tommi and Jordan, Michael I and Singh, Satinder P},
  booktitle={Advances in neural information processing systems},
  pages={703--710},
  year={1994}
}

@article{cryer2010hypoglycemia,
  title={Hypoglycemia in type 1 diabetes mellitus},
  author={Cryer, Philip E},
  journal={Endocrinology and Metabolism Clinics},
  volume={39},
  number={3},
  pages={641--654},
  year={2010},
  publisher={Elsevier}
}

@inproceedings{bastani2021safe,
  title={Safe reinforcement learning with nonlinear dynamics via model predictive shielding},
  author={Bastani, Osbert},
  booktitle={2021 American control conference (ACC)},
  pages={3488--3494},
  year={2021},
  organization={IEEE}
}

@inproceedings{sootla2022saute,
  title={Saut{\'e} rl: Almost surely safe reinforcement learning using state augmentation},
  author={Sootla, Aivar and Cowen-Rivers, Alexander I and Jafferjee, Taher and Wang, Ziyan and Mguni, David H and Wang, Jun and Ammar, Haitham},
  booktitle={International Conference on Machine Learning},
  pages={20423--20443},
  year={2022},
  organization={PMLR}
}

@misc{hettiarachchi2022glucoenv,
     author={Hettiarachchi, Chirath},
     title={GluCoEnv v0.1.0(2022)},
     year = {2022},
     publisher = {GitHub},
     journal = {GitHub repository},
     howpublished = {\url{https://github.com/chirathyh/GluCoEnv}},
   }

@article{klissarov2021flexible,
  title={Flexible option learning},
  author={Klissarov, Martin and Precup, Doina},
  journal={Advances in Neural Information Processing Systems},
  volume={34},
  pages={4632--4646},
  year={2021}
}

@article{man2014uva,
  title={The UVA/PADOVA type 1 diabetes simulator: new features},
  author={Man, Chiara Dalla and Micheletto, Francesco and Lv, Dayu and Breton, Marc and Kovatchev, Boris and Cobelli, Claudio},
  journal={Journal of diabetes science and technology},
  volume={8},
  number={1},
  pages={26--34},
  year={2014},
  publisher={SAGE Publications Sage CA: Los Angeles, CA}
}

@article{jafar2024personalized,
  title={Personalized insulin dosing using reinforcement learning for high-fat meals and aerobic exercises in type 1 diabetes: a proof-of-concept trial},
  author={Jafar, Adnan and Kobayati, Alessandra and Tsoukas, Michael A and Haidar, Ahmad},
  journal={Nature Communications},
  volume={15},
  number={1},
  pages={6585},
  year={2024},
  publisher={Nature Publishing Group UK London}
}

@article{zhu2023offline,
  title={Offline deep reinforcement learning and off-policy evaluation for personalized basal insulin control in type 1 diabetes},
  author={Zhu, Taiyu and Li, Kezhi and Georgiou, Pantelis},
  journal={IEEE Journal of Biomedical and Health Informatics},
  volume={27},
  number={10},
  pages={5087--5098},
  year={2023},
  publisher={IEEE}
}

@inproceedings{haarnoja2018soft,
  title={Soft actor-critic: Off-policy maximum entropy deep reinforcement learning with a stochastic actor},
  author={Haarnoja, Tuomas and Zhou, Aurick and Abbeel, Pieter and Levine, Sergey},
  booktitle={International conference on machine learning},
  pages={1861--1870},
  year={2018},
  organization={PMLR}
}

@article{hubbs2020orgym,
  title={{OR-Gym}: A Reinforcement Learning Library for Operations Research Problems},
  author={Hubbs, Christian D. and Perez, Hector D. and Sarwar, Owais and Sahinidis, Nikolaos V. and Grossmann, Ignacio E. and Wassick, John M.},
  journal={arXiv preprint arXiv:2008.06319},
  year={2020}
}

@inproceedings{achiam2017constrained,
  title={Constrained Policy Optimization},
  author={Achiam, Joshua and Held, David and Tamar, Aviv and Abbeel, Pieter},
  booktitle={International Conference on Machine Learning},
  pages={22--31},
  year={2017},
  organization={PMLR}
}

@article{schulman2017proximal,
  author    = {John Schulman and
               Filip Wolski and
               Prafulla Dhariwal and
               Alec Radford and
               Oleg Klimov},
  title     = {Proximal Policy Optimization Algorithms},
  journal={arXiv preprint arXiv:1707.06347},
  year={2017}
}

@article{sutton1999between,
  title={Between MDPs and semi-MDPs: A framework for temporal abstraction in reinforcement learning},
  author={Sutton, Richard S and Precup, Doina and Singh, Satinder},
  journal={Artificial intelligence},
  volume={112},
  number={1-2},
  pages={181--211},
  year={1999},
  publisher={Elsevier}
}

@InProceedings{pmlr-v202-mguni23a,
  title = 	 {{MANSA}: Learning Fast and Slow in Multi-Agent Systems},
  author =       {Mguni, David Henry and Chen, Haojun and Jafferjee, Taher and Wang, Jianhong and Yue, Longfei and Feng, Xidong and Mcaleer, Stephen Marcus and Tong, Feifei and Wang, Jun and Yang, Yaodong},
  booktitle = 	 {Proceedings of the 40th International Conference on Machine Learning},
  pages = 	 {24631--24658},
  year = 	 {2023},
  volume = 	 {202},
  series = 	 {Proceedings of Machine Learning Research},
  month = 	 {23--29 Jul},
  publisher =    {PMLR},
}

@inproceedings{mguni2022timing,
  title={Timing is {Everything}: Learning to Act Selectively with Costly Actions and Budgetary Constraints},
  author={Mguni, David and Sootla, Aivar and Ziomek, Juliusz and Slumbers, Oliver and Dai, Zipeng and Shao, Kun and Wang, Jun},
  booktitle={International Conference on Learning Representations},
  year={2023}
}

@book{sutton2018reinforcement,
  title={Reinforcement learning: An introduction},
  author={Sutton, Richard S and Barto, Andrew G},
  year={2018},
  publisher={MIT press}
}

@article{klissarov2025discovering,
  title={Discovering Temporal Structure: An Overview of Hierarchical Reinforcement Learning},
  author={Klissarov, Martin and Bagaria, Akhil and Luo, Ziyan and Konidaris, George and Precup, Doina and Machado, Marlos C.},
  journal={arXiv preprint arXiv:2506.14045},
  year={2025}
}

@inproceedings{li2024when,
  title={When Do Skills Help Reinforcement Learning? A Theoretical Analysis of Temporal Abstractions},
  author={Li, Zhening and Poesia, Gabriel and Solar-Lezama, Armando},
  booktitle={Proceedings of the 41st International Conference on Machine Learning},
  pages={28568--28596},
  year={2024},
  organization={PMLR}
}

@inproceedings{zadem2024reconciling,
  title={Reconciling Spatial and Temporal Abstractions for Goal Representation},
  author={Zadem, Mehdi and Mover, Sergio and Nguyen, Sao Mai},
  booktitle={International Conference on Learning Representations},
  year={2024}
}

@article{hu2025state,
  title={State-novelty guided action persistence in deep reinforcement learning},
  author={Hu, Jianshu and Weng, Paul and Ban, Yutong},
  journal={Machine Learning},
  volume={114},
  number={2},
  pages={45},
  year={2025},
  publisher={Springer}
}

@inproceedings{sabbioni2023simultaneously,
  title={Simultaneously Updating All Persistence Values in Reinforcement Learning},
  author={Sabbioni, Luca and Al Daire, Luca and Bisi, Lorenzo and Metelli, Alberto Maria and Restelli, Marcello},
  booktitle={Proceedings of the AAAI Conference on Artificial Intelligence},
  volume={37},
  pages={9668--9676},
  year={2023}
}

@inproceedings{lee2024learning,
  title={Learning Uncertainty-Aware Temporally-Extended Actions},
  author={Lee, Joongkyu and Park, Seung Joon and Tang, Yunhao and Oh, Min-hwan},
  booktitle={Proceedings of the AAAI Conference on Artificial Intelligence},
  volume={38},
  pages={13391--13399},
  year={2024}
}

@inproceedings{karimi2023dynamic,
  title={Dynamic Decision Frequency with Continuous Options},
  author={Karimi, Amirmohammad and Jin, Jun and Luo, Jun and Mahmood, A. Rupam and J{\"a}gersand, Martin and Tosatto, Samuele},
  booktitle={2023 IEEE/RSJ International Conference on Intelligent Robots and Systems (IROS)},
  pages={7545--7552},
  year={2023},
  organization={IEEE}
}

@inproceedings{wachi2024survey,
  title={A Survey of Constraint Formulations in Safe Reinforcement Learning},
  author={Wachi, Akifumi and Shen, Xun and Sui, Yanan},
  booktitle={Proceedings of the Thirty-Third International Joint Conference on Artificial Intelligence},
  pages={8262--8271},
  year={2024}
}

@inproceedings{ji2023safety,
  title={Safety Gymnasium: A Unified Safe Reinforcement Learning Benchmark},
  author={Ji, Jiaming and Zhang, Borong and Zhou, Jiayi and Pan, Xuehai and Huang, Weidong and Sun, Ruiyang and Geng, Yiran and Zhong, Yifan and Dai, Juntao and Yang, Yaodong},
  booktitle={Advances in Neural Information Processing Systems},
  volume={36},
  pages={18964--18993},
  year={2023}
}

@article{konighofer2025shields,
  title={Shields for Safe Reinforcement Learning},
  author={K{\"o}nighofer, Bettina and Bloem, Roderick and Jansen, Nils and Junges, Sebastian and Pranger, Stefan},
  journal={Communications of the ACM},
  volume={68},
  number={11},
  pages={80--90},
  year={2025},
  publisher={ACM}
}

@inproceedings{banerjee2024dynamic,
  title={Dynamic Model Predictive Shielding for Provably Safe Reinforcement Learning},
  author={Banerjee, Arko and Rahmani, Kia and Biswas, Joydeep and Dillig, Isil},
  booktitle={Advances in Neural Information Processing Systems},
  volume={37},
  pages={100131--100159},
  year={2024}
}

@inproceedings{varricchione2024pure,
  title={Pure-Past Action Masking},
  author={Varricchione, Giovanni and Alechina, Natasha and Dastani, Mehdi and De Giacomo, Giuseppe and Logan, Brian and Perelli, Giuseppe},
  booktitle={Proceedings of the AAAI Conference on Artificial Intelligence},
  volume={38},
  pages={21646--21655},
  year={2024}
}

@article{corsi2024verification,
  title={Verification-Guided Shielding for Deep Reinforcement Learning},
  author={Corsi, Davide and Amir, Guy and Rodr{\'i}guez, Andoni and S{\'a}nchez, C{\'e}sar and Katz, Guy and Fox, Roy},
  journal={Reinforcement Learning Journal},
  volume={4},
  pages={1759--1780},
  year={2024}
}

@inproceedings{lin2023safe,
  title={Safe Offline Reinforcement Learning with Real-Time Budget Constraints},
  author={Lin, Qian and Tang, Bo and Wu, Zifan and Yu, Chao and Mao, Shangqin and Xie, Qianlong and Wang, Xingxing and Wang, Dong},
  booktitle={Proceedings of the 40th International Conference on Machine Learning},
  pages={21127--21152},
  year={2023},
  organization={PMLR}
}

@inproceedings{cen2024feasibility,
  title={Feasibility Consistent Representation Learning for Safe Reinforcement Learning},
  author={Cen, Zhepeng and Yao, Yihang and Liu, Zuxin and Zhao, Ding},
  booktitle={Proceedings of the 41st International Conference on Machine Learning},
  pages={6002--6019},
  year={2024},
  organization={PMLR}
}

@article{pignatelli2024survey,
  title={A Survey of Temporal Credit Assignment in Deep Reinforcement Learning},
  author={Pignatelli, Eduardo and Ferret, Johan and Geist, Matthieu and Mesnard, Thomas and van Hasselt, Hado and Pietquin, Olivier and Toni, Laura},
  journal={Transactions on Machine Learning Research},
  year={2024}
}

@inproceedings{wu2024boosting,
  title={Boosting Reinforcement Learning with Strongly Delayed Feedback Through Auxiliary Short Delays},
  author={Wu, Qingyuan and Zhan, Simon Sinong and Wang, Yixuan and Wang, Yuhui and Lin, Chung-Wei and Lv, Chen and Zhu, Qi and Schmidhuber, J{\"u}rgen and Huang, Chao},
  booktitle={Proceedings of the 41st International Conference on Machine Learning},
  pages={53973--53998},
  year={2024},
  organization={PMLR}
}

@article{emerson2023offline,
  title={Offline reinforcement learning for safer blood glucose control in people with type 1 diabetes},
  author={Emerson, Harry and Guy, Matthew and McConville, Ryan},
  journal={Journal of Biomedical Informatics},
  volume={142},
  pages={104376},
  year={2023},
  publisher={Elsevier}
}

@article{marchetti2025deep,
  title={Deep reinforcement learning for Type 1 Diabetes: Dual {PPO} controller for personalized insulin management},
  author={Marchetti, Alessandro and Sasso, Daniele and D'Antoni, Federico and Morandin, Francesco and Parton, Maurizio and Matarrese, Margherita Anna Grazia and Merone, Mario},
  journal={Computers in Biology and Medicine},
  volume={191},
  pages={110147},
  year={2025},
  publisher={Elsevier}
}

@article{denesfazakas2024reinforcement,
  title={Reinforcement Learning: A Paradigm Shift in Personalized Blood Glucose Management for Diabetes},
  author={D{\'e}nes-Fazakas, Lehel and Szil{\'a}gyi, L{\'a}szl{\'o} and Kov{\'a}cs, Levente and De Gaetano, Andrea and Eigner, Gy{\"o}rgy},
  journal={Biomedicines},
  volume={12},
  number={9},
  pages={2143},
  year={2024},
  publisher={MDPI}
}

@article{boughton2024role,
  title={The role of automated insulin delivery technology in diabetes},
  author={Boughton, Charlotte K. and Hovorka, Roman},
  journal={Diabetologia},
  volume={67},
  number={10},
  pages={2034--2044},
  year={2024},
  publisher={Springer}
}

@article{nwokolo2023artificial,
  title={The Artificial Pancreas and Type 1 Diabetes},
  author={Nwokolo, Munachiso and Hovorka, Roman},
  journal={The Journal of Clinical Endocrinology \& Metabolism},
  volume={108},
  number={7},
  pages={1614--1623},
  year={2023},
  publisher={Oxford University Press}
}

@article{wu2023value,
  title={A value-based deep reinforcement learning model with human expertise in optimal treatment of sepsis},
  author={Wu, XiaoDan and Li, RuiChang and He, Zhen and Yu, TianZhi and Cheng, ChangQing},
  journal={npj Digital Medicine},
  volume={6},
  number={1},
  pages={15},
  year={2023},
  publisher={Nature Publishing Group}
}

@article{kalimouttou2025optimal,
  title={Optimal Vasopressin Initiation in Septic Shock: The {OVISS} Reinforcement Learning Study},
  author={Kalimouttou, Alexandre and Kennedy, Jason N. and Feng, Jean and Singh, Harvineet and Saria, Suchi and Angus, Derek C. and Seymour, Christopher W. and Pirracchio, Romain},
  journal={JAMA},
  volume={333},
  number={19},
  pages={1688--1698},
  year={2025},
  publisher={American Medical Association}
}

@article{choi2024deep,
  title={Deep reinforcement learning extracts the optimal sepsis treatment policy from treatment records},
  author={Choi, Yunho and Oh, Songmi and Huh, Jin Won and Joo, Ho-Taek and Lee, Hosu and You, Wonsang and Bae, Cheng-mok and Choi, Jae-Hun and Kim, Kyung-Joong},
  journal={Communications Medicine},
  volume={4},
  number={1},
  pages={245},
  year={2024},
  publisher={Nature Publishing Group}
}

@inproceedings{luo2024position,
  title={Position: Reinforcement Learning in Dynamic Treatment Regimes Needs Critical Reexamination},
  author={Luo, Zhiyao and Pan, Yangchen and Watkinson, Peter and Zhu, Tingting},
  booktitle={Proceedings of the 41st International Conference on Machine Learning},
  pages={33432--33465},
  year={2024},
  organization={PMLR}
}

@article{leluc2023marlim,
  title={{MARLIM}: Multi-Agent Reinforcement Learning for Inventory Management},
  author={Leluc, R{\'e}mi and Kadoche, Elie and Bertoncello, Antoine and Gourv{\'e}nec, S{\'e}bastien},
  journal={arXiv preprint arXiv:2308.01649},
  year={2023}
}

@inproceedings{khirwar2023cooperative,
  title={Cooperative Multi-agent Reinforcement Learning for Inventory Management},
  author={Khirwar, Madhav and Gurumoorthy, Karthik S. and Jain, Ankit Ajit and Manchenahally, Shantala},
  booktitle={Machine Learning and Knowledge Discovery in Databases: Applied Data Science and Demo Track (ECML PKDD)},
  pages={619--634},
  year={2023},
  organization={Springer}
}

@article{maggiar2025structure,
  title={Structure-Informed Deep Reinforcement Learning for Inventory Management},
  author={Maggiar, Alvaro and Andaz, Sohrab and Bagaria, Akhil and Eisenach, Carson and Foster, Dean and Gottesman, Omer and Perrault-Joncas, Dominique},
  journal={arXiv preprint arXiv:2507.22040},
  year={2025}
}

@article{towers2024gymnasium,
  title={Gymnasium: A Standard Interface for Reinforcement Learning Environments},
  author={Towers, Mark and Kwiatkowski, Ariel and Terry, Jordan K. and Balis, John U. and De Cola, Gianluca and Deleu, Tristan and Goul{\~a}o, Manuel and Kallinteris, Andreas and Krimmel, Markus and KG, Arjun and Perez-Vicente, Rodrigo and Pierr{\'e}, Andrea and Schulhoff, Sander and Tai, Jun Jet and Tan, Hannah and Younis, Omar G.},
  journal={arXiv preprint arXiv:2407.17032},
  year={2024}
}
%
\makeatletter
\renewcommand{\citep}[1]{(\csname apxcite@#1\endcsname)}
\@namedef{apxcite@tsitsiklis1999optimal}{Tsitsiklis and Van Roy 1999}
\@namedef{apxcite@jaakkola1994convergence}{Jaakkola, Jordan, and Singh 1994}
\@namedef{apxcite@sootla2022saute}{Sootla et al. 2022}
\makeatother

\clearpage
\onecolumn
\appendix
\setcounter{secnumdepth}{2}
\setlength{\arrayrulewidth}{0.4pt}
\makeatletter
\renewcommand{\@seccntformat}[1]{\csname the#1\endcsname.\hspace{0.55em}}
\def\section{\@startsection{section}{1}{\z@}%
  {-16pt plus -4pt minus -2pt}{7pt plus 2pt minus 1pt}{\large\bf\raggedright}}
\def\subsection{\@startsection{subsection}{2}{\z@}%
  {-12pt plus -3pt minus -2pt}{4pt plus 1pt}{\normalsize\bf\raggedright}}
\makeatother
\renewcommand{\topfraction}{0.94}
\renewcommand{\bottomfraction}{0.75}
\renewcommand{\textfraction}{0.05}
\renewcommand{\floatpagefraction}{0.80}
\setcounter{topnumber}{4}
\setcounter{bottomnumber}{3}
\setcounter{totalnumber}{6}
{\raggedright\LARGE Appendix\par}
\vspace{4pt}

\noindent
This supplement documents everything needed to interpret and reproduce the
results reported in the main text, and it adds the evidence that did not fit in
the body. It is organised around a single question: \emph{which parts of
\textsc{MINT} are actually doing the work?} Section~\ref{app:setup} fixes the
experimental contract, meaning the environments, the observation and action
spaces, the resource contract, the predictive shield, the evaluation protocol,
and the exact software and hardware used. Section~\ref{app:hyper} lists every
hyperparameter for \textsc{MINT} and for all six baselines.
Section~\ref{app:ablation} reports the component ablation: three single-factor
removals on the T1DM benchmark, each holding the environment, budget, shield,
network, optimiser and training seed fixed, so that every comparison is paired
by construction. It shows that the intervention selector is the load-bearing
component, that exposing the persistent state contributes a smaller but
consistent further gain, and that \textsc{MINT}'s hard resource contract costs a
few points of regulation quality while a soft penalty in its place abandons the
contract entirely. Section~\ref{app:additional} then presents cross-domain
evidence at a level of detail the body could not accommodate: per-seed endpoints
for all $105$ formal runs, channel-allocation and shield-activation accounting,
the decision-time mechanism behind the budget result, full numeric tables for
both sweeps, and a step-level anatomy of one closed-loop T1DM episode.
Sections~\ref{app:algorithms} onwards give the algorithm listings, the
spectral-decay construction, the relationship between the cases, and the proofs
of the theoretical results. Every number reported here comes from the same
frozen artefacts as the main text: the fixed $200$K-step endpoint of each run,
evaluated on a frozen panel of environment seeds, with no checkpoint selection
and no seed dropped. Where a result is descriptive rather than inferential, and
in particular for the single-seed ablations of Section~\ref{app:ablation}, we
say so explicitly and attach no significance test. The figures in this
supplement report quantities at a granularity the body does not, and no figure
is duplicated between the body and the supplement.

\medskip
\noindent
\textbf{A note on two figures quoted in the main text.} A reader reconciling the
main text against Table~\ref{tab:main-results} will find an apparent
one-tenth-of-a-point discrepancy in the headline time-in-range comparisons, and
we resolve it here rather than leave it to inference. The main text states that
\textsc{MINT} improves time in range over the strongest baseline, the
augmented-option agent, by $21.5$ percentage points, and over the flat augmented
policy by $26.3$ points. Subtracting the entries printed in
Table~\ref{tab:main-results} gives $90.9-69.5=21.4$ and $90.9-64.7=26.2$
instead. Neither number is a typographical error, because the gap is double
rounding. The seed means before rounding are $90.9444$ for \textsc{MINT},
$69.4722$ for augmented options and $64.6528$ for the flat augmented policy, so
the true differences are $21.4722$ and $26.2917$, which round \emph{once} to
$21.5$ and $26.3$. Subtracting values that have already been rounded to one
decimal place rounds a second time and loses the tenth. All contrasts in the
main text and in this supplement are computed from unrounded seed means and
rounded once at the end, while the table entries are rounded independently for
display. Where the two conventions differ we state both, as here. The
accompanying confidence interval is unaffected: the unpaired Welch $95\%$
interval on the augmented-option contrast is $[5.8712, 37.0732]$, printed as
$[5.9, 37.1]$.

\section{Table of Notation}
\label{app:notation}
\begin{table}[h]
    \centering
    \begin{tabular}{c c c}
         \toprule
 \multicolumn{3}{c}{\textit{Variables}} \\\hline
         \textbf{Name}  & \textbf{Range}  & \textbf{Meaning} \\
         \midrule
         $Z^F(t)$ & $\mathbb{R}_{\geq 0}$ & immediate-acting intervention process   \\
        $Z^L(t)$ & $\mathbb{R}_{\geq 0}$ & Long-acting intervention process   \\
        $\bmZ(t)$ & $\mathbb{R}^2_{\geq 0}$ & Dual Intervention process   \\
         $X(t)$ & $\mathbb{R}^d_{\geq 0}$ & Underlying bodily process   \\ 
         $Y(t)$&$\mathbb{R}^d_{\geq 0}$  & State Process\\ 
        $W(t)$ &$\mathbb{R}^{(\dim(K)\times 1)}$  & Brownian
motion process for process $K$\\
         $E(t)$ & $[0,1]$ & Long control variate scaling factor \\
         $F$ & $[0,1]$ & Long control variate scaling distribution \\
         \midrule
          \multicolumn{3}{c}{\textit{Constants}} \\
         \midrule
         $M$& $\mathbb{R}_{\geq 0}$  & Target value   \\
         $l$ & $\mathbb{R}_{\geq 0}$  & Tolerance range   \\
         $N_0$& $\mathbb{N}$ & Violation tolerance  \\ 
         $n_Z$&$\mathbb{N}$  & Intervention count budget   \\
         $\sigma_K$& $\mathbb{R}^{(\dim(K)\times 1)}$  & Brownian
motion process scaling parameter for process $K$\\
        $\alpha$ &$\mathbb{R}_{\geq 0}$ & Long-acting intervention cost parameter \\
        $\beta$ &$\mathbb{R}_{\geq 0}$ & immediate-acting intervention cost parameter \\
        $\gamma$ &$[0,1)$ & Cost objective discount factor \\
        $\Delta$ &$\mathbb{R}_{>0}\times\{\infty\}$ & Constraint violation cost \\
         \midrule
          \multicolumn{3}{c}{\textit{Controls}} \\\hline
        $\eta^I$&$\mathbb{R}_{\geq 0}$ & immediate-acting intervention control variate \\
        $\eta^P$&$\{0,1\}$ & Long-acting intervention control variate \\
        $\rho_k$&$\mathcal{F}$ & $k^{\rm th}$ immediate-acting intervention time\\
        $\tau_k$&$\mathcal{F}$ &$k^{\rm th}$ immediate-acting intervention time \\   \midrule
          \multicolumn{3}{c}{\textit{Sets}} \\\hline
        $\mathcal{H}^F$&$\mathbb{R}$ & immediate-acting intervention control set \\
        $\mathcal{H}^L$&$\mathbb{R}$ & Long-acting intervention control set \\
        $\bm\cY$&$\mathbb{R}^m$ & State space \\
        $\bm{\hat{\cY}}$&$\mathbb{R}^j$ & Augmented state space \\
        $\bm\cB$&$\mathbb{R}^q$ & Space of budget constraints \\
        $B^i$& $\mathbb{R}^l$ & i$^{th}$ constraint set \\
        $\bm{\mathcal{H}}$&$\mathbb{R}^p$ & Joint intervention control set \\
        $\cE$& $[0,1]^r$ & Finite spectra set \\
        $\boldsymbol{\mathcal{X}}$&$\mathbb{R}^w$ & Process $X$ space   \\
        \midrule
                  \multicolumn{3}{c}{\textit{Functions}} \\\hline
        $v^\pi_M$ & $\mathbb{R}$ & Case M Value function under $\pi$ \\
        $Q^\pi_M$&$\mathbb{R}$ & Case M action-value function under $\pi$\\
         \bottomrule
    \end{tabular}
    \caption{Table of notation.}
    \label{tab:my_label}
\end{table}
\clearpage

\clearpage
\section{Experimental Setup and Reproducibility}
\label{app:setup}

\subsection{Benchmarks and the Augmented State Contract}
\label{app:envs}

Table~\ref{tab:app-envs} specifies the three benchmarks exactly as run. The
important column is the decomposition of the agent-facing observation, because
it makes Proposition~\ref{prop:markov} operational: the physical observation is
augmented by the residual intervention state $z_t$ and, where a hard budget is
active, by the remaining-budget coordinates $b_t$. Nothing else is added. Every
method in the comparison receives the identical augmented observation and the
identical two-channel action, with the single deliberate exception of the
fixed-option baseline, from which $z_t$ is withheld precisely so that the
augmented-option baseline isolates what observing it is worth. Any remaining
performance difference is therefore attributable to how a method uses that
information, not to what it can see. How much that exception removes differs
sharply across the three domains, and the table states it explicitly so that the
fixed-option rows are read correctly. On T1DM the residual state is one
coordinate of the $29$ non-budget coordinates and on HalfCheetah it is six of
$23$, but on Inventory the outstanding-order pipeline accounts for $30$ of the
$33$, so the fixed-option baseline there sees on-hand stock and its remaining
budget and very little else. Its weak Inventory row in
Table~\ref{tab:main-results} should be attributed partly to that, and it is the
augmented-option baseline, which does observe the pipeline, that carries the
temporal-abstraction comparison in this domain.
Two details deserve emphasis. First, on the budgeted benchmarks the augmented
observation includes a normalised time-to-go coordinate wherever the horizon is
finite and the budget renews (Inventory), because a finite-horizon budgeted
process is Markov in $(x_t,z_t,b_t)$ only once remaining time is observable: the
same physical state and the same remaining budget call for different actions at
the start and at the end of an episode. Second, the persistent channel is
genuinely part of the environment dynamics rather than a policy-level
abstraction. The recursion $z_{t+1}=\rho z_t+B_P\eta_t^P$ is applied by the
environment wrapper on every physical step, whether or not the agent acts, which
is precisely what distinguishes a persistent-effect intervention from an option:
committing to $m_t=I$ at time $t$ does not terminate the residual effect of a
persistent intervention initiated earlier.

\begin{table}[!ht]
\centering
\caption{Benchmark specifications as run. The observation decomposition shows
how the augmented state $\hat y_t=(x_t,z_t,b_t)$ of
Proposition~\ref{prop:markov} is realised in each domain. All methods in the
comparison receive the identical augmented observation and the identical
two-channel action space in each column, apart from the fixed-option baseline,
which by design does not observe $z_t$.}
\label{tab:app-envs}
\small
\setlength{\tabcolsep}{5pt}
\begin{tabular}{@{}p{0.235\textwidth} p{0.228\textwidth} p{0.228\textwidth} p{0.228\textwidth}@{}}
\toprule
& \textbf{Persistent HalfCheetah} & \textbf{Inventory} & \textbf{T1DM} \\
\midrule
Base simulator & MuJoCo \texttt{HalfCheetah-v5} & OR-Gym \texttt{InvManagement-v1} (lost sales), vendored verbatim & GluCoEnv (UVA/Padova T1DM model) \\
Episode horizon & $1000$ steps & $100$ periods & $288$ steps ($24$\,h at $5$\,min) \\
Stochasticity & MuJoCo initial state & Poisson demand ($\mu=20$), lead times $L=(3,5,10)$ & meal timing/size, CGM sensor noise \\
\addlinespace[2pt]
Physical observation $x_t$ & $17$ & $3$ (on-hand stock, one per stage) & $28$ ($12$ CGM lags, $12$ insulin lags, meal announcement and countdown, time-of-day sine/cosine) \\
Residual state $z_t$ & $6$ & $30$ (outstanding-order pipeline, $3$ stages $\times$ $10$ lead-time slots) & $1$ \\
Budget coordinates $b_t$ & $0$ (unbudgeted main setting) & $3$ (total, immediate, time-to-go) & $2$ (total, immediate) \\
\textbf{Agent observation} $\hat y_t$ & $\mathbf{23}$ & $\mathbf{36}$ & $\mathbf{31}$ \\
\addlinespace[2pt]
Action dimension & $12$ ($6$ immediate $+$ $6$ persistent) & $6$ ($3+3$) & $2$ ($1+1$) \\
Immediate channel $\eta^I$ & instantaneous actuation & emergency replenishment & bolus insulin \\
Persistent channel $\eta^P$ & residual actuation & standard replenishment (supply pipeline) & basal insulin \\
Persistence $\rho$ & $0.9$ main, with $\{0,0.5,0.9,0.98\}$ swept & pipeline lead time & $0.9$ \\
Persistent gain $B_P$ & $1.0$ & $1.0$ & $0.2$ \\
\addlinespace[2pt]
Hard budget $n_Z$ & $\infty$ main, with $\{50,100,200\}$ swept & $60$ (at most $15$ immediate) & $40$ (at most $12$ immediate) \\
Budget renewal & n/a & per episode & per $288$-step day \\
Predictive shield & no state-safety term (budget term only when a budget binds) & budget term only & budget term $+$ glucose term \\
\addlinespace[2pt]
Reward & native task reward $-\,c_I-\,c_P$ & target-zone inventory cost & target-zone glucose reward, terminal penalty $25$ \\
Cost weights $(\alpha_P,\beta_I)$ & $(0.01,\,0.001)$ & $(0.02,\,0.1)$ & $(0.005,\,0.0005)$ \\
Reward scale & $1.0$ & $0.01$ & $1.0$ \\
\addlinespace[2pt]
Evaluation seeds & $33000$--$33009$ ($10$ episodes) & $31000$--$31009$ ($10$) & $32000$--$32004$ ($5$) \\
Primary metric & return & return & time in range, full horizon \\
\bottomrule
\end{tabular}
\end{table}

\subsection{The Resource Contract: Hard Feasibility versus Soft Penalty}
\label{app:resource}

The hard resource contract of Equation~\ref{eq:feasible-actions} is implemented
as an environment-level wrapper, not as an agent-level heuristic, and it is
therefore identical for every method. Table~\ref{tab:app-resource} states the
two mechanisms we compare. Under the hard contract, the remaining budget enters
both the observation and the admissible action set: once $b_t^i$ reaches zero
the only feasible action on the corresponding channel is the null intervention,
so violations are impossible by construction rather than improbable in
expectation. This is why
Table~\ref{tab:main-results} can record zero budget violations for
\emph{every} hard-contract method, ours and the baselines alike. The contract is
a property of the environment interface, and we claim no credit for satisfying
it. What we do claim is that satisfying it cheaply requires structure, and that
is exactly what Sections~\ref{app:ablation} and~\ref{app:decision-times}
demonstrate. The alternative mechanism is a scalar penalty
$-\lambda L^i(y_t,\eta_t)$ added to the reward. It is the mechanism through
which the two constrained-RL baselines of the main text address the budget, and
the third ablation of Section~\ref{app:ablation} isolates it by putting it in
place of the feasibility constraint altogether. The penalty is a soft signal, in
that it changes the objective the learner optimises but neither bounds
consumption nor makes remaining resources observable.
Section~\ref{app:no-budget} quantifies exactly what that costs.

\begin{table}[!ht]
\centering
\caption{The two resource mechanisms compared in this work. The hard contract is
the environment-level mechanism under which \textsc{MINT} and the
temporal-abstraction and flat baselines operate. The soft penalty is how the two
constrained-RL baselines address the budget, and it stands on its own in the
third ablation of Section~\ref{app:ablation}.}
\label{tab:app-resource}
\small
\begin{tabular}{@{}p{0.225\textwidth} p{0.35\textwidth} p{0.35\textwidth}@{}}
\toprule
& \textbf{Hard contract (\textsc{MINT} and hard-contract baselines)} & \textbf{Soft penalty (constrained-RL baselines and ablation)} \\
\midrule
Budget in observation & yes: $b_t$ normalised to $[0,1]$ per constraint & no \\
Budget in action set & yes: $\mathcal{H}(\hat y_t)$ masks infeasible channels & no \\
Guarantee & consumption $\leq n_Z$ almost surely, by construction & none, since expected cost is traded against reward \\
Failure mode when tight & agent must learn \emph{when} to spend & agent under- or over-spends depending on $\lambda$ \\
Multiplier tuning & none required & requires $\lambda$ or a dual ascent schedule \\
Reported as & \emph{budget consumption} (feasibility) and \emph{channel activations} (usage) & diagnostic counters only \\
\bottomrule
\end{tabular}
\end{table}

\subsection{Predictive Safety Shield}
\label{app:shield}

The shield is a safety layer, not a component of \textsc{MINT}, and it is
enabled identically for every method wherever the domain has a state-safety
constraint. It has two terms, reported separately throughout this supplement
because they measure very different things. The \emph{budget term} is exact and
free. Budgets evolve deterministically by Equation~\ref{eq:budget}, so a
one-step lookahead needs no simulator rollout: the wrapper replaces any proposal
that would exceed a remaining budget with the null intervention. The
\emph{glucose term} is the $K$-step predictive component of the main text. It
monitors the sensed glucose trend, projects it forward $K$ steps, and blocks an
insulin proposal if the projection enters the hypoglycaemic region or if the
projected residual insulin state would exceed a physiological cap. Blocking in
both cases means substituting the null intervention rather than rescaling the
proposal, on both channels when the trend or floor check fires and on the
persistent channel alone when the cap check fires, so that a simultaneous
immediate correction is preserved whenever only the residual tail is unsafe.
Table~\ref{tab:app-shield} gives the settings. A shield activation is therefore
not a failure, since it is the safety layer doing its job, but \emph{which} term
activates is highly informative about the policy underneath it, and we exploit
that in Section~\ref{app:shield-decomp}. A policy whose budget term fires
constantly is a policy that has not learned to abstain, whereas a policy whose
activations are almost entirely physiological is one whose resource behaviour is
already correct before the shield sees it.

\begin{table}[!ht]
\centering
\caption{Predictive shield settings. Enabled identically for every method
including all baselines. HalfCheetah carries no state-safety constraint under
our protocol, so no predictive term is applied there, and its budget term is
inactive in the unbudgeted main setting and engages only in the budget sweep of
Section~\ref{app:decision-times}.}
\label{tab:app-shield}
\small
\begin{tabular}{@{}llll@{}}
\toprule
Parameter & T1DM & Inventory & HalfCheetah \\
\midrule
Budget term (exact one-step) & enabled & enabled & budget sweep only \\
Glucose term ($K$-step predictive) & enabled & n/a & n/a \\
Prediction horizon $K$ & $6$ steps ($30$\,min) & n/a & n/a \\
Glucose floor $\underline{g}$ & $90$ mg/dL & n/a & n/a \\
Danger threshold & $70$ mg/dL & n/a & n/a \\
Persistent-state cap $\bar z$ & $0.8$ & n/a & n/a \\
Replacement action & null intervention on both channels when the glucose term fires, null intervention on the persistent channel alone when the cap term fires & null intervention & n/a \\
\bottomrule
\end{tabular}
\end{table}

\subsection{Evaluation Protocol and Statistics}
\label{app:protocol}

Every method is trained for exactly $200$K physical environment steps. There is
no checkpoint selection: the reported number is always the fixed $200$K
endpoint, taken from the final model, never a best-so-far validation
checkpoint. Checkpoints are stored every $20$K steps and are used only for the
descriptive convergence audits of Section~\ref{app:convergence} and for the
learning curves in Figure~\ref{fig:app-ablation-curves}, and they never feed a
reported endpoint. All methods are evaluated on the same frozen panel of
environment seeds listed in Table~\ref{tab:app-envs}.
We report the mean $\pm$ standard deviation over $5$ training seeds
($0$--$4$), treating the seed rather than the episode as the statistical unit.
Confidence intervals are two-sided Student-$t$ $95\%$ intervals on $n=5$,
paired contrasts pair by training seed, and within each sweep we apply a
Benjamini--Hochberg correction across the contrasts of that sweep. No seed is
dropped anywhere, including the single collapsed \textsc{MINT} HalfCheetah seed
analysed in Section~\ref{app:collapse}. Three limitations of this protocol
should be stated plainly, and none of them is repaired by anything in this
supplement. Hyperparameters were selected on development seeds disjoint from the
reported training seeds but using the same evaluation episode panel, so the
panel is shared with development rather than held out. The T1DM benchmark uses a
single virtual patient (\texttt{adolescent\#001}) and a single meal scenario
(AGVP), so there is no cross-patient validation. And the ablations of
Section~\ref{app:ablation} use one training seed each, so they are descriptive
comparisons and not estimates with sampling error.

\subsection{Software and Hardware}
\label{app:software}

All experiments were run on CPU only. No GPU was used at any stage, in training,
evaluation or analysis. Table~\ref{tab:app-software} lists the environment, and
the frozen \texttt{pip freeze} manifest and a SHA-256 checksum for every
produced artefact accompany each run directory.

\begin{table}[!ht]
\centering
\caption{Software and hardware environment. Identical for every run reported in
the paper and in this supplement.}
\label{tab:app-software}
\small
\begin{tabular}{@{}p{0.26\textwidth}p{0.68\textwidth}@{}}
\toprule
Component & Setting \\
\midrule
Processor & Apple M4 Pro, $12$ cores \\
Memory & $24$ GB unified \\
Accelerator & none, all runs CPU-only \\
Operating system & macOS $15.7$ (Darwin $24.6$) \\
Python & $3.13.12$ \\
\midrule
\texttt{torch} & $2.13.0$ \\
\texttt{numpy} & $2.5.1$ \\
\texttt{scipy} & $1.18.0$ \\
\texttt{gymnasium} & $1.3.0$ \\
\texttt{stable-baselines3} & $2.9.0$ \\
\texttt{mujoco} & $3.10.0$ \\
\texttt{torchdiffeq} & $0.2.5$ \\
\texttt{pandas} & $3.0.3$ \\
\midrule
Determinism & per-run seeding of Python, NumPy and \texttt{torch}, with frozen evaluation seeds \\
Archived per run & final model, optimiser and training state, replay buffer, chronological transition archive, $10$ intermediate checkpoints, $10$ frozen validation evaluations, resolved config, \texttt{pip freeze}, SHA-256 manifest \\
\bottomrule
\end{tabular}
\end{table}

\subsection{Compute Budget}
\label{app:compute}

Table~\ref{tab:app-compute} reports measured wall-clock time, online parameter
count and number of gradient-update iterations for every method on every
benchmark. Two observations matter for interpreting the main results.
First, \textsc{MINT} is the most expensive method in the comparison, by roughly
$2.2\times$ over the flat augmented policy on the two cheap benchmarks and
$1.28\times$ on T1DM, and it carries the largest parameter count ($373$K on
T1DM against $225$K) because it maintains a selector in addition to the two
conditional intervention policies. We report this openly, since the gains
documented in this paper are not free and a reader budgeting compute should know
the factor. The comparison is nonetheless controlled where it counts, namely in
environment interaction, where every method sees exactly $200$K physical steps,
and in gradient updates, where \textsc{MINT}'s $195{,}001$ iterations are
within one update of the flat policy's $195{,}000$. \textsc{MINT} does not win
by taking more optimisation steps.
Second, the on-policy and off-policy families are separated by two orders of
magnitude in gradient updates ($1{,}000$ for PPO and Lagrangian PPO, $50$
trust-region updates for CPO, against $\approx\!195$K for every off-policy
method) at equal environment interaction. This is a property of the learner
families at a $200$K-step budget, not of the constraint mechanisms, and it is
why the main text declines to read the T1DM constrained-RL column as evidence
that constrained RL cannot regulate glucose. The like-for-like comparison for
\textsc{MINT} is and remains the flat augmented SAC policy, which matches it in
observation, action space, optimiser family, network width and gradient-update
count.

\begin{table}[!ht]
\centering
\caption{Measured compute. Wall-clock is mean $\pm$ SD over the $5$ training
seeds, on the CPU-only machine of Table~\ref{tab:app-software}. Parameter and
gradient-update counts are for the T1DM runs at seed $0$ and are representative
of the other domains up to input-dimension differences. Physical environment
steps are fixed at $200$K for every entry in the table.}
\label{tab:app-compute}
\small
\setlength{\tabcolsep}{6pt}
\begin{tabular}{@{}lccccc@{}}
\toprule
& \multicolumn{3}{c}{\textbf{Training wall-clock (hours)}} & \multicolumn{2}{c}{\textbf{T1DM, seed $0$}} \\
\cmidrule(lr){2-4}\cmidrule(lr){5-6}
Method & HalfCheetah & Inventory & T1DM & Online params & Gradient updates \\
\midrule
PPO & $0.03\pm0.01$ & $0.03\pm0.01$ & $5.17\pm0.07$ & $148{,}741$ & $1{,}000$\\
Flat augmented SAC & $0.27\pm0.03$ & $0.29\pm0.00$ & $6.06\pm0.05$ & $224{,}518$ & $195{,}000$\\
Fixed options & $0.25\pm0.07$ & $0.31\pm0.01$ & $5.93\pm0.04$ & $223{,}750$ & $195{,}012$\\
Augmented options & $0.25\pm0.08$ & $0.36\pm0.01$ & $5.74\pm0.03$ & $224{,}518$ & $195{,}009$\\
Lagrangian PPO & $0.03\pm0.01$ & $0.04\pm0.00$ & $7.54\pm0.02$ & $148{,}741$ & $1{,}000$\\
CPO & $0.03\pm0.00$ & $0.04\pm0.00$ & $7.47\pm0.05$ & $12{,}518$ & $50$\\
\midrule
\textsc{MINT} (ours) & $0.60\pm0.12$ & $0.64\pm0.12$ & $7.77\pm0.05$ & $373{,}257$ & $195{,}001$\\
\bottomrule
\end{tabular}
\end{table}

The T1DM column is dominated by the physiological simulator rather than by the
learner: the UVA/Padova ordinary differential equations are integrated at every
one of the $200$K steps, which is why even PPO, with a thousandth of the
gradient updates, still takes over five hours. The learner accounts for the
$2.6$-hour spread across the column, the simulator for the five-hour floor.

\section{Hyperparameter Settings}
\label{app:hyper}

\subsection{Tuning Protocol}

Hyperparameters were selected once, on development seeds ($100$, $101$ in the
first stage and $105$ in the second) that are disjoint from the reported
training seeds $0$--$4$, and then frozen for every campaign in the paper. No
hyperparameter was retuned per benchmark after the formal runs began, no run was
repeated because of its score, and no seed was substituted. Where a value
differs across benchmarks in Table~\ref{tab:app-hp-mint} it is because the
search selected it during that one tuning stage, not because it was adjusted
afterwards. The shared backbone, consisting of $256\times256$ multilayer
perceptrons, Adam at $3\times10^{-4}$, batch size $256$, $\gamma=0.99$,
$\tau=0.005$, $5$K warm-up steps and one gradient step per environment step, is
deliberately the \emph{same} for \textsc{MINT} and for the flat augmented SAC
baseline, so the comparison between them isolates the policy decomposition and
nothing else.

\subsection{\textsc{MINT}}

\begin{table}[!ht]
\centering
\caption{\textsc{MINT} hyperparameters. The optimiser block is shared verbatim
with the flat augmented SAC baseline of Table~\ref{tab:app-hp-baselines}, so
that the two differ only in the policy decomposition. $\texttt{auto\_0.2}$
denotes automatic entropy tuning initialised at $0.2$.}
\label{tab:app-hp-mint}
\small
\setlength{\tabcolsep}{6pt}
\begin{tabular}{@{}llll@{}}
\toprule
Hyperparameter & HalfCheetah & Inventory & T1DM \\
\midrule
\multicolumn{4}{@{}l}{\textit{Architecture}}\\
Selector $g_\psi$ network & $256\times256$ MLP, ReLU & $256\times256$ & $256\times256$\\
Intervention policies $\pi_I,\pi_P$ & $256\times256$ MLP, ReLU & $256\times256$ & $256\times256$\\
Critics (twin, per channel) & $256\times256$ MLP, ReLU & $256\times256$ & $256\times256$\\
Mode-action contract & \texttt{exact\_active\_channel} & \texttt{exact\_active\_channel} & \texttt{exact\_active\_channel}\\
Online parameters & $373{,}277$ & $383{,}761$ & $373{,}257$\\
\addlinespace[2pt]
\multicolumn{4}{@{}l}{\textit{Optimisation}}\\
Optimiser & Adam & Adam & Adam \\
Learning rate (all heads) & $3\times10^{-4}$ & $3\times10^{-4}$ & $3\times10^{-4}$\\
Batch size & $256$ & $256$ & $256$\\
Replay capacity & $200{,}000$ & $200{,}000$ & $200{,}000$\\
Warm-up steps before learning & $5{,}000$ & $5{,}000$ & $5{,}000$\\
Train frequency & every step & every step & every step\\
Gradient steps per update & $1$ & $1$ & $1$\\
Discount $\gamma$ & $0.99$ & $0.99$ & $0.99$\\
Target smoothing $\tau$ & $0.005$ & $0.005$ & $0.005$\\
\addlinespace[2pt]
\multicolumn{4}{@{}l}{\textit{Entropy temperatures}}\\
Continuous, $\alpha_{\text{cont}}$ & $0.2$ (fixed) & \texttt{auto\_0.2} & \texttt{auto\_0.2}\\
Discrete selector, $\alpha_{\text{switch}}$ & $0.1$ & $0.001$ & $0.01$\\
\addlinespace[2pt]
\multicolumn{4}{@{}l}{\textit{Normalisation}}\\
Observation normalisation & running mean/variance & running mean/variance & running mean/variance\\
Reward normalisation & off & off & off\\
Action gate & n/a & positive, threshold $0$ & positive, threshold $0$\\
Persistent action clipping & on & on & on\\
\addlinespace[2pt]
\multicolumn{4}{@{}l}{\textit{Problem constants}}\\
Persistence $\rho$ & $0.9$ & lead-time pipeline & $0.9$\\
Persistent gain $B_P$ & $1.0$ & $1.0$ & $0.2$\\
Persistent cost $\alpha_P$ & $0.01$ & $0.02$ & $0.005$\\
Immediate cost $\beta_I$ & $0.001$ & $0.1$ & $0.0005$\\
Budget violation cost $\Delta$ & n/a & $100$ & $25$\\
Hard budget $(n_Z, n_I)$ & n/a & $(60, 15)$ & $(40, 12)$\\
\bottomrule
\end{tabular}
\end{table}

Reward normalisation is switched \emph{off} everywhere. This is deliberate and
worth a sentence, because it is the setting under which an earlier version of
our inventory results collapsed: with running reward normalisation active, the
scale of the target-zone penalty drifts as the policy improves, and the selector
receives a non-stationary comparison between $Q_0$, $Q_I$ and $Q_P$ in
Equation~\ref{eq:mode}. Since the mode decision is a comparison of values rather
than a magnitude, it is more sensitive to reward rescaling than a monolithic
continuous policy is. We instead apply a single fixed reward scale per benchmark
(Table~\ref{tab:app-envs}) and leave the normaliser off for every method
including the baselines.

\subsection{Baselines}

Table~\ref{tab:app-hp-baselines} lists every baseline configuration. Two of
them deserve to be read closely, because they are the controls that make the
comparison in Table~\ref{tab:main-results} interpretable. The flat augmented
SAC baseline shares \textsc{MINT}'s optimiser block verbatim, so it differs from
\textsc{MINT} in exactly one respect: it emits both intervention channels from a
single monolithic policy instead of routing them through a learned selector. The
augmented-option baseline differs from the fixed-option baseline in exactly one
respect: it observes the residual state $z_t$. Together they separate the value
of persistent-state information from the value of temporal abstraction and from
the value of mode selection.

\begin{table}[!ht]
\centering
\caption{Baseline hyperparameters. Where a value is shared with
\textsc{MINT} it is stated as such, to make explicit that the flat augmented SAC
baseline is a genuine like-for-like control: it receives the same augmented
observation, the same two-channel action space, the same optimiser, the same
network width and the same number of gradient updates, and differs from
\textsc{MINT} only in emitting both channels from a single monolithic policy.}
\label{tab:app-hp-baselines}
\small
\setlength{\tabcolsep}{6pt}
\begin{tabular}{@{}p{0.30\textwidth} p{0.64\textwidth}@{}}
\toprule
Method & Settings \\
\midrule
\textbf{PPO} & \texttt{MlpPolicy}, actor and critic $256\times256$ ReLU, learning rate $3\times10^{-4}$, rollout $n_{\text{steps}}=2000$, minibatch $125$, $10$ epochs per update, $\gamma=0.99$, GAE $\lambda=0.95$, initial log-std $-1.0$, observation normalisation on, reward normalisation off, observation clip $10$. Yields $100$ policy updates at $200$K steps.\\
\addlinespace[3pt]
\textbf{Flat augmented SAC} & Identical to \textsc{MINT}'s optimiser block: $256\times256$ ReLU actor and twin critics, learning rate $3\times10^{-4}$, replay $200$K, batch $256$, $\gamma=0.99$, $\tau=0.005$, warm-up $5$K, one gradient step per environment step, entropy \texttt{auto\_0.2}. One monolithic policy emits the full augmented action $(\eta^I_t,\eta^P_t)$ at every physical step, with no selector and no option holding, and with the persistent state exposed.\\
\addlinespace[3pt]
\textbf{Fixed options} & A SAC actor selects a full continuous dual-channel intervention $(\eta^I_t,\eta^P_t)$ at each option boundary and that action is then held open-loop for $k=3$ physical steps, $256\times256$ ReLU actor and twin critics, learning rate $3\times10^{-4}$ (Adam), replay $100$K, batch $256$, $\gamma=0.99$, Polyak $\tau=0.005$, warm-up $5$K, $3$ gradient steps per option transition, initial entropy $0.2$. Replay holds semi-Markov transitions whose option return is the discounted intra-option reward sum $\sum_i\gamma^i r_{t+i}$ with $\gamma^k$ bootstrapping. A continuous option is the stronger baseline here because it can express the signed multidimensional actuation that MuJoCo requires, which a fixed magnitude grid cannot. Residual state $z_t$ \emph{not} exposed.\\
\addlinespace[3pt]
\textbf{Augmented options} & As fixed options, with $z_t$ exposed in the observation. This isolates temporal abstraction from persistent-state information.\\
\addlinespace[3pt]
\textbf{Lagrangian PPO} & PPO block above, plus a Lagrangian multiplier on the pre-shield intervention proposal count: limit $40$ (T1DM) / $60$ (Inventory), multiplier learning rate $10^{-3}$, initialised at $0$, updated every $2000$ steps, cost measured per step on proposals. Hard task budget and shield retained, so the multiplier is an additional soft term rather than a replacement.\\
\addlinespace[3pt]
\textbf{CPO} & PyTorch port of the reference implementation (\texttt{jachiam/cpo}, commit \texttt{5c83925}), hidden sizes $(64,32)$, rollout $4000$ steps, $\gamma=0.995$, GAE $\lambda=0.95$, cost discount $1.0$ and cost GAE $\lambda=1.0$, cost limit $40$ (T1DM) / $60$ (Inventory), trust region $\delta_{\text{KL}}=0.01$, $10$ conjugate-gradient iterations, damping $10^{-5}$, subsample $0.2$, backtrack ratio $0.8$ with at most $15$ backtracks, value learning rate $10^{-3}$ with $40$ iterations, initial log-std $-1.0$. Yields $50$ trust-region updates at $200$K steps.\\
\bottomrule
\end{tabular}
\end{table}

\subsection{Legacy Switching-Agent Configuration}
\label{sec:hyperparameters}

For completeness we retain the hyperparameter grid of the earlier
switching-control formulation from which the present method descends. Values in
square brackets denote the ranges searched during that earlier round of
performance tuning. These settings are \emph{not} those of the runs reported in
the main text, which are given in Tables~\ref{tab:app-hp-mint}
and~\ref{tab:app-hp-baselines}, and they are recorded here so that the lineage
of the design is auditable.

\begin{center}
\small
    \begin{tabular}{c|c}
        \toprule
        Clip Gradient Norm & 1\\
        $\gamma_{E}$ & 0.99\\
        $\lambda$ & 0.95\\
        Learning rate & $1$x$10^{-4}$ \\
        Number of minibatches & 4\\
        Number of optimisation epochs & 4\\
        Number of parallel actors & 16\\
        Optimisation algorithm & Adam\\
        Rollout length & 128\\
        Sticky action probability & 0.25\\
        Use Generalised Advantage Estimation & True\\
        \midrule
        Coefficient of extrinsic reward & [1, 5]\\
        Coefficient of intrinsic reward & [1, 2, 5, 10, 20, 50]\\
        Switching agent discount factor & 0.99\\
        Probability of terminating option & [0.5, 0.75, 0.8, 0.9, 0.95]\\
        $L$ function output size & [2, 4, 8, 16, 32, 64, 128, 256]\\
        \bottomrule
    \end{tabular}
\end{center}

\section{Component Ablation on T1DM}
\label{app:ablation}

\subsection{Design and One-Factor Contract}

We ablate on T1DM AGVP, the benchmark where the two time scales carry their
clearest physical meaning: basal insulin is the persistent-effect intervention
and bolus insulin the immediate one, the resource budget is a real clinical
constraint on injections, and the safety region is a documented physiological
target. Each ablation removes exactly one contract term relative to full
\textsc{MINT} and holds \emph{everything} else fixed: the patient, the meal
scenario, the horizon, the persistence parameter, the reward, the safety shield,
the network architecture, the optimiser, the replay capacity, the training seed,
the validation panel, the evaluation panel and the $200$K-step budget.
Table~\ref{tab:app-ablation-contract} records what changed and what did not.
Because only one factor moves, the comparison against full \textsc{MINT} at the
same seed is paired by construction and no cross-seed variance enters it. The
price of this design is that each row is a single training seed, so the
comparisons are descriptive. We fixed the reference run (seed $0$ of the
\textsc{MINT} T1DM campaign) and the ablation seed \emph{before} observing any
ablation result, forbade score-based reruns, seed substitution, early stopping
and checkpoint substitution, and froze each ablation configuration together with
a SHA-256 hash of the configuration, the training script and the reference
configuration before training started. We attach no significance test to these
rows and make no cross-seed claim from them.

\begin{table}[!ht]
\centering
\caption{One-factor ablation contract. The final row shows the resulting
agent-facing observation dimension: removing the persistent-state exposure
deletes the single $z_t$ coordinate, and replacing the hard budget by a soft
penalty deletes the two remaining-budget coordinates. The action space is
unchanged in every row, so every configuration retains access to both
intervention channels.}
\label{tab:app-ablation-contract}
\small
\setlength{\tabcolsep}{5pt}
\begin{tabular}{@{}p{0.205\textwidth} p{0.238\textwidth} p{0.238\textwidth} p{0.238\textwidth}@{}}
\toprule
& \textbf{No switching selector} & \textbf{No persistent-state exposure} & \textbf{No hard budget} \\
\midrule
Factor removed & the learned mode selector $g_\psi$ & $z_t$ from the agent observation & hard budget state and feasibility masking \\
Operational change & one monolithic SAC policy emits $(\eta^I_t,\eta^P_t)$ at every physical step & \texttt{expose\_persistent\_}\-\texttt{state}:\ true $\rightarrow$ false & \texttt{budget.mode}: hard $\rightarrow$ soft, $\lambda=0.1$ reward penalty \\
\addlinespace[2pt]
Persistent dynamics $z_{t+1}=\rho z_t+B_P\eta^P_t$ & retained & retained (still driven, merely unobserved) & retained \\
Both action channels available & yes & yes & yes \\
Safety shield & retained & retained & retained \\
Budget enforced & yes, $n_Z=40$ & yes, $n_Z=40$ & \textbf{no} \\
Budget observable & yes & yes & no \\
\addlinespace[2pt]
Seed / steps & $0$ / $200$K & $0$ / $200$K & $0$ / $200$K \\
Entropy scheme & SAC \texttt{auto\_0.2} (no discrete head exists) & unchanged & unchanged \\
Observation dimension & $31$ & $30$ & $29$ \\
\bottomrule
\end{tabular}
\end{table}

\subsection{Results}

\begin{figure}[!ht]
\centering
\includegraphics[width=\textwidth]{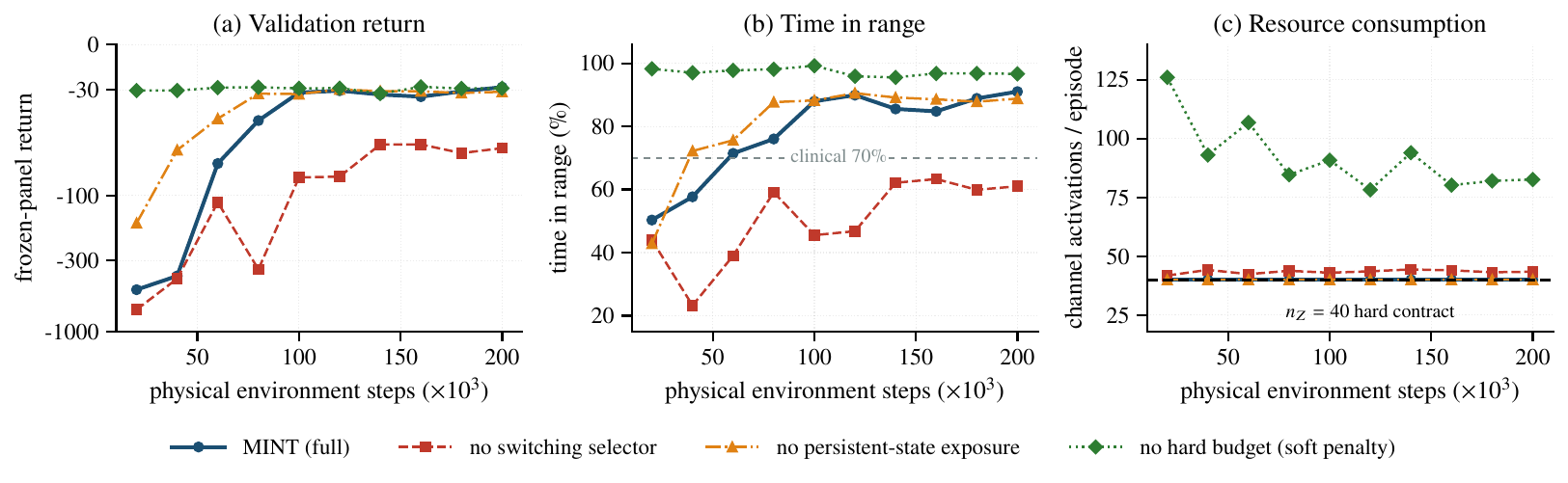}
\caption{Frozen-panel validation curves for full \textsc{MINT} and the three
ablations on T1DM AGVP, seed $0$, evaluated every $20$K physical steps on the
same five held-fixed evaluation seeds. (a) Return on a symmetric-log axis.
(b) Time in range over the full $288$-step horizon. (c) Channel activations per
episode against the $n_Z=40$ contract. These curves are diagnostic only: every
number reported in Table~\ref{tab:ablation} is the fixed $200$K endpoint, not a
best checkpoint from these traces. Panel (c) is the whole argument about the
no-budget configuration in one line, since it never approaches the contract from
below at any point in training.}
\label{fig:app-ablation-curves}
\end{figure}

Table~\ref{tab:ablation} reports the endpoints and
Figure~\ref{fig:app-ablation-curves} the frozen-panel learning curves. The
headline is that the three components are not interchangeable: they contribute
on very different scales, and the ordering matches the design argument of the
main text rather than cutting across it.

\begin{table}[!ht]
\centering
\caption{Component ablation on T1DM AGVP (patient \texttt{adolescent\#001},
seed $0$, $200$K physical steps, $5$ frozen evaluation episodes, mean over
episodes). Each row changes one contract term relative to \textsc{MINT} and
holds everything else fixed. \emph{Interv.}\ is mean channel activations per
episode and \emph{Budget} states whether the $n_Z=40$ contract is enforced.
Every configuration holds time below range at $0.0\%$ and completes every
episode. The no-budget row consumes $82.6$ activations because no contract
constrains it. That is not a budget violation, since there is no budget to
violate, and it is exactly the point of that ablation.}
\label{tab:ablation}
\small
\setlength{\tabcolsep}{5pt}
\resizebox{\textwidth}{!}{%
\begin{tabular}{@{}lcccccccc@{}}
\toprule
Configuration & Budget & Return $\uparrow$ & TIR (\%) $\uparrow$ & TBR (\%) $\downarrow$ & Interv.\ $\downarrow$ & Immediate & Persistent & Mean BG \\
\midrule
\textsc{MINT} (full) & $n_Z=40$ & $\mathbf{-28.27}$ & $91.04$ & $0.00$ & $\mathbf{40.0}$ & $0.0$ & $40.0$ & $152.8$\\
\midrule
$-$ switching selector & $n_Z=40$ & $-68.64$ & $61.04$ & $0.00$ & $43.4$ & $12.0$ & $31.4$ & $176.3$\\
$-$ persistent-state exposure & $n_Z=40$ & $-31.33$ & $88.82$ & $0.00$ & $40.0$ & $0.2$ & $39.8$ & $154.9$\\
$-$ hard budget ($\lambda=0.1$) & none & $-28.99$ & $\mathbf{96.67}$ & $0.00$ & $82.6$ & $29.8$ & $52.8$ & $144.3$\\
\midrule
\multicolumn{9}{@{}l}{\textit{Change relative to full \textsc{MINT} at the same seed}}\\
$-$ switching selector & & $-40.37$ & $-30.00$ & $0.00$ & $+3.4$ & $+12.0$ & $-8.6$ & $+23.5$\\
$-$ persistent-state exposure & & $-3.05$ & $-2.22$ & $0.00$ & $\pm0.0$ & $+0.2$ & $-0.2$ & $+2.1$\\
$-$ hard budget ($\lambda=0.1$) & & $-0.72$ & $+5.63$ & $0.00$ & $+42.6$ & $+29.8$ & $+12.8$ & $-8.5$\\
\bottomrule
\end{tabular}}
\end{table}

\subsection{Removing the Switching Selector}
\label{app:no-switching}

Replacing the selector and the two conditional policies by a single monolithic
SAC policy over the same augmented action space is by far the most damaging
change. Time in range falls from $91.0\%$ to $61.0\%$, a loss of $30.0$
percentage points, and return falls from $-28.3$ to $-68.6$, while activations
\emph{rise} from $40.0$ to $43.4$. The monolithic controller regulates
substantially worse \emph{and} intervenes more often: it is dominated on both
axes simultaneously, which is the strongest form the ablation could have taken.
Mean glucose rises by $23.5$ mg/dL to $176.3$, right at the upper edge of the
target zone, and the configuration spends $38.96\%$ of the horizon above range
against \textsc{MINT}'s $8.96\%$.
The mechanism is visible in Figure~\ref{fig:app-trajectory}. On the frozen
evaluation episode, the monolithic policy leaves all $40$ activations unspent for
the first $6.25$ hours, during which glucose climbs to $255$ mg/dL, and only then
begins to intervene, whereas \textsc{MINT} acts from the first step and holds
glucose between $127$ and $179$ mg/dL for the entire $24$-hour horizon. This is
exactly the failure mode that Equation~\ref{eq:mode} is designed to prevent. A
monolithic continuous policy has no representation for the decision ``intervene
now rather than later''. It can only emit magnitudes, and abstention has to be
encoded as a near-zero magnitude that competes with the same Gaussian head that
also has to produce therapeutic doses. \textsc{MINT}'s selector makes the timing
decision an explicit comparison of $Q_0$, $Q_I$ and $Q_P$, and the comparison is
made at every step.
Two further readings support rather than qualify this. First, the monolithic
policy is not a weakened variant that we constructed for the ablation: it is
exactly the flat augmented SAC baseline of Table~\ref{tab:main-results}, and at
seed $0$ the ablation run and the baseline run are the same run. The ablation
and the headline baseline comparison are therefore the same evidence viewed
twice, which is why the $30$-point gap here and the $26.3$-point gap in the main
text are consistent. Second, the monolithic policy splits its budget across both
channels ($12.0$ immediate, $31.4$ persistent) whereas \textsc{MINT} commits
entirely to the persistent channel ($0.0$, $40.0$). Committing is the correct
behaviour in this domain, because basal insulin is the appropriate instrument
for $24$-hour regulation under a $40$-injection budget, and only a policy that
can compare modes is able to discover it.

\subsection{Removing the Persistent-State Exposure}
\label{app:no-state}

Hiding $z_t$ from the agent while leaving the recursion
$z_{t+1}=\rho z_t + B_P\eta^P_t$ running costs $2.22$ points of time in range
and $3.05$ of return. The effect is an order of magnitude smaller than removing
the selector, which is the expected ordering: with $\rho=0.9$ and a
one-dimensional residual state, twelve steps of insulin history are already in
the observation (Table~\ref{tab:app-envs}), so a sufficiently expressive critic
can partially reconstruct the residual load from them. The ablation is therefore
a test of whether \emph{explicit} sufficiency helps when an implicit proxy is
available, and it does: the loss is small but it is consistent in sign across
return, time in range, time above range and mean glucose, and the degradation is
present at the endpoint and in the last three checkpoints alike.
This is the empirical counterpart of Proposition~\ref{prop:markov}. The
proposition establishes that $(x_t,z_t,b_t)$ is a sufficient statistic for the
intervention history, and the ablation shows what happens when the learner is
denied the $z_t$ coordinate of that statistic and must recover it from a
finite observation window. Two consequences are worth noting. Time below range
stays at $0.00\%$ and every episode completes, so the cost of hiding $z_t$ is
paid in regulation quality rather than in safety, with the shield still holding
the floor. And the hidden-state configuration spends its budget almost exactly as
\textsc{MINT} does ($0.2$ immediate, $39.8$ persistent, $40.0$ total), so the
loss is not a resource-allocation failure but a dosing one: without observing
the accumulated residual, the policy misjudges how much additional persistent
intervention the current state can absorb, and Figure~\ref{fig:app-trajectory}
shows the resulting excursion to $191$ mg/dL where \textsc{MINT} peaks at $179$.

\subsection{Removing the Hard Budget Contract}
\label{app:no-budget}

The third ablation replaces the hard contract by a soft reward penalty
($\lambda=0.1$) and removes the remaining-budget coordinates from the
observation. It is the row a sceptical reader should look at first, because in
isolation its time in range is the highest in the table, $96.67\%$ against
\textsc{MINT}'s $91.04\%$. Read with the resource column, it is the strongest
single piece of evidence in this supplement for the design that the main text
argues for.
\paragraph{What the row actually says.} The soft-penalty agent consumes $82.6$
channel activations per episode. The contract every other configuration
respects, and that the clinical setting imposes, is $40$. That is a $106.5\%$
overrun, and the agent has no mechanism that could have prevented it: with the
budget absent from both the observation and the feasible-action set, there is
nothing for a policy to be careful about, and the penalty term merely makes
interventions slightly more expensive in a reward that is already dominated by
glycaemic error. Figure~\ref{fig:app-ablation-curves}(c) shows that consumption
never approaches $40$ at any point during training, since it starts at $126$
activations and settles around $82$. A soft penalty is not a weaker version of a
hard contract but a different object, and it does not deliver a resource
guarantee at any value of $\lambda$ we would be able to tune without seeing the
test panel.

\begin{figure}[!ht]
\centering
\includegraphics[width=\textwidth]{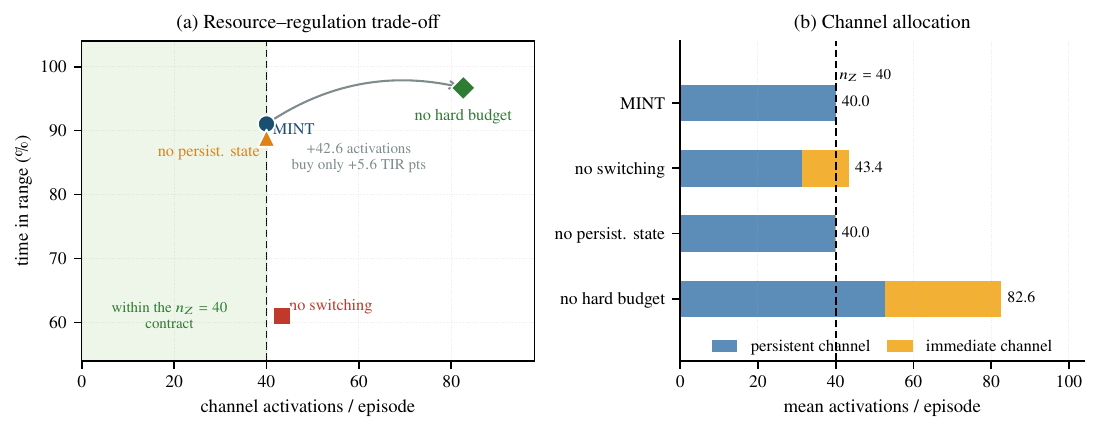}
\caption{(a) Resource--regulation trade-off across the ablations at the
$200$K endpoint. The shaded region is the $n_Z=40$ contract, and only the three
hard-contract configurations lie inside it. The arrow measures what the
soft-penalty configuration pays for its extra time in range: $42.6$ additional
activations for $5.63$ points, against \textsc{MINT}'s average of $0.44$
activations per point. (b) Channel allocation. \textsc{MINT} commits entirely to
the persistent channel, removing the selector splits the budget across both
channels, and removing the contract more than doubles total consumption.}
\label{fig:app-resource-frontier}
\end{figure}

\paragraph{What it costs \textsc{MINT} to honour the contract.} The comparison
is therefore not ``$96.67$ beats $91.04$'' but ``what does the last $5.63$ points
of time in range cost in resources?'' The answer is $42.6$ additional
activations, or $7.57$ activations per percentage point of time in range at the
margin, against \textsc{MINT}'s average of $0.44$ activations per point, which
is a marginal cost $17\times$ the average. Figure~\ref{fig:app-resource-frontier}(a)
plots the two points and the arrow between them: \textsc{MINT} sits at the knee
of the resource--regulation curve, and the soft-penalty agent sits far out on the
flat part of it, having more than doubled its resource use to buy a $6.2\%$
relative improvement in regulation. In efficiency terms \textsc{MINT} delivers
$2.28$ points of time in range per activation against $1.17$, so it is $1.94\times$
more resource-efficient. Return, the quantity the learner actually optimises,
is $-28.27$ against $-28.99$: \textsc{MINT} is \emph{ahead} by $0.72$, because
the reward charges for interventions and the unconstrained agent's extra $42.6$
of them are not free.

\paragraph{What the extra activations are spent on.} The soft-penalty agent
loses \textsc{MINT}'s channel specialisation:
Figure~\ref{fig:app-resource-frontier}(b) shows it firing the immediate channel
$29.8$ times and the persistent channel $52.8$ times, where \textsc{MINT} uses
the persistent channel exclusively. Its budget-shield activation rate is $0.000$
by definition, its glucose-shield rate falls to $0.415$ from
\textsc{MINT}'s $0.676$ at the same seed, and it delivers $5.18$ U of insulin
against \textsc{MINT}'s $4.42$ U, a $17\%$ larger dose. In other words, having lost the
resource contract, the agent reverts to spraying both channels and dosing more
heavily. That is a perfectly reasonable response to the objective it was given.
It is also unusable in the setting the benchmark models, and it is precisely why
the budget belongs in the state and the feasible-action set rather than in the
reward.

\paragraph{Relation to the main text.} This ablation isolates the constraint
mechanism from the learner family, which the constrained-RL baselines of
Table~\ref{tab:main-results} cannot do, since those change the learner as well.
Here the learner, the network, the optimiser and the seed are all identical to
\textsc{MINT}'s and only the budget mechanism moves. It therefore supplies the
controlled evidence for the claim made in the main text's discussion of those
baselines: because the budget enters \textsc{MINT}'s state and feasible-action
set rather than a penalty term, \textsc{MINT} never trades task reward against
constraint satisfaction, and when the same learner is given a penalty term
instead, it stops satisfying the constraint.

\subsection{Convergence Audit}
\label{app:convergence}

Because the ablations are single-seed, we ran an automated descriptive audit on
each to establish that the reported endpoint is a settled value rather than a
point sampled from an unconverged or collapsing trajectory.
Table~\ref{tab:app-convergence} reports it. The audit checks four things: that
all $200$K steps completed, that the last three checkpoints lie within a
tolerance set by the within-panel evaluation spread (\emph{stable late window}),
that the endpoint has not degraded materially from the best three-checkpoint
window, and that the frozen-panel return improved over the initial training
return.
All three ablations pass the completeness, late-stability and
no-late-degradation criteria. The late-window range is $5.75$ for the
no-switching run, $1.15$ for the hidden-state run and $1.07$ for the no-budget
run, in each case well inside the corresponding stability tolerance, and for
the two smaller ones inside a single point of return. One flag requires
explanation and we state it explicitly rather than suppressing it: the
composite \texttt{descriptive\_convergence\_pass} field is \texttt{false} for the
no-budget run. The cause is \emph{not} a late collapse. Its final three
checkpoint returns are $-28.07$, $-29.13$ and $-28.99$, a total late-window
spread of $1.07$, which is the flattest of the three ablations. The cause is the
opposite: without a budget to learn to respect, the configuration is already at
$-30.5$ return and $98.3\%$ time in range at the very first $20$K checkpoint, so
its total improvement over the initial training return, $0.43$, falls below the
automated learning-improvement threshold. The audit is measuring how much the run
learned, and this run had very little left to learn, which is itself
informative, since it suggests that under our protocol the difficulty of the
T1DM task lies largely in the resource contract rather than in the glycaemic
control problem. The two contract-respecting ablations, by contrast, improved by
$527.5$ and $521.3$ of return over their initial training performance before
settling.

\begin{table}[!ht]
\centering
\caption{Descriptive convergence audit for the three ablations, computed from
the ten frozen validation evaluations of each run. \emph{Late range} is the
spread of the last three checkpoint returns and \emph{tolerance} is the
stability threshold derived from the within-panel episode spread. The
\texttt{false} learning flag for the no-budget run reflects a high initial score
rather than a late collapse, as discussed in the text.}
\label{tab:app-convergence}
\small
\setlength{\tabcolsep}{5pt}
\begin{tabular}{@{}lcccccc@{}}
\toprule
Configuration & Late range & Tolerance & Stable late & No late & Improved over & Endpoint \\
& (last $3$ ckpts) & & window & degradation & initial training & return \\
\midrule
$-$ switching selector & $5.75$ & $50.18$ & \checkmark & \checkmark & \checkmark ($+521.3$) & $-68.64$\\
$-$ persistent-state exposure & $1.15$ & $11.18$ & \checkmark & \checkmark & \checkmark ($+527.5$) & $-31.33$\\
$-$ hard budget & $1.07$ & $9.91$ & \checkmark & \checkmark & $\times$ ($+0.43$) & $-28.99$\\
\bottomrule
\end{tabular}
\end{table}

\subsection{Summary of the Ablation}

The three removals separate cleanly. Removing the intervention selector costs
$30.0$ percentage points of time in range, which is the entire gap between
\textsc{MINT} and a monolithic policy with identical observation and action
access, and it degrades regulation while increasing resource use at the same
time. Exposing the persistent state accounts for a
further $2.2$ points and is what makes the sufficient-statistic property
available to the learner rather than merely true in principle. The hard
resource contract, far from being the source of \textsc{MINT}'s performance,
\emph{costs} it $5.6$ points of time in range relative to an otherwise identical
agent that ignores the contract entirely, and buys, in exchange, a $51.6\%$
reduction in resource consumption, a guarantee that holds almost surely rather
than in expectation, and a higher return. Taken together the three rows say that
\textsc{MINT}'s advantage comes from structured intervention selection under a
resource contract, which is the claim the paper makes.

\section{Additional Cross-Domain Results}
\label{app:additional}

This section reports evidence at a granularity the main text could not
accommodate. All of it derives from the same $105$ formal runs ($7$ methods
$\times$ $3$ benchmarks $\times$ $5$ seeds) and the two sweeps.

\subsection{Per-Seed Endpoints}
\label{app:perseed}

\begin{figure}[!ht]
\centering
\includegraphics[width=\textwidth]{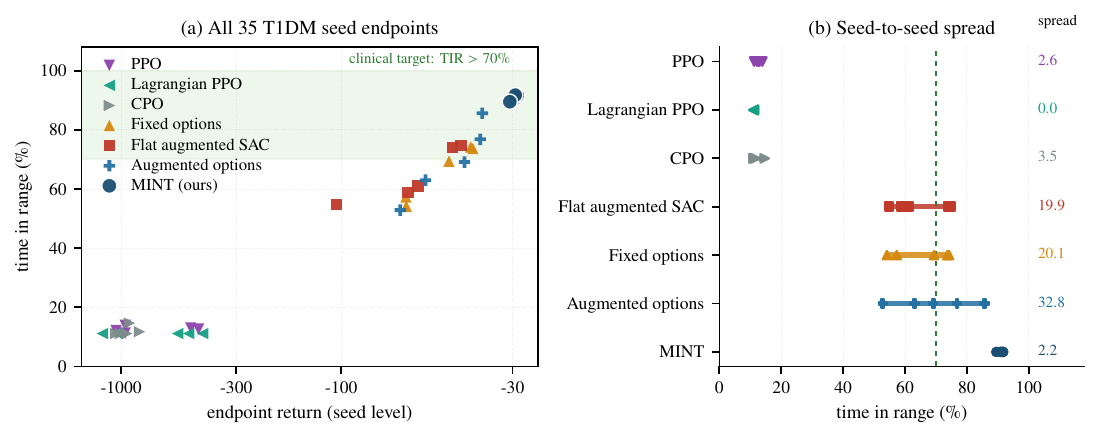}
\caption{(a) All $35$ T1DM seed endpoints. \textsc{MINT}'s five runs occupy the
top-right corner with no overlap with any baseline run, the weakest of them
sitting $3.9$ points of time in range above the strongest baseline run.
(b) Seed-to-seed spread in time in range, with
the numeric spread at the right of each row. Among methods that regulate at all,
\textsc{MINT}'s spread is $8.9$--$14.7$ times tighter. Lagrangian PPO's spread of
$0.00$ is degenerate, because all five seeds sit at the $11.11\%$ reset floor.}
\label{fig:app-seed-scatter}
\end{figure}

Tables~\ref{tab:app-perseed-t1dm}--\ref{tab:app-perseed-mujoco} give every
seed-level endpoint behind Table~\ref{tab:main-results}, so that a reader can
verify the aggregates and inspect the dispersion directly rather than through a
standard deviation. Figure~\ref{fig:app-seed-scatter} plots the T1DM endpoints.
The T1DM table supports a stronger statement than the aggregate does.
\textsc{MINT}'s \emph{worst} seed attains $89.51\%$ time in range, while the
\emph{best} seed of any of the six baselines attains $85.62\%$ (augmented
options, seed $4$). Every one of \textsc{MINT}'s five runs therefore beats every
one of the thirty baseline runs on the primary clinical metric, with no overlap
at all. The same holds for return, where \textsc{MINT}'s worst seed is $-31.13$
and the best baseline seed is $-42.38$. Separation of this kind cannot be read
off a mean and a standard deviation, and it is the reason we report the
per-seed table.
Dispersion is the second story. Among the methods that regulate at all, meaning
those reaching above $50\%$ time in range, \textsc{MINT}'s seed-to-seed spread is
$2.23$ percentage points, against $19.86$ for the flat augmented policy,
$20.07$ for fixed options and $32.77$ for augmented options, so between $8.9$ and
$14.7$ times tighter. The three on-policy methods show smaller spreads still,
but at the wrong end of the scale: Lagrangian PPO records a spread of exactly
$0.00$ because all five of its seeds sit at $11.11\%$, the value obtained by
keeping the $32$ steps that are already in range at reset and never bringing
glucose back. A degenerate policy is perfectly reproducible, which is not the
same property. Reproducibility is worth reporting because a controller that
attains $91\%$ on one seed and $55\%$ on another is not a controller one could
deploy, whatever its mean.

\begin{table}[!ht]
\centering
\caption{T1DM per-seed endpoints, reported as return\,/\,time in range (\%) over
the full $288$-step horizon. Every \textsc{MINT} seed exceeds every baseline seed
on both quantities. Time below range is $0.00\%$ in every cell of this table.}
\label{tab:app-perseed-t1dm}
\small
\setlength{\tabcolsep}{4pt}
\resizebox{\textwidth}{!}{%
\begin{tabular}{@{}lccccccc@{}}
\toprule
Method & Seed $0$ & Seed $1$ & Seed $2$ & Seed $3$ & Seed $4$ & Mean return & Mean TIR (\%) \\
\midrule
PPO & $-958.68$ / $11.25$ & $-1053.62$ / $12.15$ & $-482.59$ / $12.99$ & $-445.08$ / $12.64$ & $-957.52$ / $13.89$ & $-779.50\pm291.09$ & $12.58\pm0.98$\\
Flat augmented SAC & $-68.64$ / $61.04$ & $-72.72$ / $58.75$ & $-51.15$ / $74.65$ & $-54.44$ / $74.03$ & $-104.85$ / $54.79$ & $-70.36\pm21.33$ & $64.65\pm9.12$\\
Fixed options & $-73.45$ / $54.17$ & $-73.60$ / $57.29$ & $-56.02$ / $69.38$ & $-46.24$ / $73.68$ & $-47.14$ / $74.24$ & $-59.29\pm13.54$ & $65.75\pm9.40$\\
Augmented options & $-65.60$ / $62.99$ & $-49.64$ / $69.10$ & $-43.24$ / $76.81$ & $-75.89$ / $52.85$ & $-42.38$ / $85.62$ & $-55.35\pm14.79$ & $69.47\pm12.58$\\
Lagrangian PPO & $-1036.59$ / $11.11$ & $-553.16$ / $11.11$ & $-1215.53$ / $11.11$ & $-493.30$ / $11.11$ & $-425.72$ / $11.11$ & $-744.86\pm356.55$ & $11.11\pm0.00$\\
CPO & $-945.36$ / $11.11$ & $-1055.95$ / $11.11$ & $-1008.81$ / $11.11$ & $-825.05$ / $11.74$ & $-917.22$ / $14.65$ & $-950.48\pm88.59$ & $11.94\pm1.54$\\
\midrule
\textsc{MINT} (ours) & $\mathbf{-28.27}$ / $\mathbf{91.04}$ & $\mathbf{-28.90}$ / $\mathbf{90.76}$ & $\mathbf{-28.09}$ / $\mathbf{91.67}$ & $\mathbf{-28.93}$ / $\mathbf{91.74}$ & $\mathbf{-31.13}$ / $\mathbf{89.51}$ & $\mathbf{-29.06\pm1.21}$ & $\mathbf{90.94\pm0.90}$\\
\bottomrule
\end{tabular}}
\end{table}

\begin{table}[!ht]
\centering
\caption{Inventory per-seed endpoint return ($n_Z=60$, $100$ periods), with
aggregate service level and channel activations. \textsc{MINT} and the flat
augmented policy both meet $100.0\%$ of demand, and \textsc{MINT} does so with
$24.7\%$ fewer activations. Fixed options' much lower return is a genuine
stockout failure at $62.6\%$ service, not a scoring artefact.}
\label{tab:app-perseed-inventory}
\small
\setlength{\tabcolsep}{5pt}
\resizebox{\textwidth}{!}{%
\begin{tabular}{@{}lcccccccc@{}}
\toprule
Method & Seed $0$ & Seed $1$ & Seed $2$ & Seed $3$ & Seed $4$ & Mean return & Service (\%) & Interv. \\
\midrule
PPO & $-234.6$ & $-176.7$ & $-553.8$ & $-253.4$ & $-252.1$ & $-294.1\pm148.5$ & $98.6\pm3.0$ & $68.22\pm5.89$\\
Flat augmented SAC & $-279.2$ & $-282.4$ & $-263.7$ & $-216.7$ & $-226.5$ & $-253.7\pm30.3$ & $\mathbf{100.0\pm0.0}$ & $51.62\pm4.29$\\
Fixed options & $-1279.5$ & $-1640.9$ & $-1555.8$ & $-1670.1$ & $-1379.8$ & $-1505.2\pm169.4$ & $62.6\pm5.5$ & $67.90\pm4.28$\\
Augmented options & $-511.3$ & $-340.5$ & $-333.6$ & $-334.6$ & $-371.9$ & $-378.4\pm75.9$ & $99.3\pm1.1$ & $68.60\pm4.72$\\
Lagrangian PPO & $-377.6$ & $-2081.5$ & $-2883.3$ & $-2395.9$ & $-1631.4$ & $-1873.9\pm952.9$ & $61.5\pm37.1$ & $40.28\pm24.67$\\
CPO & $-1472.7$ & $-1235.5$ & $-1191.0$ & $-1453.9$ & $-1260.1$ & $-1322.6\pm131.0$ & $64.1\pm1.5$ & $75.00\pm0.00$\\
\midrule
\textsc{MINT} (ours) & $-256.0$ & $-199.9$ & $-196.0$ & $-217.2$ & $-224.1$ & $\mathbf{-218.7\pm23.9}$ & $\mathbf{100.0\pm0.0}$ & $\mathbf{38.86\pm2.40}$\\
\bottomrule
\end{tabular}}
\end{table}

\begin{table}[!ht]
\centering
\caption{HalfCheetah per-seed endpoint return ($\rho=0.9$, unbudgeted), with
the reward decomposed into the native task reward and the control cost paid.
\textsc{MINT}'s four healthy seeds earn a higher base task reward than any
baseline while paying an order of magnitude less control cost, and seed $4$ is the
collapsed run analysed in Section~\ref{app:collapse} and is retained in every
aggregate. Lagrangian PPO reduces exactly to PPO here because HalfCheetah is
unbudgeted under our protocol, which is why the two rows are identical.}
\label{tab:app-perseed-mujoco}
\small
\setlength{\tabcolsep}{5pt}
\resizebox{\textwidth}{!}{%
\begin{tabular}{@{}lcccccccc@{}}
\toprule
Method & Seed $0$ & Seed $1$ & Seed $2$ & Seed $3$ & Seed $4$ & Mean return & Base reward & Control cost \\
\midrule
PPO & $1416.9$ & $1377.1$ & $966.7$ & $1398.3$ & $1325.6$ & $1296.9\pm187.7$ & $1317.5\pm186.3$ & $20.59\pm3.50$\\
Flat augmented SAC & $1813.2$ & $1548.0$ & $1666.2$ & $1506.7$ & $1844.2$ & $\mathbf{1675.7\pm151.9}$ & $1716.4\pm153.6$ & $40.77\pm1.82$\\
Fixed options & $118.8$ & $-116.2$ & $668.3$ & $-148.1$ & $-284.0$ & $47.7\pm376.0$ & $85.5\pm377.7$ & $37.73\pm1.73$\\
Augmented options & $-117.8$ & $-400.4$ & $230.0$ & $-149.3$ & $124.7$ & $-62.5\pm247.7$ & $-22.8\pm247.4$ & $39.72\pm4.38$\\
Lagrangian PPO & $1416.9$ & $1377.1$ & $966.7$ & $1398.3$ & $1325.6$ & $1296.9\pm187.7$ & $1317.5\pm186.3$ & $20.59\pm3.50$\\
CPO & $315.0$ & $12.5$ & $537.6$ & $-117.9$ & $104.3$ & $170.3\pm258.9$ & $173.8\pm261.2$ & $3.56\pm2.41$\\
\midrule
\textsc{MINT} (ours) & $2150.0$ & $2146.3$ & $2158.6$ & $2196.5$ & $-398.9$ & $1650.5\pm1145.8$ & $1657.9\pm1133.3$ & $\mathbf{7.44\pm12.53}$\\
\textsc{MINT}, healthy seeds only & $2150.0$ & $2146.3$ & $2158.6$ & $2196.5$ & n/a & $\mathbf{2162.8\pm23.1}$ & $\mathbf{2164.7\pm23.2}$ & $\mathbf{1.84\pm0.03}$\\
\bottomrule
\end{tabular}}
\end{table}

The HalfCheetah decomposition in Table~\ref{tab:app-perseed-mujoco} is worth
reading alongside the main text, where the two methods are reported as
indistinguishable in raw return. That parity is real and we do not walk it back:
retaining all five seeds, the paired difference is $-25.2$ with a $95\%$
interval of $[-1573.3, 1522.9]$. But the decomposition shows the parity is a
sum of two large and opposite effects. On its four healthy seeds \textsc{MINT}
earns $2164.7$ of native task reward against the flat policy's $1716.4$, a
$26.1\%$ improvement, while paying $1.84$ of control cost against $40.77$, or
$22.2$ times less. The flat policy achieves its return by actuating both
channels at all $1000$ steps and absorbing the resulting cost, whereas
\textsc{MINT} achieves a higher one by actuating a single channel per step and
paying almost nothing for it. Aggregated with the collapsed seed the two
numbers cancel. Disaggregated, they are the efficiency claim of the paper
expressed in the reward function itself.

\subsection{Channel Allocation and Mode Usage}
\label{app:channels}

Table~\ref{tab:app-channels} gives the resource accounting behind the
intervention columns of Table~\ref{tab:main-results}. Two distinctions carry the
result and are easy to conflate. \emph{Channel activations} count non-null
channel outputs and are what the tables report as intervention use.
\emph{Intervention steps} count distinct time steps at which the controller
acted at all. For a monolithic policy that emits both channels jointly these
differ by a factor of up to two, because a single decision consumes two
activations, whereas for \textsc{MINT}, whose selector commits to one mode per
step, they coincide.
The consequence is visible in every row. On HalfCheetah the flat policy spends
$2000$ activations across $1000$ decision times while \textsc{MINT} spends
$1000$ across the same $1000$, so decision coverage is identical at half the
resource cost. On Inventory \textsc{MINT} uses $38.86$ activations at $38.86$
decision times against $51.62$ at $40.26$, so it acts at $96.5\%$ as many time
steps for $75.3\%$ of the resource. On T1DM the flat policy needs $43.80$
activations to act at $40.00$ time steps, whereas \textsc{MINT} needs exactly
$40.00$: the flat policy pays a $9.5\%$ resource premium purely for its
inability to abstain on one channel while acting on the other.

\begin{table}[!ht]
\centering
\caption{Channel-allocation accounting, mean $\pm$ SD over $5$ seeds.
\emph{Activations} counts non-null channel outputs and \emph{decision times}
counts distinct steps at which the controller acted. They coincide for
\textsc{MINT} because its selector commits to one mode per step, and diverge for
a monolithic policy that emits both channels jointly. \emph{Persist.\ share} is
the fraction of activations placed on the persistent channel. The HalfCheetah
\textsc{MINT} row is dominated by the single collapsed seed, and the
healthy-seed value is $0.8\pm0.1\%$.}
\label{tab:app-channels}
\small
\setlength{\tabcolsep}{5pt}
\resizebox{\textwidth}{!}{%
\begin{tabular}{@{}llcccccc@{}}
\toprule
Benchmark & Method & Activations & Decision times & Immediate & Persistent & Persist.\ share (\%) & Mean $\|z_t\|$\\
\midrule
HalfCheetah ($\rho=0.9$) & Flat augmented SAC & $2000.00\pm0.00$ & $1000.00\pm0.00$ & $1000.0\pm0.0$ & $1000.0\pm0.0$ & $50.0\pm0.0$ & $2.643\pm0.371$\\
 & \textsc{MINT} & $999.98\pm0.04$ & $999.98\pm0.04$ & $794.0\pm442.7$ & $206.0\pm442.7$ & $20.6\pm44.3$ & $2.522\pm5.544$\\
\addlinespace[2pt]
Inventory ($n_Z=60$) & Flat augmented SAC & $51.62\pm4.29$ & $40.26\pm3.54$ & $15.0\pm0.0$ & $36.6\pm4.3$ & $70.8\pm2.4$ & $265.9\pm31.3$\\
 & \textsc{MINT} & $38.86\pm2.40$ & $38.86\pm2.40$ & $15.0\pm0.0$ & $23.9\pm2.4$ & $61.3\pm2.4$ & $193.4\pm12.7$\\
\addlinespace[2pt]
T1DM ($n_Z=40$) & Flat augmented SAC & $43.80\pm0.79$ & $40.00\pm0.00$ & $12.0\pm0.0$ & $31.8\pm0.8$ & $72.6\pm0.5$ & $0.369\pm0.025$\\
 & \textsc{MINT} & $40.00\pm0.00$ & $40.00\pm0.00$ & $0.0\pm0.0$ & $40.0\pm0.0$ & $100.0\pm0.0$ & $0.507\pm0.011$\\
\bottomrule
\end{tabular}}
\end{table}

The persistent-channel share column shows that the selector reaches a different
allocation in every domain, and in each case the allocation matches the domain's
physics. On T1DM it commits entirely to basal insulin, the instrument suited to
$24$-hour regulation under an injection budget. On Inventory it places $61.3\%$
of activations on standard replenishment and saturates the $15$-unit emergency
allowance, using the pipeline channel for baseline supply and the immediate
channel only for shortfalls. On HalfCheetah, where the persistent actuator
duplicates a capability the immediate actuator already provides at lower cost,
the healthy seeds place under $1\%$ of activations on it, so the selector learns
that a channel exists and should not be used. A monolithic policy over a joint
continuous action space has no comparable freedom. On HalfCheetah it is pinned
at exactly $50\%$ by construction, and on the two budgeted benchmarks it spends
its full immediate allowance in every seed ($15.0\pm0.0$ and $12.0\pm0.0$), so
its split is fixed by the sub-budget rather than selected. That is the
structural asymmetry Equation~\ref{eq:mode} removes.

\subsection{Shield-Activation Decomposition}
\label{app:shield-decomp}

\begin{figure}[!ht]
\centering
\includegraphics[width=\textwidth]{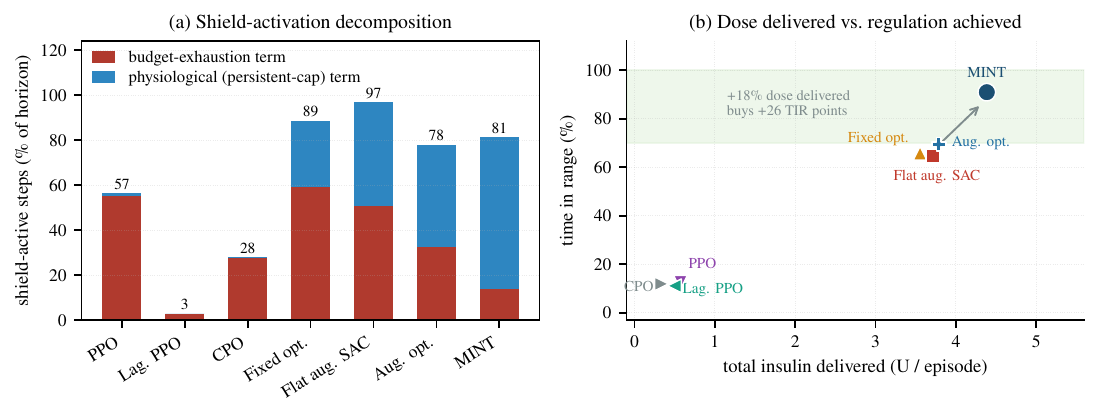}
\caption{(a) Shield activations on T1DM decomposed into the exact
budget-exhaustion term and the $K$-step physiological term. \textsc{MINT} has
the lowest budget-exhaustion rate of any method that regulates: its selector
abstains without needing the resource layer to abstain for it. (b) Insulin
delivered against time in range. \textsc{MINT} converts an $18\%$ larger dose
into $26$ additional points of time in range relative to the flat augmented
policy, while the two constrained-RL baselines fail by under-dosing at under a
tenth of that dose.}
\label{fig:app-shield}
\end{figure}

Since the shield is identical for every method, the composition of its
activations is a clean read-out of the policy underneath it.
Table~\ref{tab:app-shield-decomp} and Figure~\ref{fig:app-shield} decompose it
on T1DM.
The pattern is sharp. \textsc{MINT}'s budget-shield rate is $0.137$, so on fewer
than $14\%$ of steps does the resource layer have to intervene to prevent an
infeasible proposal. The flat augmented policy's is $0.506$ and fixed options'
is $0.591$, meaning that these policies propose infeasible interventions on half
of all steps and rely on the shield to abstain on their behalf. \textsc{MINT}'s
selector, by contrast, has learned when not to act, so the resource layer is
mostly idle. This is the same fact as the activation counts in
Table~\ref{tab:app-channels}, seen from the environment's side rather than the
agent's, and it is the sense in which \textsc{MINT} satisfies the contract
endogenously rather than being clamped into satisfying it.
\textsc{MINT}'s total shield rate is nonetheless the highest in the
table, at $0.812$, and the reason is instructive rather than awkward. Its
activations are overwhelmingly the \emph{physiological} term ($0.675$ of the
horizon), which is the suppression of a persistent-channel proposal that would
have carried the residual state past the $\bar z = 0.8$ cap. \textsc{MINT} keeps the persistent
channel loaded almost continuously, at $\|z_t\|=0.507$ against the flat
policy's $0.369$, with Figure~\ref{fig:app-trajectory}(b) showing $z_t$ riding
just below the cap for nineteen hours, so it is the method that interacts
with the cap most often. Operating a persistent actuator at the boundary of
its physiological envelope, from the first step of the episode, is how it
reaches $91\%$ time in range at $0\%$ time below range. The cap is a limit
\textsc{MINT} saturates by design, not a rescue it requires.
The dose column completes the picture. \textsc{MINT} delivers $4.38$ U per
episode against the flat policy's $3.71$ U, an $18.1\%$ larger dose, using
$8.7\%$ fewer activations, which is the same resource concentrated into
better-timed and larger persistent interventions. At the other end, the two
constrained-RL baselines deliver $0.50$ and $0.34$ U, under a tenth of
\textsc{MINT}'s dose, which is exactly why their time below range is $0.0\%$ and
their time in range is at the floor. They fail by under-dosing, and the scalar
cost term is what discourages the interventions the task requires.

\begin{table}[!ht]
\centering
\caption{Shield-activation decomposition and dose accounting on T1DM
($n_Z=40$), mean $\pm$ SD over $5$ seeds. Rates are fractions of the $288$-step
horizon. The shield is enabled identically for every method, so differences
reflect the policy proposing into it. A low \emph{budget} rate means the policy
abstained on its own, whereas a high \emph{glucose} rate means the policy is
operating the persistent channel near its physiological envelope.}
\label{tab:app-shield-decomp}
\small
\setlength{\tabcolsep}{5pt}
\begin{tabular}{@{}lccccccc@{}}
\toprule
Method & Budget rate & Glucose rate & Total rate & Insulin (U) & Mean BG & Completion & Failure \\
\midrule
PPO & $0.553\pm0.281$ & $0.013\pm0.016$ & $0.565\pm0.272$ & $0.57\pm0.43$ & $295.3\pm28.4$ & $0.943\pm0.055$ & $0.400$\\
Flat augmented SAC & $0.506\pm0.057$ & $0.463\pm0.064$ & $0.748\pm0.075$ & $3.71\pm0.26$ & $171.7\pm11.9$ & $1.000\pm0.000$ & $0.000$\\
Fixed options & $0.591\pm0.118$ & $0.296\pm0.090$ & $0.711\pm0.094$ & $3.55\pm0.24$ & $171.0\pm7.7$ & $1.000\pm0.000$ & $0.000$\\
Augmented options & $0.327\pm0.082$ & $0.454\pm0.109$ & $0.701\pm0.082$ & $3.79\pm0.35$ & $168.4\pm8.7$ & $1.000\pm0.000$ & $0.000$\\
Lagrangian PPO & $0.029\pm0.039$ & $0.000\pm0.000$ & $0.029\pm0.039$ & $0.50\pm0.40$ & $297.3\pm32.0$ & $0.957\pm0.061$ & $0.320$\\
CPO & $0.279\pm0.328$ & $0.001\pm0.002$ & $0.280\pm0.328$ & $0.34\pm0.29$ & $314.2\pm15.5$ & $0.919\pm0.068$ & $0.440$\\
\midrule
\textsc{MINT} (ours) & $\mathbf{0.137\pm0.043}$ & $0.675\pm0.025$ & $0.812\pm0.053$ & $4.38\pm0.10$ & $\mathbf{153.5\pm0.7}$ & $1.000\pm0.000$ & $\mathbf{0.000}$\\
\bottomrule
\end{tabular}
\end{table}

\subsection{Decision-Time Accounting under a Binding Budget}
\label{app:decision-times}

\begin{figure}[!ht]
\centering
\includegraphics[width=\textwidth]{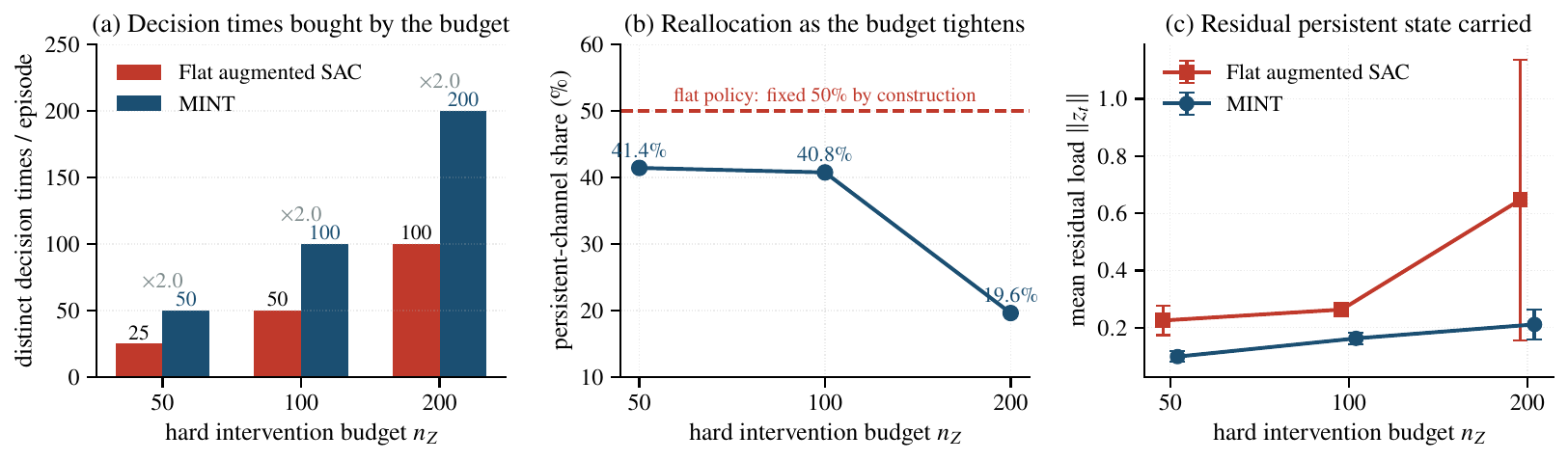}
\caption{Budget sweep mechanism on HalfCheetah ($\rho=0.9$). (a) Distinct
decision times purchased by the same activation budget: exactly $2.0\times$ for
\textsc{MINT} at every binding budget, with zero seed variance, because the
factor is structural. (b) \textsc{MINT} reallocates onto the persistent channel
as interventions are rationed, an adaptation the flat policy cannot express.
(c) \textsc{MINT} carries less accumulated residual load at every budget.
Bands and whiskers are two-sided Student-$t$ $95\%$ intervals over $5$ seeds.}
\label{fig:app-decision-times}
\end{figure}

The budget sweep of the main text reports return. Table~\ref{tab:app-budget}
and Figure~\ref{fig:app-decision-times} report the resource accounting that
explains it, which is where the mechanism lives.
At every binding budget both methods exhaust exactly $n_Z$ activations with zero
violations, so the resource contract is equally satisfied. What differs is what
the budget buys. The flat policy fires both channels at every decision, so
$n_Z$ activations purchase $n_Z/2$ decision times: $25$, $50$ and $100$ at
$n_Z=50$, $100$ and $200$. \textsc{MINT}'s selector commits to one channel per
decision, so the same $n_Z$ activations purchase $n_Z$ decision times: $50$,
$100$ and $200$. The factor is exactly $2.0$ at every budget, with zero variance
across seeds, because it is structural rather than learned. Turning the same
resource allowance into twice as many opportunities to act is what separating
timing from magnitude buys, and it is the direct explanation for the return
ordering in the main text: $14.8$ against $7.8$, $34.3$ against $19.1$, and
$196.5$ against $29.4$.
The reallocation column adds a second effect. As the budget tightens
\textsc{MINT} shifts weight onto the persistent channel, placing $41.4\%$ and
$40.8\%$ of activations there at $n_Z=50$ and $100$ against $19.6\%$ at
$n_Z=200$, which is what Equation~\ref{eq:mode} predicts: when interventions are
rationed, the mode whose effect outlives the decision has higher value per
activation. The flat policy cannot express this at all, since its split is
pinned at $50\%$ in every column. And the residual-load column shows
\textsc{MINT} carrying consistently less accumulated persistent state than the
flat policy at every budget ($0.099$ against $0.225$, $0.163$ against $0.263$,
$0.211$ against $0.647$), that is, achieving more with less residual actuation.
Two caveats from the main text carry over unchanged. The return gap is not
ordered by scarcity, being largest at $n_Z=200$, the loosest binding budget,
and no individual return contrast survives Benjamini--Hochberg correction
(Table~\ref{tab:app-budget-paired}, smallest $q=0.238$). What carries the
result is the consistent sign across all three binding budgets together with
the decision-time mechanism, which has no sampling error at all.

\begin{table}[!ht]
\centering
\caption{Budget sweep on HalfCheetah ($\rho=0.9$), full accounting, $5$ seeds
each. \emph{Dec.\ times} is the number of distinct steps at which the controller
acted and \emph{Act.} is channel activations. Both methods exhaust exactly $n_Z$
activations with zero violations at every binding budget, but \textsc{MINT}
converts them into exactly twice as many decision times. \emph{Shield} is the
budget-shield rate, i.e.\ the fraction of the $1000$-step episode spent
resource-exhausted.}
\label{tab:app-budget}
\small
\setlength{\tabcolsep}{5pt}
\begin{tabular}{@{}llccccccc@{}}
\toprule
$n_Z$ & Method & Return & Dec.\ times & Act. & Immediate & Persistent & Persist.\ (\%) & Shield / $\|z_t\|$\\
\midrule
$50$ & Flat augmented SAC & $7.8\pm8.1$ & $25$ & $50$ & $25.0$ & $25.0$ & $50.0$ & $0.975$ / $0.225$\\
 & \textsc{MINT} & $\mathbf{14.8\pm6.0}$ & $\mathbf{50}$ & $50$ & $29.3$ & $20.7$ & $41.4$ & $0.950$ / $\mathbf{0.099}$\\
\addlinespace[2pt]
$100$ & Flat augmented SAC & $19.1\pm16.8$ & $50$ & $100$ & $50.0$ & $50.0$ & $50.0$ & $0.950$ / $0.263$\\
 & \textsc{MINT} & $\mathbf{34.3\pm20.7}$ & $\mathbf{100}$ & $100$ & $59.2$ & $40.8$ & $40.8$ & $0.900$ / $\mathbf{0.163}$\\
\addlinespace[2pt]
$200$ & Flat augmented SAC & $29.4\pm88.6$ & $100$ & $200$ & $100.0$ & $100.0$ & $50.0$ & $0.900$ / $0.647$\\
 & \textsc{MINT} & $\mathbf{196.5\pm82.6}$ & $\mathbf{200}$ & $200$ & $160.8$ & $39.2$ & $19.6$ & $0.800$ / $\mathbf{0.211}$\\
\addlinespace[2pt]
$\infty$ & Flat augmented SAC & $1675.7\pm151.9$ & $1000$ & $2000$ & $1000.0$ & $1000.0$ & $50.0$ & $0.000$ / $2.643$\\
 & \textsc{MINT} & $1650.5\pm1145.8$ & $1000$ & $\mathbf{1000}$ & $794.0$ & $206.0$ & $20.6$ & $0.000$ / $2.522$\\
\bottomrule
\end{tabular}
\end{table}

\begin{table}[!ht]
\centering
\caption{Paired contrasts for the budget sweep, \textsc{MINT} minus flat
augmented SAC, pairing by training seed ($n=5$). Intervals are two-sided
Student-$t$ $95\%$ intervals and $q$ is the Benjamini--Hochberg adjusted
$p$-value within the
sweep. No individual contrast survives correction, which is why the main text
rests the claim on the consistent sign across the three binding budgets and on
the decision-time mechanism of Table~\ref{tab:app-budget} rather than on any one
contrast.}
\label{tab:app-budget-paired}
\small
\begin{tabular}{@{}lccccc@{}}
\toprule
$n_Z$ & Paired difference & $95\%$ CI & $p$ & $q$ (BH) & Cohen's $d_z$ \\
\midrule
$50$ & $7.0$ & $[-9.9, 23.9]$ & $0.314$ & $0.419$ & $0.51$\\
$100$ & $15.2$ & $[-17.1, 47.5]$ & $0.262$ & $0.419$ & $0.58$\\
$200$ & $167.1$ & $[-10.7, 345.0]$ & $0.059$ & $0.238$ & $1.17$\\
$\infty$ & $-25.2$ & $[-1573.3, 1522.9]$ & $0.966$ & $0.966$ & $-0.02$\\
\bottomrule
\end{tabular}
\end{table}

\subsection{Persistence Sweep: Full Numeric Detail}
\label{app:persistence}

\begin{figure}[!ht]
\centering
\includegraphics[width=\textwidth]{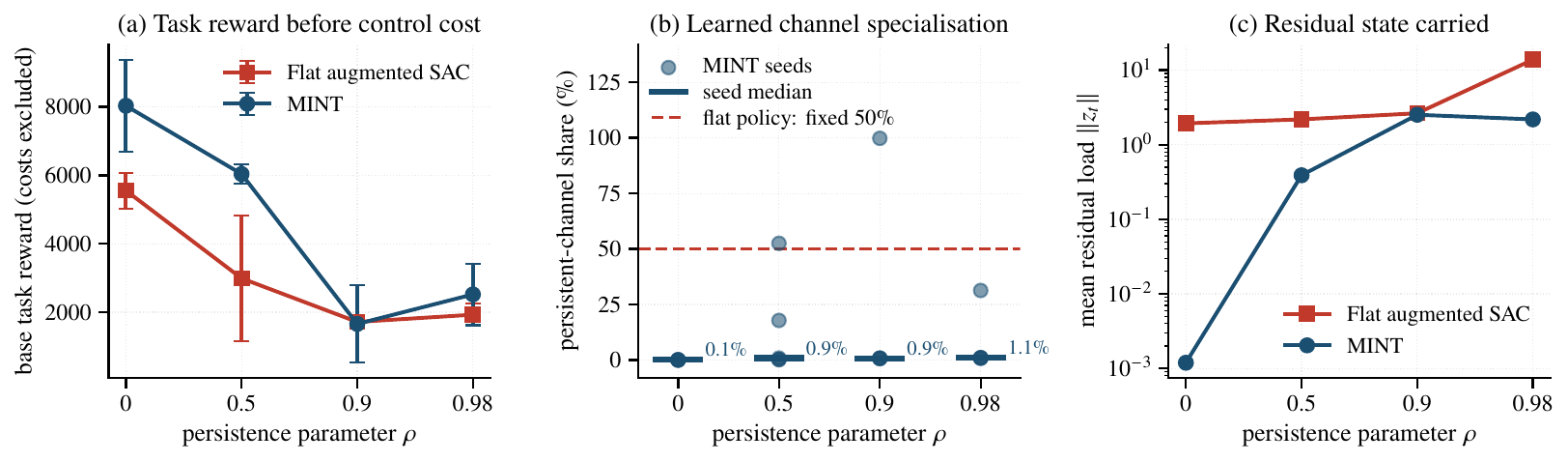}
\caption{Persistence sweep, quantities not shown in the main text. (a) Native
task reward before intervention costs are charged. (b) Per-seed
persistent-channel share with the seed median: whichever channel a
\textsc{MINT} seed selects it specialises, while the flat policy is pinned at
$50\%$ by construction. (c) Accumulated residual load on a log axis:
\textsc{MINT} abstains almost entirely from the persistent channel at $\rho=0$
where it offers nothing, and carries over six times less residual load than the flat
policy at $\rho=0.98$.}
\label{fig:app-persistence}
\end{figure}

Table~\ref{tab:app-persistence} gives the complete numbers behind the
persistence figure of the main text, together with quantities that figure does
not show, and Figure~\ref{fig:app-persistence} plots them.
The return column reproduces the main text's finding and its honest reading: the
raw-return advantage is \emph{non-monotonic} in $\rho$, peaking at $\rho=0.5$
($+3084.2$) and vanishing at $\rho=0.9$ ($-25.2$), and no individual contrast
reaches $q<0.05$ after correction (Table~\ref{tab:app-persistence-paired}). We
do not claim the return advantage grows with persistence. What does hold at
every $\rho$ tested is the resource ratio: \textsc{MINT} reaches its returns on
$1000$ activations against the flat policy's $2000$, in every column, so its
return per activation exceeds the flat policy's by $5.27$, $4.62$, $0.81$ and
$1.57$ at $\rho=0$, $0.5$, $0.9$ and $0.98$ respectively, three of the four
surviving correction.
The two columns the main text does not show are the more mechanistic ones. The
median persistent-channel share is at or below $1.1\%$ at every $\rho$, so on
HalfCheetah \textsc{MINT} typically discovers that the persistent actuator
duplicates the immediate one at higher cost and declines to use it, with
individual seeds at $\rho=0.5$, $0.9$ and $0.98$ committing to the persistent
channel instead. Whichever channel a seed selects, it specialises, whereas the
flat policy is fixed at $50\%$ throughout. And the residual-load column shows the
selector responding to $\rho$ directly: at $\rho=0$, where the persistent
channel degenerates into a second immediate actuator with no memory,
\textsc{MINT} carries $\|z_t\|=0.001$, three orders of magnitude below the flat
policy's $1.934$, having learned to abstain from a channel that offers
nothing. As $\rho$ grows its residual load rises to $0.389$ and $2.522$, then
falls back to $2.185$ at $\rho=0.98$, where the flat policy's rises to
$13.808$, over six times higher. Explicit mode selection is what makes it
possible to decline a channel, whereas a monolithic policy must always fire it
and pay for it.

\begin{table}[!ht]
\centering
\caption{Persistence sweep on HalfCheetah, full detail, $5$ seeds per cell.
\emph{Base reward} excludes intervention costs. \emph{Persist.\ share} is the
seed \emph{median} rather than the mean, since at $\rho\in\{0.5,0.9,0.98\}$ a
single seed commits to the persistent channel and dominates the mean, and the
per-seed values are shown in Figure~\ref{fig:app-persistence}(b).
\textsc{MINT} uses exactly half as many activations as the flat policy in every
row.}
\label{tab:app-persistence}
\small
\setlength{\tabcolsep}{6pt}
\begin{tabular}{@{}llccccc@{}}
\toprule
$\rho$ & Method & Return & Base reward & Activations & Persist.\ share (\%) & Mean $\|z_t\|$\\
\midrule
$0.00$ & Flat augmented SAC & $5505.6\pm519.3$ & $5553.4$ & $2000$ & $50.0$ & $1.934$\\
 & \textsc{MINT} & $\mathbf{8027.3\pm1345.1}$ & $\mathbf{8030.7}$ & $\mathbf{1000}$ & $0.1$ & $\mathbf{0.001}$\\
\addlinespace[2pt]
$0.50$ & Flat augmented SAC & $2944.5\pm1837.1$ & $2989.6$ & $2000$ & $50.0$ & $2.187$\\
 & \textsc{MINT} & $\mathbf{6028.7\pm296.1}$ & $\mathbf{6036.7}$ & $\mathbf{1000}$ & $0.9$ & $\mathbf{0.389}$\\
\addlinespace[2pt]
$0.90$ & Flat augmented SAC & $\mathbf{1675.7\pm151.9}$ & $1716.4$ & $2000$ & $50.0$ & $2.643$\\
 & \textsc{MINT} & $1650.5\pm1145.8$ & $1657.9$ & $\mathbf{1000}$ & $0.9$ & $\mathbf{2.522}$\\
\addlinespace[2pt]
$0.98$ & Flat augmented SAC & $1886.5\pm311.2$ & $1927.7$ & $2000$ & $50.0$ & $13.808$\\
 & \textsc{MINT} & $\mathbf{2516.0\pm895.3}$ & $\mathbf{2520.0}$ & $\mathbf{1000}$ & $1.1$ & $\mathbf{2.185}$\\
\bottomrule
\end{tabular}
\end{table}

\begin{table}[!ht]
\centering
\caption{Paired contrasts for the persistence sweep, \textsc{MINT} minus flat
augmented SAC, pairing by training seed ($n=5$), with Benjamini--Hochberg
correction within the sweep. The advantage is large and positive at $\rho=0$ and
$\rho=0.5$ but does not survive correction, and it is absent at $\rho=0.9$. The
main text states this plainly and claims no monotone trend in $\rho$.}
\label{tab:app-persistence-paired}
\small
\begin{tabular}{@{}lccccc@{}}
\toprule
$\rho$ & Paired difference & $95\%$ CI & $p$ & $q$ (BH) & Cohen's $d_z$ \\
\midrule
$0.00$ & $2521.7$ & $[438.6, 4604.9]$ & $0.028$ & $0.057$ & $1.50$\\
$0.50$ & $3084.2$ & $[532.7, 5635.7]$ & $0.028$ & $0.057$ & $1.50$\\
$0.90$ & $-25.2$ & $[-1573.3, 1522.9]$ & $0.966$ & $0.966$ & $-0.02$\\
$0.98$ & $629.5$ & $[-157.3, 1416.3]$ & $0.090$ & $0.121$ & $0.99$\\
\bottomrule
\end{tabular}
\end{table}

\subsection{The One Collapsed HalfCheetah Seed}
\label{app:collapse}

\begin{figure}[!ht]
\centering
\includegraphics[width=\textwidth]{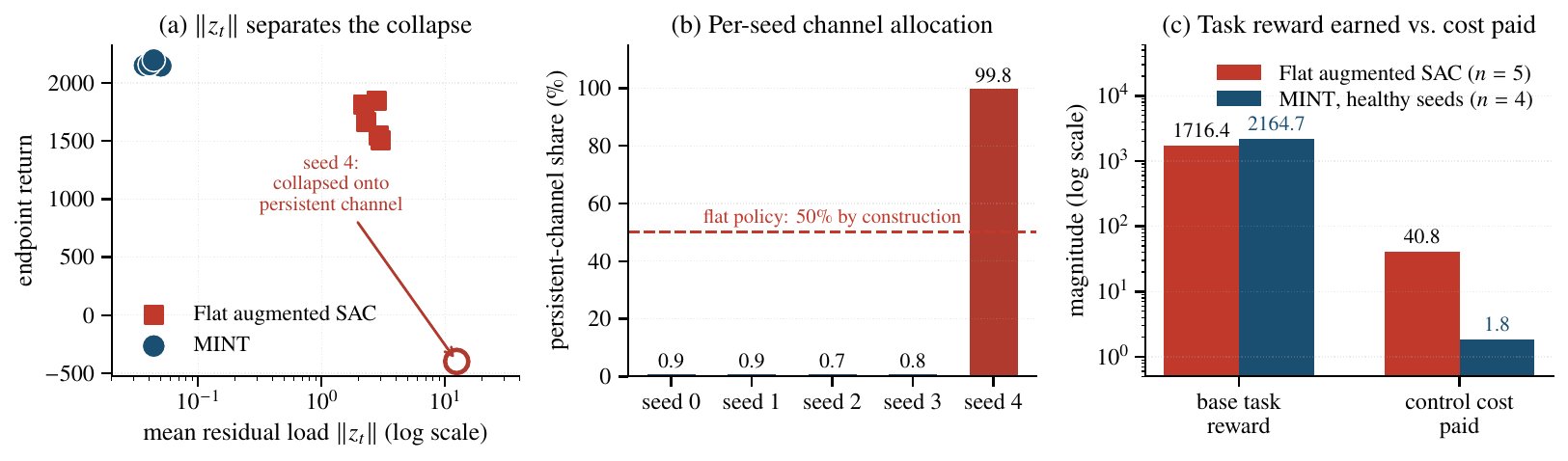}
\caption{Diagnosis of the single collapsed \textsc{MINT} HalfCheetah seed
($\rho=0.9$, unbudgeted). (a) Endpoint return against accumulated residual load
on a log axis: the collapsed run (hollow marker) is separated from the four
healthy runs by more than two orders of magnitude in $\|z_t\|$, a quantity
observable online. (b) Per-seed persistent-channel share: seed $4$ places
$99.8\%$ of its activations on the persistent channel where the healthy seeds
place under $1\%$. (c) Native task reward earned against control cost paid, on a
log axis: on its healthy seeds \textsc{MINT} earns $26.1\%$ more task reward
than the flat augmented policy while paying $22.2\times$ less control cost.}
\label{fig:app-collapse}
\end{figure}

We retain every seed in every aggregate, which means one \textsc{MINT}
HalfCheetah run at $\rho=0.9$ contributes $-398.9$ to a column whose other four
entries lie between $2146.3$ and $2196.5$, and inflates the reported standard
deviation from $23.1$ to $1145.8$. Reporting it is the correct choice and we do
not exclude it anywhere. Diagnosing it is also possible, and the diagnosis is
specific rather than a generic instability.
Figure~\ref{fig:app-collapse} shows that the failure is a mode-selection
collapse and that the augmented state detects it exactly. The four healthy seeds
place $0.7$--$0.9\%$ of their activations on the persistent channel and carry
$\|z_t\|\approx0.04$, whereas seed $4$ places $99.8\%$ on the persistent channel
and carries $\|z_t\|=12.44$, over $300$ times more, and pays $29.85$ of
persistent control cost against the healthy seeds' $0.21$. The selector has
converged to the wrong mode and then saturated the residual state, at which
point the immediate channel it needs for locomotion is effectively unavailable.
Three points follow. First, $\|z_t\|$ separates the collapsed run from the
healthy runs by two orders of magnitude on a quantity that is available online
and requires no reward information, so this failure mode is cheaply detectable
in deployment. Second, it is a failure of the selector's exploration in a domain
with no resource contract and no safety shield, since HalfCheetah is unbudgeted
under our protocol, and it does not occur on either of the two budgeted
benchmarks, where all five seeds behave consistently
(Tables~\ref{tab:app-perseed-t1dm} and~\ref{tab:app-perseed-inventory}). A hard
budget both constrains and informs mode selection, so the setting in which the
selector is least supervised is the setting in which it occasionally fails.
Third, the discrete entropy temperature $\alpha_{\text{switch}}$ is the natural
control for exactly this failure and is fixed at $0.1$ on HalfCheetah by our
frozen tuning protocol. We did not retune it after observing the collapse,
because doing so would have meant selecting a hyperparameter on the reported
panel. We flag it as the obvious first thing to change and leave it as a stated
limitation rather than a repaired result.

\clearpage
\subsection{Closed-Loop Trajectory Anatomy}
\label{app:trajectory}

\begin{figure}[!ht]
\centering
\includegraphics[width=\textwidth]{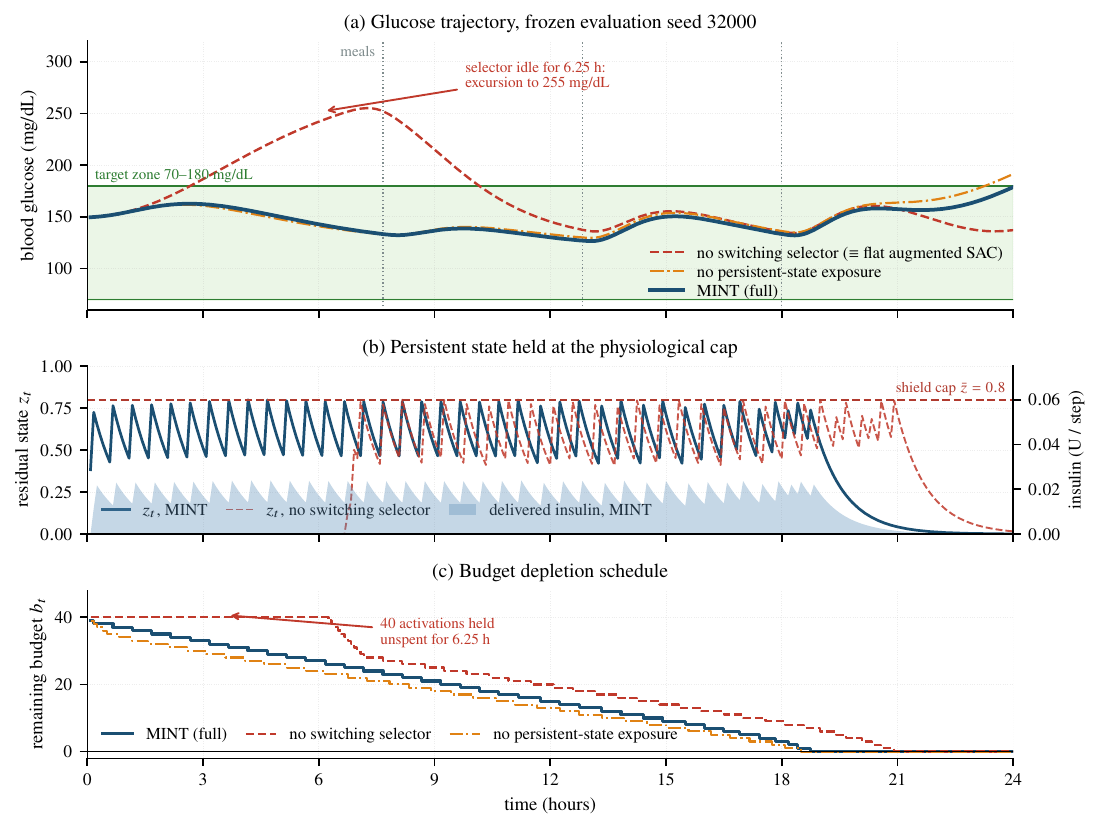}
\caption{Step-level anatomy of one frozen T1DM evaluation episode (seed
$32000$, patient \texttt{adolescent\#001}, AGVP scenario, $288$ steps at $5$
minutes). Dotted verticals mark meals. The no-switching ablation is the flat
augmented SAC baseline, so at this seed the two are the same run and are drawn
once. (a) Glucose against the $70$--$180$ mg/dL target zone. (b) \textsc{MINT}'s
residual persistent state held just below the $\bar z=0.8$ physiological cap,
with delivered insulin on the right axis. (c) Remaining budget $b_t$: all three
configurations respect the $n_Z=40$ contract exactly, and differ only in when
they spend it.}
\label{fig:app-trajectory}
\end{figure}

Figure~\ref{fig:app-trajectory} shows one frozen $24$-hour evaluation episode
(seed $32000$) step by step, for full \textsc{MINT} and for two of the
ablations. It makes concrete what the aggregate metrics compress, and every
panel is a quantity recorded by the environment rather than a reconstruction.
Panel (a) is the control outcome. \textsc{MINT} holds glucose between $127$ and
$179$ mg/dL for the entire horizon, giving $100\%$ time in range on this
episode, with the trajectory never approaching either boundary. The monolithic
policy leaves its budget untouched for the first $6.25$ hours, lets glucose
climb to $255$ mg/dL, and recovers only after the first meal has passed, giving
$69.1\%$ time in range on the same episode from the same initial state. The
hidden-state configuration acts from the first step like \textsc{MINT} but
overshoots to $191$ mg/dL, at $96.9\%$ in range, so it has the timing right and
the dose slightly wrong, which is exactly the signature one expects when the
accumulated residual is unobserved.
Panel (b) shows the mechanism on \textsc{MINT}'s side. The residual state $z_t$
is driven up to the $\bar z = 0.8$ physiological cap within the first hour and
then held in a narrow band just below it for nineteen hours, each activation
topping up a decaying reservoir rather than delivering an isolated dose. This is
persistent-effect control used as intended, with overlapping interventions
superposing in $z_t$ and the controller still deciding at every one of the
$288$ steps. The monolithic policy's $z_t$ trace stays at zero for the first
$6.25$ hours and then saturates the cap in a much noisier band.
Panel (c) shows the resource schedule and explains panel (a). \textsc{MINT}
spends its $40$ activations at a near-uniform rate from step $1$, exhausting
the budget at hour $18.8$, whereas the monolithic policy holds all $40$ unspent
until hour $6.25$ and exhausts them at hour $20.9$. Both respect the contract
exactly and neither violates it. The difference is entirely in \emph{when} the
budget is spent, which is the quantity \textsc{MINT}'s selector optimises and a
monolithic magnitude policy has no representation for. On this episode the
outcome turns at least as much on intervention timing as on intervention size,
which is the mechanism the aggregate metrics compress.

\subsection{What This Evidence Does and Does Not Establish}

Collected in one place, the supplementary evidence supports four claims. The
intervention selector is the load-bearing component of \textsc{MINT}
(Section~\ref{app:no-switching}). The advantage takes the form of a better
performance-per-resource trade-off rather than a uniformly larger return, and
that trade-off holds in every regime tested, at every persistence value, at
every binding budget and on all three domains
(Sections~\ref{app:channels},~\ref{app:decision-times},~\ref{app:persistence}).
The hard resource contract is close to free, costing $5.6$ points of time in
range for a $51.6\%$ reduction in resource use and a guarantee that holds almost
surely (Section~\ref{app:no-budget}). And the method is markedly more
reproducible across seeds than any baseline that regulates at all
(Section~\ref{app:perseed}).
It does not establish four other things, and we state them explicitly. The
ablations are single-seed and descriptive. The evaluation panel is shared with
hyperparameter development rather than held out. The T1DM results use one
virtual patient and one meal scenario in simulation, and establish no clinical
safety or efficacy whatsoever. And the selector can fail: one HalfCheetah seed
collapses onto the wrong mode, in the one setting where no resource contract
supervises the choice (Section~\ref{app:collapse}).

\clearpage
\section{Algorithms}
\label{app:algorithms}

 \begin{algorithm}[ht]
\begin{algorithmic}[1] 
\STATE Define: 
\\$\bm{\hat{y}}_t:=(b^1_t,b^2_t,\bm{y}_t)$,
\\$b^1_t:= X(\bmy_t)-\frac{1}{2}M-\sum_{k=0}^t\eta^I_k$, {\fontfamily{qcr}\textcolor{blue}{//State component for immediate-acting admissible range}} 
\\$b^2_t:=H(|X_t-M|-l)-N_0$ \textcolor{blue}{//State component for BGL admissible range}
                     \\$\bm{\hat{y}'}_{t+1}= \tilde{P}(\cdot|\eta^I_t,\bm{\hat{y}'}_t)$
		\STATE {\bfseries Input:} Stepsize $\alpha$, batch size $B$, episodes $K$, steps per episode $T$, mini-epochs $e$, immediate-acting intervention  cost parameter $\beta$, $\Delta$ budget violation parameter.
		\STATE {\bfseries Initialise:} Policy network (intervention acting) $\boldsymbol{\pi}$, 
  Critic network (acting )$V_{\boldsymbol{\pi}}$. 
		\STATE Given reward objective function, $\boldsymbol{\hat{r}}_t$, initialise Rollout Buffers $\mathcal{B}_{\pi}$.
		\FOR{$N_{episodes}$}
		    \STATE Reset state $s_0$, Reset Rollout Buffers $\mathcal{B}_{\pi}$, $\mathcal{B}_{\mathfrak{g}}$,
		    \FOR{$t=0,1,\ldots$}
    		    \STATE Sample  $\eta^I_t \sim \pi(\cdot|\bm{\hat{y}}_t)$.

  \IF{$\bm{r}_{t+1}\sim\cR(\bm{\hat{y}'}_{t+1},\eta^I_{t+1},\bmb_{t+1})>-\Delta$}              
                \STATE Apply the action so $\bm{\hat{y}}_{t+1}\sim P(\cdot|\eta^I_t,\bm{\hat{y}}_t),$
        	\STATE Receive rewards $\bm{\hat{r}}_t\sim \cR(\bm{\hat{y}}_t,\eta^I_t,\bmb_t)$.
            \STATE Store 
            $(\bm{\hat{y}}_t,\eta^I_t, \bm{\hat{y}}_{t+1}, \bm{\hat{r}}_t)$ in $\mathcal{B}_{\bm\pi}$ 
                    \ELSE
    		         \STATE Apply the null action so $\bm{\hat{y}}_{t+1}\sim P(\cdot|\bm0,\bm{\hat{y}}_t),$
        	\STATE Receive rewards $\boldsymbol{\hat{r}}_t\sim \cR(\bm{\hat{y}}_t,0,\bmb_t)$.
            \STATE Store 
            $(\bm{\hat{y}}_t,\bm0, \bm{\hat{y}}_{t+1}, \bm{\hat{r}}_t)$ in $\mathcal{B}_{\bm\pi}$ 
                \ENDIF
    	    	\ENDFOR
    	\STATE{\textbf{// Learn the individual policy}}
        \STATE Update policy  $\boldsymbol{\pi}$ and critic $V_{\boldsymbol{\pi}}$ networks using $\mathfrak{B}_{\boldsymbol{\pi}}$ 
        \ENDFOR
	\caption{Our Method (Actor-Critic version)}
	\label{algo:our_method_no_budget_version_sans_carbs} 
\end{algorithmic}        \clearpage 
\end{algorithm}

 \begin{algorithm}[ht]
\begin{algorithmic}[1] 
\STATE Define: \\$\bm{\hat{y}}_t:=(b^1_t,b^2_t,b^3_t,b^4_t,\bm{y}_t)$, \\$b^1_t:=n_Z -\sum_{l=0}^t\sum_{j,k\geq 1}\left(\boldsymbol{1}_{\{l={\tau_k}\}} +  \boldsymbol{1}_{\{l={\rho_j}\}}\right)$  {\fontfamily{cmss} \textcolor{blue}{//State component for injections budget}}
\\$b^2_t:= X(\bmy_t)-\frac{1}{2}M-\sum_{k=0}^tE_k\eta^P_k$, {\fontfamily{cmss}\textcolor{blue}{//State component for long-acting admissible range}} 
\\$b^3_t:= X(\bmy_t)-\frac{1}{2}M-\sum_{k=0}^t\eta^I_k$, {\fontfamily{cmss}\textcolor{blue}{//State component for immediate-acting admissible range}}
\\$b^4_t:=H(|X_t-M|-l)-N_0$ \textcolor{blue}{//State component for process admissible range} 
                    
$ \bm{\hat{y}''}_{t+1}= \tilde{P}(\cdot|\eta^P_t,\bm{\hat{y}''}_{t})$
                    \\$\bm{\hat{y}'}_{t+1}= \tilde{P}(\cdot|\eta^I_t,\bm{\hat{y}'}_{t})$
		\STATE {\bfseries Input:} Stepsize $\alpha$, batch size $B$, episodes $K$, steps per episode $T$, mini-epochs $e$, immediate-acting intervention  cost parameter $\beta$, long-acting cost parameter $\alpha$, Spectral decay distribution $F$, $\Delta$ budget violation parameter.
		\STATE {\bfseries Initialise:} Policy network (intervention acting) $\boldsymbol{\pi}$, Policy network (switching) $\mathfrak{g}$,
  Critic network (acting )$V_{\boldsymbol{\pi}}$, Critic network (switching )$V_{\mathfrak{g}}$, $\forall t< 0$ set termination probability $E_t\equiv 0$. 
		\STATE Given reward objective function, $\boldsymbol{\hat{r}}_t$, initialise Rollout Buffers $\mathcal{B}_{\pi}$, $\mathcal{B}_{\mathfrak{g}}$ (use Replay Buffer for SAC).
		\FOR{$N_{episodes}$}
		    \STATE Reset state $s_0$, Reset Rollout Buffers $\mathcal{B}_{\pi}$, $\mathcal{B}_{\mathfrak{g}}$,
		    \FOR{$t=0,1,\ldots$}
    		    \STATE Sample  $(\eta^P_t,\eta^I_t)\equiv \bm{\eta}_t \sim \boldsymbol{\pi}(\cdot|\bm{\hat{y}}_t)$,  $(g^L_t,g^F_t)\equiv \bmg_t \sim \mathfrak{g}(\cdot|\bm{\hat{y}}_t)$,  $e_t\sim F(Z)$.
    	\IF{$e_tg^L_{t-1}>0$ i.e.,  $e_tg^L_{t-1}\in (0,1]$ and  $\cR(\bm{\hat{y}''}_{t+1},\eta^P_{t+1},\bmb_{t+1}),\dots, \cR(\bm{\hat{y}''}_{t+K},\eta^P_{t+K},\bmb_{t+K})>-\Delta$}
         \STATE Set $g^L_t\equiv g^L_{t-1}$. Apply $(z\eta^P)_t$ so $\bm{\hat{y}}_{t+1}\sim P(\cdot|((z\eta^P)_t,0),\bm{\hat{y}}_t)$
                    
        		    \STATE Receive rewards $\boldsymbol{\hat{r}}_{S,t} =\boldsymbol{\hat{r}}_t$ and $\boldsymbol{\hat{r}}_t\sim \cR(\bm{\hat{y}}_t,\eta^P_t,\bmb_t)$. 
                    \STATE Store 
                  $(\bm{\hat{y}}_t,((z\eta^P)_t,0), \bm{\hat{y}}_{t+1}, \bm{\hat{r}}_t)$ and $(\bm{\hat{y}}_t,\textcolor{black}{((z\eta^P)_t,0)},(g^L_t=1,0), \bm{\hat{y}}_{t+1}, \bm{\hat{r}}_{S,t})$ in $\mathcal{B}_{\bm\pi}$ and $\mathcal{B}_{\mathfrak{g}}$ respectively.
        \ELSE 
            \IF{$g^L_t=1$ and $\cR(\bm{\hat{y}''}_{t+1},\eta^P_{t+1},\bmb_{t+1}),\dots, \cR(\bm{\hat{y}''}_{t+K},\eta^P_{t+K},\bmb_{t+K})>-\Delta$} 
         \STATE Apply $\eta^P_{t}$ so $\bm{\hat{y}}_{t+1}\sim P(\cdot|(\eta^P_t,0),\bm{\hat{y}}_t),$
                    
        		    \STATE Receive rewards $\boldsymbol{\hat{r}}_{S,t} = -\alpha+\boldsymbol{\hat{r}}_t$ and $\boldsymbol{\hat{r}}_t\sim \cR(\bm{\hat{y}}_t,\eta^P_t,\bmb_t)$. 
                    \STATE Store $(\bm{\hat{y}}_t,(\eta^P_t,0), \bm{\hat{y}}_{t+1}, \bm{\hat{r}}_t)$ and $(\bm{\hat{y}}_t, \textcolor{black}{(\eta^P_t,0)},(g^L_t=1,0), \bm{\hat{y}}_{t+1}, \boldsymbol{\hat{r}}_{S,t})$ in $\mathcal{B}_{\bm\pi}$ and $\mathcal{B}_{\mathfrak{g}}$ respectively.
    		    \ELSE
    		     \IF{$g^F_t=1$ and $\cR(\bm{\hat{y}'}_{t+1},\eta^I_{t+1},\bmb_{t+1})>-\Delta$} 
         \STATE Apply $\eta^I_{t}$ so $\bm{\hat{y}}_{t+1}\sim P(\cdot|(0,\eta^I_t,0),\bm{\hat{y}}_t),$
                    
        		    \STATE Receive rewards $\boldsymbol{\hat{r}}_{S,t} = -\beta\cdot (\eta^I_{t})^2+\boldsymbol{\hat{r}}_t$ and $\boldsymbol{\hat{r}}_t\sim \cR(\bm{\hat{y}}_t,\eta^I_t,\bmb_t)$. 
                    \STATE Store $(\bm{\hat{y}}_t,(0,\eta^I_t), \bm{\hat{y}}_{t+1}, \bm{\hat{r}}_t)$ and $(\bm{\hat{y}}_t,\textcolor{black}{(0,\eta^I_t)},(0,g^F_t=1), \bm{\hat{y}}_{t+1}, \bm{\hat{r}}_{S,t})$ in $\mathcal{B}_{\bm\pi}$ and $\mathcal{B}_{\mathfrak{g}}$ respectively.
    		    \ELSE
    		         \STATE Apply the null action so $\bm{\hat{y}}_{t+1}\sim P(\cdot|\bm0,\bm{\hat{y}}_t),$
        	\STATE Receive rewards $\boldsymbol{\hat{r}}_{S,t} = \boldsymbol{r}_t\sim \cR(\bm{\hat{y}}_t,0,\bmb_t)$ and $\boldsymbol{\hat{r}}_t$.
            \STATE Store 
            $(\bm{\hat{y}}_t,\bm0, \bm{\hat{y}}_{t+1}, \bm{\hat{r}}_t)$ and $(\bm{\hat{y}}_t,\textcolor{black}{\bma_t\equiv\bm0},\bmg_t\equiv\bm0, \bm{\hat{y}}_{t+1}, \bm{\hat{r}}_{S,t})$ in $\mathcal{B}_{\bm\pi}$ 
                   and $\mathcal{B}_{\mathfrak{g}}$ respectively.
                \ENDIF
                   \ENDIF
    		    \ENDIF

        	\ENDFOR
    	\STATE{\textbf{// Learn the individual policies}}
        \STATE Update policy  $\boldsymbol{\pi}$ and critic $V_{\boldsymbol{\pi}}$ networks using $\mathfrak{B}_{\boldsymbol{\pi}}$ 
        \STATE Update policy  $\mathfrak{g}$ and critic $V_{\mathfrak{g}}$ networks using $\mathfrak{B}_{\mathfrak{g}}$
        \ENDFOR
	\caption{Our Method (Actor-Critic Budget version)}
	\label{algo:our_method_budget_version_sans_carbs} 
\end{algorithmic}         
\end{algorithm}
\clearpage

\section{Spectral Decay}\label{app:decay}\label{sec:decay}
To model the decay of the effect of the long-acting intervention over time, the control variate $\eta^P_t$ is scaled by a value $E_t$ where $E_t\sim  \max( F(\cE),1-\eta^P_{t-1})$, which also ensures that $Z\equiv 1$ at activation times $\tau_1,\tau_3,\ldots,$ (since $\eta^P_t\equiv 0$ on the interval $\tau_{2k}\leq t<\tau_{2k+1}$ for any $k=0,1,\ldots,$) and $E_t\in \cE$ elsewhere. The distribution 
$F$ is chosen such that $\operatorname{Supp}(F)(E_t)=\{z\in \cE|z\leq  E_{t-1}\}$.  Intuitively, this leads to stochastic decay of the magnitude of the long-acting intervention (measured by the value $\eta^P e $) since it is performed at times $\tau_1,\tau_3,\ldots,$ at which points $\eta^P e\sim 1\cdot\max( F(\cE),1)\equiv 1$, and, 
we observe $\eta^P e\sim 1\cdot\max( F(\cE),0)=F(\cE)$ which approaches $0$ almost surely as $t \uparrow \tau_{2k}$ by construction of $F$ and $\cE$. Lastly, we can now define $\tau_{2k}:=\operatorname{inf}\{t> \tau_{2k-1}|E_t=0\}\in\mathcal{F}$.

\section{How Each Case Relates}\label{app:cases}
Case I is a degenerate case of Case II. To degenerate Case II into Case I the Switching policy for the long-acting treatment is fixed to $0$ for all states and the Switching policy for the fast-acting policy fixed to $1$ for all states and the set of allowed values for the fast-acting intervention is augmented to include $0$. The state is reduced to three dimensions $\boldsymbol{y}_t=[Z_t\;\mathcal{C}_t\;X_t
]^\top\in \cY\subset\mathbb{R}^3_{\geq 0}$.

\newpage

\normalsize
\newpage
\clearpage\Huge
\begin{center}\textbf{Additional Theoretical Results}
\end{center}
\normalsize
\medskip
\begin{proposition}\label{prop:switching_times}
Let $Q^{\pi}$ be the solution in Theorem \ref{theorem:existence}, for any $ \bmy_t\in\bm{\mathcal{Y}}$ the policy $\hat{\mathfrak{g}}$ is given by:\\ $\mathfrak{g}(\bmy_t)= \begin{cases}
 L,  \bm\cM_{\rm long}\bmQ^{\pi}\geq \bmQ^{\pi},
 \\
  F, \bm\cM_{\rm fast}\bmQ^{\pi}\geq \bmQ^{\pi}, \bm\cM_{\rm long}\bmQ^{\pi}<\bmQ^{\pi}
  \\0, \text{ otherwise}
\end{cases}$and the intervention times 
are\\ $\tau_k=\inf\{\tau>\tau_{k-1}|\bm\cM_{\rm fast}\bmQ^{\pi}= \bmQ^{\pi}\}$ and $\rho_j=\inf\{\rho>\rho_{j-1}|\bm\cM_{\rm long}\bmQ^{\pi}= \bmQ^{\pi}\}$. 
\end{proposition}
 Prop. \ref{prop:switching_times} characterises the optimal intervention conditions in which each policy should be executed. The condition can be evaluated online therefore allowing the $\mathfrak{g}$ policy to be computed online. A key aspect of Prop. \ref{prop:switching_times} is that it exploits the cost structure of the problem to determine when the agent should perform an intervention. 

\medskip
\noindent
The next theorem gives the tabular Q-learning instance of the framework, in
which the intervention policies and the switching policy are learned
concurrently. It is the multi-timescale counterpart of
Theorem~\ref{thm:qlearning} in the main text and is the statement proved later
in this supplement.
\begin{theorem}\label{theorem:q_learning}
Consider the following Q-learning variant:
\begin{align} \nonumber
\bmQ_{t+1}&(\bmy_t,\boldsymbol{\eta}_t,g)=\bmQ_{t}(\bmy_t,\boldsymbol{\eta}_t,g)
\\&\begin{aligned}+\alpha_t(\bmy_t,\boldsymbol{\eta}_t)\Bigg[\max\left\{\bm{\mathcal{M}}_i\bmQ_{t}(\bmy_t,\boldsymbol{\eta}_t,g), R(\bmy_t,\boldsymbol{\eta}_t,g)+\gamma\underset{\bm{\eta'}\in\bm{\mathcal{H}}}{\max}\;\bmQ_{t}(\bmy_{t+1},\bm{\eta'},g)\right\}&
\\-\bmQ_{t}(\bmy_t,\boldsymbol{\eta}_t,g)\Bigg]&,
\end{aligned}
\end{align}
then $\bmQ_{t}$ converges to $\bmQ^\star$ with probability $1$, where $\bmy_t,\bmy_{t+1}\in\bm{\cY}$ and $\bm\eta_t\in\bm{\mathcal{H}}$.
\end{theorem}
\clearpage\Huge
\begin{center}\textbf{Proof of Theoretical Results}
\end{center}
\normalsize
\section*{Assumptions \& Definitions}

The results of the paper are built under the following assumptions which are standard within RL and stochastic approximation methods.

\noindent \textbf{Assumption 1.}
The stochastic process governing the system dynamics is ergodic, that is  the process is stationary and every invariant random variable of $\{\bmy_t\}_{t\geq 0}$ is equal to a constant with probability $1$.

\noindent\textbf{Assumption 2.}
The function $R$ is in $L_2$.

\noindent\textbf{Assumption 3.}
For any positive scalar $c$, there exists a scalar $\kappa_c$ such that for all $\bmy\in\bm{\cY}$ and for any $t\in\mathbb{N}$ we have: $
    \mathbb{E}\left[1+\|\bmy_t\|^c|s_0=s\right]\leq \kappa_c(1+\|\bmy\|^c)$.

\noindent\textbf{Assumption 4.}
There exists scalars $C_1$ and $c_1$ such that for any function $v$ satisfying $|v(\bmy)|\leq C_2(1+\|s\|^{c_2})$ for some scalars $c_2$ and $C_2$ we have that: $
    \sum_{t=0}^\infty\left|\mathbb{E}\left[v(\bmy_t)|\bmy_0=\bmy\right]-\mathbb{E}[v(s_0)]\right|\leq C_1C_2(1+\|\bmy_0\|^{c_1c_2})$.

\noindent\textbf{Assumption 5.}
There exists scalars $c$ and $C$ such that for any $\bmy\in\bm{\cY}$ we have that $
    |R(\bmy,\cdot)|\leq C(1+\|\bmy\|^c)$.
In what follows, we denote by $\left( \mathcal{V},\|\|\right)$ any finite normed vector space.

Additionally, in keeping with the notion that activating a long-acting intervention is more costly than a fast-acting intervention activation we make the following assumption: 

\noindent\textbf{Assumption 6.}  The intervention costs are equal: $0<\beta= \alpha:=c$.







We begin the analysis with some preliminary results and definitions required for proving our main results.

\begin{definition}{A.1}
Given a normed space $\|\cdot\|_X, X$, an operator $T: X\to X$ is a contraction if there exists some constant $c\in[0,1[$ for which for any measurable functions $J_1,J_2\in  X$ the following bound holds: $    \|TJ_1-TJ_2\|\leq c\|J_1-J_2\|$.
\end{definition}

\begin{definition}{A.2}
An operator $T: X\to  X$ is non-expansive if $\forall J_1,J_2\in  X$ the following bound holds: $    \|TJ_1-TJ_2\|\leq \|J_1-J_2\|$.
\end{definition}

Since the following result is well-known, we state it without proof.
\begin{lemma}
\label{max_lemma}
For any
$f: \cY\to\mathbb{R}: X\to\mathbb{R}$, we have that the following inequality holds:
\begin{align}
\left\|\underset{a\in X}{\max}\:f(a)-\underset{a\in X}{\max}\: g(a)\right\| \leq \underset{a\in X}{\max}\: \left\|f(a)-g(a)\right\|.
\end{align}
\end{lemma}
\begin{lemma}\citep{tsitsiklis1999optimal}\label{non_expansive_P}
The probability transition kernel $P$ is non-expansive so that if $\forall J_1,J_2\in  X$ the following holds: $    \|PJ_1-PJ_2\|\leq \|J_1-J_2\|$.
\end{lemma}
\begin{lemma}\label{lemma:max_min}
The following inequality holds:
\[
\left| \max_a \max_b f(a,b) - \max_a \min_b g(a,b) \right| \leq \max_a \max_b |f(a,b) - g(a,b)|.
\]
\end{lemma}
\begin{proof}
Define by
$
M := \max_a \max_b f(a,b), \;
N := \max_a \min_b g(a,b), \;
D := \max_a \max_b |f(a,b) - g(a,b)|$. Our goal is to show that \( |M - N| \leq D \). First we establish an upper bound on \( M - N \):
Let \( a^\ast, b^\ast \) be such that $
f(a^\ast, b^\ast) = \max_a \max_b f(a,b) = M$. Then for any \( a \), we have $
\min_b g(a,b) \leq g(a, b^\ast) \quad \Rightarrow \quad \max_a \min_b g(a,b) \leq \max_a g(a, b^\ast) \leq g(a^\ast, b^\ast)$, Therefore, $
N \leq g(a^\ast, b^\ast) \quad \Rightarrow \quad M - N \leq f(a^\ast, b^\ast) - g(a^\ast, b^\ast) \leq |f(a^\ast, b^\ast) - g(a^\ast, b^\ast)| \leq D$.

Next we seek to establish an upper bound on \( N - M \). Let \( a', b' \) be such that $
g(a', b') = \min_b g(a', b), \quad \text{and} \quad N = \max_a \min_b g(a,b) = \min_b g(a', b)$.
Then $
M = \max_a \max_b f(a,b) \geq f(a', b'),
$ hence $
N - M \leq g(a', b') - f(a', b') \leq |f(a', b') - g(a', b')| \leq D$, After combining both results we deduce $
|M - N| \leq D$
which is the desired result.
\end{proof}
\begin{lemma}\label{lemma:basic_inequality}
Suppose \( y \geq x \) and $
|y - z| \geq x - \max\{y, z\}$,  then the following inequality holds:
\[
y - x \leq |y - z|
.\]
\end{lemma}
\begin{proof}
To prove the lemma, we proceed by case analysis.

\textbf{Case 1: \( y \geq z \)}. In this case, \( \max\{y, z\} = y \), so the assumption becomes $|y - z| \geq x - y$. 
Since by assumption, \( y \geq x \) the inequality holds. Now, we observe that $
y - x \leq |y - z| \iff y \leq x + |y - z|$. 
But since \( |y - z| \geq x - y \), we can write $
|y - z| + y \geq x \Rightarrow y \leq x + |y - z|$, which is the desired result.

\textbf{Case 2: \( z > y \)}. In this case, \( \max\{y, z\} = z \), so the assumption becomes $
|y - z| \geq x - z$.
Note that \( |y - z| = z - y \), since \( z > y \). Therefore $
z - y \geq x - z \Rightarrow z + z \geq x + y \Rightarrow 2z \geq x + y$. Our aim is to prove $
y - x \leq |y - z|$ or $y \leq x + |y - z|$. 
Substituting \( |y - z| = z - y \) yields $
y \leq x + (z - y) \Rightarrow 2y \leq x + z,
$ then after subtracting $y+x$ from both sides we deduce that
\[
y-x \leq  z-y=|y-z|,\]
which is the required result.
\end{proof}
\clearpage
\Large
\begin{center}\textbf{Main Proofs}
\end{center}
\normalsize
We now state and prove the main results of the paper. Our first proposition proves the convergence of the switcher agent's policy for a given pair of intervention policies. In what follows, we denote by $\bm\cM_i\bm{Q}\in \{\bm\cM_{\rm long} Q_1,\bm\cM_{\rm fast} Q_2\} $ for any pair of measurable functions $Q_1$ and $Q_2$ and for any $\bmy,\bmy'\in\bm{\mathcal{Y}}$, we write $\cP^{\eta^i}_{\bmy\bmy'}:=\sum_{\bmy'\in\bm{\cY}}P(\bmy';\eta^i,\bmy)$ and $\mathcal{P}^{\pi^i}_{\bmy\bmy'}=:\sum_{\eta^i\in\mathcal{H}^i}\pi^i(\eta^i|\bmy)\mathcal{P}^{\eta^i}_{\bmy\bmy'}$ where $i\in\{L,F\}$.

\begin{proposition}\label{theorem:existence}
Let $v:\bm{\cY}\to\mathbb{R}$ then for any fixed joint policy $\boldsymbol{\pi}=(\pi^L,\pi^F)\in \boldsymbol{\Pi}$, the solution of Switching agent's problem is given by 
\begin{align}
\underset{k\to\infty}{\lim}T^kv(\bmy|\boldsymbol{\pi},\mathfrak{g})=\underset{\hat{\mathfrak{g}}}{\max}\;v(\bmy|\boldsymbol{\pi},\hat{\mathfrak{g}})=v^\ast(\bmy|\bm\pi),\quad \forall \bmy\in\bm{\cY},
\end{align} 
where 
$
T v(\bmy|\boldsymbol{\pi},\mathfrak{g}):=\max\Bigg[\max\left\{\bm\cM_{\rm long}^{\pi^L}Q^{\pi^L}_1, R(\bmy,\bm0)+\gamma\sum_{\bmy'\in\bm{\mathcal{Y}}}P(\bmy';\bm0,\bmy)v(\bmy')\right\}
,\bm\cM_{\rm fast}^{\pi^F}Q^{\pi^F}_2\Bigg]$ given $\bm\pi\equiv (\pi^L,\pi^F)$, and the intervention operators are defined by 

$\bm\cM_{\rm long}^{\pi^L}Q_1^{\pi^L}(\bmy_{\tau_k},\eta^P_{\tau_k}):=R(\bmy_{\tau_k},\eta^P_{\tau_k},0)-\alpha+\gamma\sum_{\bmy'\in\bm{\mathcal{Y}}}P(\bmy';\eta^P_{\tau_k},0,\bmy)Q_1^{\pi^L}(\bmy',\eta^P_{\tau_k})|\eta^P_{\tau_k}\sim\pi^L(\cdot|\bmy_{\tau_k})
$

and

$\bm\cM_{\rm fast}^{\pi^F}Q_2^{\pi^F}(\bmy_{\rho_k},\eta^I_{\rho_k}):=R(\bmy_{\rho_k},0,\eta^I_{\rho_k})-\beta+\gamma\sum_{\bmy'\in\bm{\mathcal{Y}}}P(\bmy';0,\eta^I_{\rho_k},\bmy)Q_1^{\pi^F}(\bmy',\eta^I_{\rho_k})|\eta^I_{\rho_k}\sim\pi^F(\cdot|\bmy_{\rho_k})
$,

which measure the expected return for the switching agent following a long-acting intervention and a fast-acting intervention under their respective policies at state $y$ minus the long-acting and fast-acting intervention costs.

\end{proposition}
The next result is a key result of the paper which proves the existence of a solution and the joint convergence of the switcher and intervention policies and that the solution is a limit point of repeated application of a Bellman operator. 
\begin{theorem}\label{theorem:joint-sol}
Let $v:\bm{\cY}\to\mathbb{R}$ then the sequence of Bellman operators acting on $v
$ converges to the solution of the game, that is to say for any $\bmy\in\bm{\cY}$ the following holds: 
\begin{align}
\underset{k\to\infty}{\lim}T^kv(\bmy|\boldsymbol{\pi},\mathfrak{g})=v^\ast(\bmy),
\end{align}
where $v^\ast(\bmy)=\max\limits_{\hat{\mathfrak{g}},\hat{\boldsymbol{\pi}}\in \bm\Pi}v^{\hat{\boldsymbol{\pi}},\hat{\mathfrak{g}}}(\bmy|\bm{\hat{\pi}},\hat{\mathfrak{g}})
$ and the operator $T$ is given by 
\\$
T v(\bmy):=\max\left[\max\left\{\bm\cM_{\rm long}Q_1, R(\bmy,\bm0)+\gamma\sum_{\bmy'\in\bm{\mathcal{Y}}}P(\bmy';\bm0,\bmy)v(\bmy')\right\}
,\bm\cM_{\rm fast}Q_2\right]$,
and the intervention operators are defined by

$\bm\cM_{\rm long}Q_1(\bmy_{\tau_k},\eta):=\underset{\eta'\in \mathcal{H}^L}{\max} \left(R(\bmy_{\tau_k},\eta',0)-\alpha+\gamma\sum_{\bmy'\in\bm{\mathcal{Y}}}P(\bmy';\eta',0,\bmy_{\tau_k})v(\bmy')\right)
$

and

$\bm\cM_{\rm fast}^{\pi^F}Q_2^{\pi^F}(\bmy_{\rho_k},\eta):=\underset{\eta'\in \mathcal{H}^F}{\max} \left(R(\bmy_{\rho_k},0,\eta')-\beta+\gamma\sum_{\bmy'\in\bm{\mathcal{Y}}}P(\bmy';0,\eta',\bmy_{\rho_k})v(\bmy')\right)
$.
\end{theorem}

\section*{Proof of Proposition \ref{theorem:existence}}
Proposition \ref{theorem:existence} is a special case of Theorem \ref{theorem:joint-sol}, namely it is achieved when the intervention policies are fixed hence, we skip the proof of the proposition and prove Theorem \ref{theorem:joint-sol} directly.

\section*{Proof of Theorem \ref{theorem:joint-sol}}
\begin{proof}

Recall that the Bellman operator acting on a function $v:\bm{\cY}\to\mathbb{R}$ is:

\begin{align}
T v(\bmy):=\max\left[\max\left\{\bm\cM_{\rm long}Q_1, R(\bmy,\bm0)+\gamma\sum_{\bmy'\in\bm{\mathcal{Y}}}P(\bmy';\bm0,\bmy)v(\bmy')\right\}
,\bm\cM_{\rm fast}Q_2\right].\label{bellman_proof_start}
\end{align}

It suffices to prove that $T$ is a contraction operator. Thereafter, we use both results to prove the existence of a solution of $\bm{\cM}$ as a limit point of a sequence generated by successively applying the Bellman operator to a test value function.   
Therefore our next result shows that the following bounds holds:
\begin{lemma}\label{lemma:bellman_contraction}
The Bellman operator $T$ is a contraction so that for any real-valued maps $v,v'$, the following bound holds: $
\left\|Tv-Tv'\right\|\leq \gamma\left\|v-v'\right\|$.
\end{lemma}


We now consider the four cases produced by \eqref{bellman_proof_start}, that is to say we prove the following statements:

i) $\qquad\qquad
\left\|\bm\cM_i\bmQ-\bm\cM_j\bmQ'\right\|\leq    \gamma\left\|v-v'\right\|,\qquad i\in\{{\rm fast},{\rm long}\}.$

ii)  $\qquad\qquad
    \left\|\bm\cM_i\bmQ-\left[ R(\cdot,\boldsymbol{0})+\gamma\mathcal{P}^{\boldsymbol{0}}_{\bm{y'}\bmy}v'\right]\right\|\leq \gamma\left\|v-v'\right\|\qquad i\in\{{\rm fast},{\rm long}\}.
$

We first prove (i). We break the proof into two cases:

\textbf{Case 1:}
\begin{align}
\underset{\eta'\in \mathcal{H}^i}{\max} \left(R(\bmy_{\tau},\eta',\eta^j)-c+\gamma\cP^{(\eta',\eta^j)}_{\bm{y'}\bmy_{\tau}}v(\bm{y'})\right)-\underset{\eta''\in \mathcal{H}^j}{\max} \left(R(\bmy_{\tau},\eta^i,\eta'')-c+\gamma\cP^{(\eta^i,\eta'')}_{\bm{y'}\bmy_{\tau}}v'(\bm{y'})\right)\leq 0
\end{align}
\begin{align*}
&\left|(\bm\cM_i\bmQ-\bm\cM_j\bmQ')(\bmy_{\tau},\bm\eta)\right|
\\&=    \left|\underset{\eta'\in \mathcal{H}^i}{\max} \left(R(\bmy_{\tau},\eta',\eta^j)+\gamma\cP^{(\eta',\eta^j)}_{\bm{y'}\bmy_{\tau}}v(\bm{y'})\right)-\underset{\eta''\in \mathcal{H}^j}{\max} \left(R(\bmy_{\tau},\eta^i,\eta'')+\gamma\cP^{(\eta^i,\eta'')}_{\bm{y'}\bmy_{\tau}}v'(\bm{y'})\right)\right|
\\&\hspace{-1cm}\begin{aligned}\leq  \Bigg|\max\left\{\underset{\eta'\in \mathcal{H}^i}{\max} \left(R(\bmy_{\tau},\eta',\eta^j)+\gamma\cP^{(\eta',\eta^j)}_{\bm{y'}\bmy_{\tau}}v(\bm{y'})\right),\underset{\eta''\in \mathcal{H}^j}{\max} \left(R(\bmy_{\tau},\eta^i,\eta'')+\gamma\cP^{(\eta^i,\eta'')}_{\bm{y'}\bmy_{\tau}}v(\bm{y'})\right)\right\}&
\\-\underset{\eta''\in \mathcal{H}^j}{\max}
\left(R(\bmy_{\tau},\eta^i,\eta'')+\gamma\cP^{(\eta^i,\eta'')}_{\bm{y'}\bmy_{\tau}}v'(\bm{y'})\right)\Bigg|&
\end{aligned}
\\&\hspace{-1cm}\leq    \Bigg|\max\left\{\underset{\eta'\in \mathcal{H}^i}{\max} \left(R(\bmy_{\tau},\eta',\eta^j)+\gamma\cP^{(\eta',\eta^j)}_{\bm{y'}\bmy_{\tau}}v(\bm{y'})\right),\underset{\eta''\in \mathcal{H}^j}{\max} \left(R(\bmy_{\tau},\eta^i,\eta'')+\gamma\cP^{(\eta^i,\eta'')}_{\bm{y'}\bmy_{\tau}}v(\bm{y'})\right)\right\}
\\&\hspace{-1cm}- \max\left\{\underset{\eta'\in \mathcal{H}^i}{\max} \left(R(\bmy_{\tau},\eta',\eta^j)+\gamma\cP^{(\eta',\eta^j)}_{\bm{y'}\bmy_{\tau}}v(\bm{y'})\right),\underset{\eta''\in \mathcal{H}^j}{\max} \left(R(\bmy_{\tau},\eta^i,\eta'')+\gamma\cP^{(\eta^i,\eta'')}_{\bm{y'}\bmy_{\tau}}v'(\bm{y'})\right)\right\}
\\&\hspace{-1cm}\begin{aligned}+
\max\left\{\underset{\eta'\in \mathcal{H}^i}{\max} \left(R(\bmy_{\tau},\eta',\eta^j)+\gamma\cP^{(\eta',\eta^j)}_{\bm{y'}\bmy_{\tau}}v(\bm{y'})\right),\underset{\eta''\in \mathcal{H}^j}{\max} \left(R(\bmy_{\tau},\eta^i,\eta'')+\gamma\cP^{(\eta^i,\eta'')}_{\bm{y'}\bmy_{\tau}}v'(\bm{y'})\right)\right\}&
\\-\underset{\eta''\in \mathcal{H}^j}{\max} \left(R(\bmy_{\tau},\eta^i,\eta'')+\gamma\cP^{(\eta^i,\eta'')}_{\bm{y'}\bmy_{\tau}}v'(\bm{y'})\right)\Bigg|&
\end{aligned}
\\&\hspace{-1cm}\leq\left|    \underset{\eta''\in \mathcal{H}^j}{\max} \left(R(\bmy_{\tau},\eta^i,\eta'')+\gamma\cP^{(\eta^i,\eta'')}_{\bm{y'}\bmy_{\tau}}v(\bm{y'})\right)-\underset{\eta''\in \mathcal{H}^j}{\max} \left(R(\bmy_{\tau},\eta^i,\eta'')+\gamma\cP^{(\eta^i,\eta'')}_{\bm{y'}\bmy_{\tau}}v'(\bm{y'})\right)\right|
\\&\hspace{-1.5cm}+\Bigg|
\max\left\{\underset{\eta'\in \mathcal{H}^i}{\max} \left(R(\bmy_{\tau},\eta',\eta^j)+\gamma\cP^{(\eta',\eta^j)}_{\bm{y'}\bmy_{\tau}}v(\bm{y'})\right)-\underset{\eta''\in \mathcal{H}^j}{\max} \left(R(\bmy_{\tau},\eta^i,\eta'')+\gamma\cP^{(\eta^i,\eta'')}_{\bm{y'}\bmy_{\tau}}v'(\bm{y'})\right),0\right\}\Bigg|
\\&   
\leq\gamma\max\limits_{\eta'\in \mathcal{H}^i}\max\limits_{\eta''\in \mathcal{H}^j}\left|\cP^{(\eta',\eta'')}_{\bm{y'}\bmy_{\tau}}v(\bm{y'})-\cP^{(\eta',\eta'')}_{\bm{y'}\bmy_{\tau}}v'(\bm{y'})\right|
\\&\leq \gamma\left\|v-v'\right\|,
\end{align*}
where we have again used the fact that for any scalars $a,b,c$ we have that $
    \left|\max\{a,b\}-\max\{b,c\}\right|\leq \left|a-c\right|$ using the non-expansiveness of $\cP$.
    
\textbf{Case 2:}
\begin{align}\nonumber
&\max\limits_{\eta'\in \mathcal{H}^i} \left(R(\bmy_{\tau},\eta',\eta^j)-c+\gamma\cP^{(\eta',\eta^j)}_{\bm{y'}\bmy_{\tau}}v(\bm{y'})\right)
\\&\qquad\qquad\qquad\quad-\underset{\eta''\in \mathcal{H}^j}{\max} \left(R(\bmy_{\tau},\eta^i,\eta'')-c+\gamma\cP^{(\eta^i,\eta'')}_{\bm{y'}\bmy_{\tau}}v'(\bm{y'})\right)> 0, \quad \forall \bm\eta=(\eta^i,\eta^j)\in\mathcal{H}^i\times\mathcal{H}^j.
\end{align}
Now we observe that
\begin{align*}
&(\bm\cM_i \bmQ-\bm\cM_j \bmQ')(\bmy_{\tau},\bm{\eta}_{\tau})
\\&=    \underset{\eta'\in \mathcal{H}^i}{\max} \left(R(\bmy_{\tau},\eta',\eta^j)-c+\gamma\cP^{\eta',\eta^j}_{\bm{y'}\bmy_{\tau}}v(\bm{y'})\right)-\underset{\eta''\in \mathcal{H}^j}{\max} \left(R(\bmy_{\tau},\eta^i,\eta'')-c+\gamma\cP^{(\eta^i,\eta'')}_{\bm{y'}\bmy_{\tau}}v'(\bm{y'})\right)
\\&\begin{aligned}\leq \underset{\eta'\in \mathcal{H}^i}{\max}\underset{\eta''\in \mathcal{H}^j}{\max} &\left(R(\bmy_{\tau},\eta',\eta'')+\gamma\cP^{(\eta',\eta'')}_{\bm{y'}\bmy_{\tau}}v(\bmy)\right)-\underset{\eta'\in \mathcal{H}^i}{\min}\underset{\eta''\in \mathcal{H}^j}{\max} \left(R(\bmy_{\tau},\eta',\eta^j)+\gamma\cP^{(\eta',\eta'')}_{\bm{y'}\bmy_{\tau}}v'(\bm{y'})\right) 
\end{aligned}
\\&\begin{aligned}\leq \Bigg|\underset{\eta'\in \mathcal{H}^i}{\max}\underset{\eta''\in \mathcal{H}^j}{\max} &\left(R(\bmy_{\tau},\eta',\eta'')+\gamma\cP^{(\eta',\eta'')}_{\bm{y'}\bmy_{\tau}}v(\bmy)\right)-\underset{\eta'\in \mathcal{H}^i}{\min}\underset{\eta''\in \mathcal{H}^j}{\max} \left(R(\bmy_{\tau},\eta',\eta^j)+\gamma\cP^{(\eta',\eta'')}_{\bm{y'}\bmy_{\tau}}v'(\bm{y'})\right)\Bigg| 
\end{aligned}
\\&\begin{aligned}
\leq \underset{\eta'\in \mathcal{H}^i}{\max}\underset{\eta''\in \mathcal{H}^j}{\max}\Bigg|& \left(R(\bmy_{\tau},\eta',\eta'')+\gamma\cP^{(\eta',\eta'')}_{\bm{y'}\bmy_{\tau}}v(\bm{y'})\right)-\left(R(\bmy_{\tau},\eta',\eta'')+\gamma\cP_{\bm{y'}\bmy_{\tau}}^{(\eta',\eta'')}v'(\bm{y'})\right) \Bigg|
\end{aligned}
\\&\leq \gamma\underset{\eta'\in \mathcal{H}^i}{\max}\underset{\eta''\in \mathcal{H}^j}{\max}\left| \cP_{\bm{y'}\bmy_{\tau}}^{(\eta',\eta'')}(v-v')(\bm{y'})\right|  
\\&\leq \gamma\|\cP\|\left\|v-v'\right\|
\\&\leq \gamma\left\|v-v'\right\|,
\end{align*}
using Lemma \ref{lemma:max_min}, the non-expansiveness of $\cP$ and, in the penultimate step the Cauchy-Schwarz inequality. Since in this case $(\bm\cM_i \bmQ-\bm\cM_j \bmQ')(\bmy_{\tau},\bm{\eta}_{\tau})=|(\bm\cM_i \bmQ-\bm\cM_j \bmQ')(\bmy_{\tau},\bm{\eta}_{\tau})|$ this completes the proof for $i\neq j$. 
The proof for $i=j$ is completely analogous, we therefore omit the proof.

We now prove ii). We split the proof of the statement into two cases:

\textbf{Case 1:} 
\begin{align}\bm\cM_i\bm{Q}(\bmy_{\tau},\bm{\eta}_{\tau})-\left(R(\bmy_{\tau},\bm0)+\gamma\mathcal{P}^{\boldsymbol{0}}_{\bm{y'}\bmy_{\tau}}v'(\bm{y'})\right)<0, \qquad i\in\{{\rm fast},{\rm long}\}.
\end{align}

We now observe the following:
\begin{align*}
&\bm\cM_i\bm{Q}(\bmy_{\tau},\bm{\eta}_{\tau})-\left(R(\bmy_{\tau},\boldsymbol{0})+\gamma\mathcal{P}^{\boldsymbol{0}}_{\bm{y'}\bmy_{\tau}}v'(\bm{y'})\right)
\\&\leq\max\left\{R(\bmy_{\tau},\boldsymbol{0})+\gamma\cP_{\bm{y'}\bmy_{\tau}}^{\bm0}v(\bm{y'}),\bm\cM_i\bm{Q}(\bmy_{\tau},\bm{\eta}_{\tau})\right\}
-\left(R(\bmy_{\tau},\boldsymbol{0})+\gamma\mathcal{P}^{\boldsymbol{0}}_{\bm{y'}\bmy_{\tau}}v'(\bm{y'})\right)
\\&\leq \Bigg|\max\left\{R(\bmy_{\tau},\boldsymbol{0})+\gamma\cP_{\bm{y'}\bmy_{\tau}}^{\bm0}v(\bm{y'}),\bm\cM_i\bm{Q}(\bmy_{\tau},\bm{\eta}_{\tau})\right\}
-\max\left\{R(\bmy_{\tau},\boldsymbol{0})+\gamma\mathcal{P}^{\boldsymbol{0}}_{\bm{y'}\bmy_{\tau}}v'(\bm{y'}),\bm\cM_i\bm{Q}(\bmy_{\tau},\bm{\eta}_{\tau})\right\}
\\&\qquad+\max\left\{R(\bmy_{\tau},\boldsymbol{0})+\gamma\mathcal{P}^{\boldsymbol{0}}_{\bm{y'}\bmy_{\tau}}v'(\bm{y'}),\bm\cM_i\bm{Q}(\bmy_{\tau},\bm{\eta}_{\tau})\right\}
-\left(R(\bmy_{\tau},\boldsymbol{0})+\gamma\mathcal{P}^{\boldsymbol{0}}_{\bm{y'}\bmy_{\tau}}v'(\bm{y'})\right)\Bigg|
\\&\leq \Bigg|\max\left\{R(\bmy_{\tau},\boldsymbol{0})+\gamma\mathcal{P}^{\boldsymbol{0}}_{\bm{y'}\bmy_{\tau}}v(\bm{y'}),\bm\cM_i\bm{Q}(\bmy_{\tau},\bm{\eta}_{\tau})\right\}
-\max\left\{R(\bmy_{\tau},\boldsymbol{0})+\gamma\mathcal{P}^{\boldsymbol{0}}_{\bm{y'}\bmy_{\tau}}v'(\bm{y'}),\bm\cM_i\bm{Q}(\bmy_{\tau},\bm{\eta}_{\tau})\right\}\Bigg|
\\&\qquad+\Bigg|\max\left\{R(\bmy_{\tau},\boldsymbol{0})+\gamma\mathcal{P}^{\boldsymbol{0}}_{\bm{y'}\bmy_{\tau}}v'(\bm{y'}),\bm\cM_i\bm{Q}(\bmy_{\tau},\bm{\eta}_{\tau})\right\}-\left(R(\bmy_{\tau},\boldsymbol{0})+\gamma\mathcal{P}^{\boldsymbol{0}}_{\bm{y'}\bmy_{\tau}}v'(\bm{y'})\right)\Bigg|
\\&\leq \gamma\max\limits_{\bm\eta\in\bm{\mathcal{H}}}\left|\cP_{\bm{y'}\bmy_{\tau}}^{\bm\eta}v(\bm{y'})-\cP_{\bm{y'}\bmy_{\tau}}^{\bm\eta}v'(\bm{y'})\right|+\left|\max\left\{0,\bm\cM_i\bm{Q}(\bmy_{\tau},\bm{\eta}_{\tau})-\left(R(\bmy_{\tau},\boldsymbol{0})+\gamma\mathcal{P}^{\boldsymbol{0}}_{\bm{y'}\bmy_{\tau}}v'(\bm{y'})\right)\right\}\right|
\\&\leq \max\limits_{\bm\eta\in\bm{\mathcal{H}}}\left\|\cP_{\bm{y'}\bmy_{\tau}}^{\bm\eta}\right\|\left\|v-v'\right\|
\\&\leq \gamma\|v-v'\|,
\end{align*}
where we have again used the fact that for any scalars $a,b,c$ we have that $
    \left|\max\{a,b\}-\max\{b,c\}\right|\leq \left|a-c\right|$ and the non-expansiveness of the $\cP$ operator.

\textbf{Case 2: }
\begin{align}\bm\cM_i\bm{Q}(\bmy_{\tau},\bm{\eta}_{\tau})-\left(R(\bmy_{\tau},\boldsymbol{0})+\gamma\mathcal{P}^{\boldsymbol{0}}_{\bm{y'}\bmy_{\tau}}v'(\bm{y'})\right)\geq 0, \qquad i\in\{{\rm fast},{\rm long}\}.\label{ii.case_2.assumption}
\end{align}

For this case,  we observe that
\begin{align*}
&\bm\cM_i\bm{Q}(\bmy_{\tau},\bm\eta_{\tau})-\left(R(\bmy_{\tau},\boldsymbol{0})+\gamma\mathcal{P}^{\boldsymbol{0}}_{\bm{y'}\bmy_{\tau}}v'(\bm{y'})\right)
\\&\begin{aligned}
=\bm\cM_i\bm{Q}(\bmy_{\tau},\bm\eta_{\tau})-\max\{\bm\cM_i\bm{Q}(\bmy_{\tau},\bm\eta_{\tau}),\bm\cM_i\bm{Q'}(\bmy_{\tau},\bm\eta_{\tau})\}+\max\{\bm\cM_i\bm{Q}(\bmy_{\tau},\bm\eta_{\tau}),\bm\cM_i\bm{Q'}(\bmy_{\tau},\bm\eta_{\tau})\}&\\-\left( R(\bmy_{\tau},\boldsymbol{0})+\gamma\mathcal{P}^{\boldsymbol{0}}_{\bm{y'}\bmy_{\tau}}v'(\bm{y'})\right).&
\end{aligned}
\end{align*}
Now, using the fact that $\max\{\bm\cM_i\bm{Q},\bm\cM_i\bm{Q'}\}-\bm\cM_i\bm{Q}\geq -\|\bm\cM_i\bm{Q}-\bm\cM_i\bm{Q'}\|$, implies 
\begin{align*}
&R(\bmy_{\tau},\boldsymbol{0})+\gamma\mathcal{P}^{\boldsymbol{0}}_{\bm{y'}\bmy_{\tau}}v'(\bm{y'}) -\|\bm\cM_i\bm{Q}-\bm\cM_i\bm{Q'}\|
\\&\leq -\bm\cM_i\bm{Q}(\bmy_{\tau},\bm\eta_{\tau})+\max\{\bm\cM_i\bm{Q}(\bmy_{\tau},\bm\eta_{\tau}),\bm\cM_i\bm{Q'}(\bmy_{\tau},\bm\eta_{\tau})\}+R(\bmy_{\tau},\boldsymbol{0})+\gamma\mathcal{P}^{\boldsymbol{0}}_{\bm{y'}\bmy_{\tau}}v'(\bm{y'})
\\&\leq\max\{\bm\cM_i\bm{Q}(\bmy_{\tau},\bm\eta_{\tau}),\bm\cM_i\bm{Q'}(\bmy_{\tau},\bm\eta_{\tau})\},
\end{align*}
using \eqref{ii.case_2.assumption}. From this we find that
\begin{align}
\|\bm\cM_i\bm{Q}-\bm\cM_i\bm{Q'}\|\geq R(\bmy_{\tau},\boldsymbol{0})+\gamma\mathcal{P}^{\boldsymbol{0}}_{\bm{y'}\bmy_{\tau}}v'(\bm{y'})-\max\{\bm\cM_i\bm{Q}(\bmy_{\tau},\bm\eta_{\tau}),\bm\cM_i\bm{Q'}(\bmy_{\tau},\bm\eta_{\tau})\}.    \label{case_2_proof_intermediate}
\end{align}
Setting $y=\bm\cM_i\bm{Q}$, $x=R(\bmy_{\tau},\boldsymbol{0})+\gamma\mathcal{P}^{\boldsymbol{0}}_{\bm{y'}\bmy_{\tau}}v'(\bm{y'})$ and $z=\bm\cM_i\bm{Q'}$ in Lemma \ref{lemma:basic_inequality} and after combining this fact with \eqref{case_2_proof_intermediate} we arrive at 
\begin{align}
\bm\cM_i\bm{Q}-\left(R(\bmy_{\tau},\boldsymbol{0})+\gamma\mathcal{P}^{\boldsymbol{0}}_{\bm{y'}\bmy_{\tau}}v'(\bm{y'})\right)\leq \|\bm\cM_i\bm{Q}-\bm\cM_i\bm{Q'}\|\leq \gamma\|v-v'\|,
\end{align}
by part (i) which is the required result.

Hence we have succeeded in showing that for any $v\in L_2$ we have that
\begin{align}
    \left\|\bm\cM\bmQ- \left(R(\cdot,\bm0)+\gamma\mathcal{P}^{\bm0}v'\right)\right\|\leq \gamma\left\|v-v'\right\|.\label{off_M_bound_gen}
\end{align}
Gathering the results of the two cases completes the proof of Theorem \ref{theorem:joint-sol}. 
\end{proof}

To prove the Theorem \ref{theorem:q_learning}, we make use of the following result:
\begin{theorem}[Theorem 1, pg 4 in \citep{jaakkola1994convergence}]
Let $\Xi_t(\bmy)$ be a random process that takes values in $\mathbb{R}^n$ and given by the following:
\begin{align}
    \Xi_{t+1}(\bmy)=\left(1-\alpha_t(\bmy)\right)\Xi_{t}(\bmy)\alpha_t(\bmy)L_t(\bmy),
\end{align}
then $\Xi_t(\bmy)$ converges to $0$ with probability $1$ under the following conditions:
\begin{itemize}
\item[i)] $0\leq \alpha_t\leq 1, \sum_t\alpha_t=\infty$ and $\sum_t\alpha_t<\infty$
\item[ii)] $\|\mathbb{E}[L_t|\mathcal{F}_t]\|\leq \gamma \|\Xi_t\|$, with $\gamma <1$
\item[iii)] ${\rm Var}\left[L_t|\mathcal{F}_t\right]\leq c(1+\|\Xi_t\|^2)$ for some $c>0$.
\end{itemize}
\end{theorem}
\begin{proof}
To prove the result, we show (i) - (iii) hold. Condition (i) holds by choice of learning rate. It therefore remains to prove (ii) - (iii). We first prove (ii). For this, we consider our variant of the Q-learning update rule:
\begin{align*}
\bmQ_{S,t+1}(\bmy_t,\boldsymbol{\eta}_t,g|\cdot)=\bmQ_t&(\bmy_t,\boldsymbol{\eta}_t,g|\cdot)
\\&\begin{aligned}+\alpha_t(\bmy_t,\boldsymbol{\eta}_t)\Big[\max\left\{\bm{\mathcal{M}}_iQ(\bmy_{\tau_k},\boldsymbol{\eta},g|\cdot), R(\bmy_{\tau_k},\boldsymbol{\eta},g)+\gamma\underset{\bm{\eta'}\in\bm{\mathcal{H}}}{\max}\;Q(\bmy_{t+1},\bm{\eta'},g|\cdot)\right\}&
\\-\bmQ_{t}(\bmy_t,\boldsymbol{\eta}_t,g|\cdot)\Big]&.
\end{aligned}
\end{align*}
After subtracting $\bmQ^\ast(\bmy_t,\boldsymbol{\eta}_t,g|\cdot)$ from both sides and some manipulation we obtain that:
\begin{align*}
&\Xi_{t+1}(\bmy_t,\boldsymbol{\eta}_t)
\\&=(1-\alpha_t(\bmy_t,\boldsymbol{\eta}_t))\Xi_{t}(\bmy_t,\boldsymbol{\eta}_t)
\\&\quad+\alpha_t(\bmy_t,\boldsymbol{\eta}_t))\left[\max\left\{\boldsymbol{\hat{\cM}}Q(\bmy_{\tau_k},\boldsymbol{\eta},g|\cdot), R(\bmy_{\tau_k},\boldsymbol{\eta},g)+\gamma\underset{\bm{\eta'}\in\bm{\mathcal{H}}}{\max}\;\bmQ(\bm{y'},\bm{\eta'},g|\cdot)\right\}-\bmQ^\ast(\bmy_t,\boldsymbol{\eta}_t,g|\cdot)\right],  \end{align*}
where $\Xi_{t}(\bmy_t,\boldsymbol{\eta}_t,g):=\bmQ_t(\bmy_t,\boldsymbol{\eta}_t,g|\cdot)-Q^\star(\bmy_t,\boldsymbol{\eta}_t,g|\cdot)$.

Let us now define by 
\begin{align*}
L_t(\bmy_{\tau_k},\boldsymbol{\eta},g):=\max\left\{\bm{\mathcal{M}}_iQ(\bmy_{\tau_k},\boldsymbol{\eta},g|\cdot), R(\bmy_{\tau_k},\boldsymbol{\eta},g)+\gamma\underset{\bm{\eta'}\in\bm{\mathcal{H}}}{\max}\;\bmQ(\bm{y'},\bm{\eta'},g|\cdot)\right\}-\bmQ^\ast(\bmy_t,\bm{\eta},g|\cdot).
\end{align*}
Then
\begin{align}
\Xi_{t+1}(\bmy_t,\boldsymbol{\eta}_t,g)=(1-\alpha_t(\bmy_t,\boldsymbol{\eta}_t))\Xi_{t}(\bmy_t,\boldsymbol{\eta}_t,g)+\alpha_t(\bmy_t,\boldsymbol{\eta}_t))\left[L_t(\bmy_{\tau_k},\boldsymbol{\eta},g)\right].   
\end{align}

We now observe that
\begin{align}\nonumber
&\mathbb{E}\left[L_t(\bmy_{\tau_k},\boldsymbol{\eta},g)|\mathcal{F}_t\right]
\\&\begin{aligned}=\sum_{\bm{y'}\in\mathcal{S}}P(\bm{y'};a,\bmy_{\tau_k})\max\left\{\bm{\mathcal{M}}_iQ(\bmy_{\tau_k},\boldsymbol{\eta},g|\cdot), R(\bmy_{\tau_k},\boldsymbol{\eta},g)+\gamma\underset{\bm{\eta'}\in\bm{\mathcal{H}}}{\max}\;\bmQ(\bm{y'},\bm{\eta'},g|\cdot)\right\}\nonumber&
\\-\bmQ^\ast(\bmy_{\tau_k},a,g|\cdot)&
\end{aligned}
\\&= T \bmQ_t(\bmy,\bm{\eta},g|\cdot)-\bmQ^\ast(\bmy,\bm{\eta},g). \label{expectation_L}
\end{align}
Now, using the fixed point property that implies $\bmQ^\ast=T \bmQ^\ast$, we find that
\begin{align}\nonumber
    \mathbb{E}\left[L_t(\bmy_{\tau_k},\boldsymbol{\eta},g)|\mathcal{F}_t\right]&=T \bmQ_t(\bmy,\bm{\eta},g|\cdot)-T \bmQ^\ast(\bmy,\bm{\eta},g|\cdot)
    \\&\leq\left\|T \bmQ_t-T \bmQ^\ast\right\|\nonumber
    \\&\leq \gamma\left\| \bmQ_t- \bmQ^\ast\right\|_\infty=\gamma\left\|\Xi_t\right\|_\infty.
\end{align}
using the contraction property of $T$ established in Lemma \ref{lemma:bellman_contraction}. This proves (ii).

We now prove iii), that is
\begin{align}
    {\rm Var}\left[L_t|\mathcal{F}_t\right]\leq c(1+\|\Xi_t\|^2).
\end{align}
Now by \eqref{expectation_L} we have that
\begin{align*}
  {\rm Var}\left[L_t|\mathcal{F}_t\right]&= {\rm Var}\left[\max\left\{\bm{\mathcal{M}}_iQ(\bmy_{\tau_k},\boldsymbol{\eta},g|\cdot), R(\bmy_{\tau_k},\boldsymbol{\eta},g)+\gamma\underset{\bm{\eta'}\in\bm{\mathcal{H}}}{\max}\;\bmQ(\bm{y'},\bm{\eta'},g|\cdot)\right\}-\bmQ^\ast(\bmy_t,\bm{\eta},g|\cdot)\right]
  \\&= \mathbb{E}\Bigg[\Bigg(\max\left\{\bm{\mathcal{M}}_i\bmQ(\bmy_{\tau_k},\boldsymbol{\eta},g|\cdot), R(\bmy_{\tau_k},\boldsymbol{\eta},g)+\gamma\underset{\bm{\eta'}\in\bm{\mathcal{H}}}{\max}\;\bmQ(\bm{y'},\bm{\eta'},g|\cdot)\right\}
  \\&\qquad\qquad\qquad\qquad\qquad\quad\quad\quad-\bmQ^\ast(\bmy_t,\bm{\eta},g|\cdot)-\left(T \bmQ_t(\bmy,\bm{\eta},g|\cdot)-\bmQ^\ast(\bmy,\bm{\eta},g|\cdot)\right)\Bigg)^2\Bigg]
      \\&= \mathbb{E}\left[\left(\max\left\{\bm{\mathcal{M}}_iQ(\bmy_{\tau_k},\boldsymbol{\eta},g|\cdot), R(\bmy_{\tau_k},\boldsymbol{\eta},g)+\gamma\underset{\bm{\eta'}\in\bm{\mathcal{H}}}{\max}\;Q(\bm{y'},\bm{\eta'},g|\cdot)\right\}-T \bmQ_t(\bmy,\bm{\eta},g|\cdot)\right)^2\right]
    \\&= {\rm Var}\left[\max\left\{\bm{\mathcal{M}}_iQ(\bmy_{\tau_k},\boldsymbol{\eta},g|\cdot), R(\bmy_{\tau_k},\boldsymbol{\eta},g)+\gamma\underset{\bm{\eta'}\in\bm{\mathcal{H}}}{\max}\;Q(\bm{y'},\bm{\eta'},g|\cdot)\right\}-T \bmQ_t(\bmy,\boldsymbol{\eta},g|\cdot))\right]
    \\&\leq c(1+\|\Xi_t\|^2),
\end{align*}
for some $c>0$ where the last line follows due to the boundedness of $Q$ (which follows from Assumptions 2 and 4). This concludes the proof of the Theorem.
\end{proof}

\section*{Proof of Proposition \ref{prop:switching_times}}
\begin{proof}
We begin by re-expressing the \textit{activation times} at which the Switching agent agent activates an intervention policy. In particular,an activation time $\tau_k$ is defined recursively $\tau_k=\inf\{t>\tau_{k-1}|\bmy_t\in A,\tau_k\in\mathcal{F}_t\}$ where $A=\{y\in\bm{\cY},g(\bmy_t)=1\}$.
The proof is given by deriving a contradiction.  Therefore suppose that $\bm{\mathcal{M}}_iv(\bmy_{\tau_k})> v(\bmy_{\tau_k})$ for $i\in\{{\rm fast},{\rm slow}\}$ and suppose that the activation time $\tau'_1>\tau_1$ is an optimal activation time. Construct the Switching agent $g'$ and $\tilde{g}$ policy activation times by $(\tau'_0,\tau'_1,\ldots,)$ and $g'^2$ policy by $(\tau'_0,\tau_1,\ldots)$ respectively.  Define by $l=\inf\{t>0;\bm{\mathcal{M}}_i\psi(\bmy_{t}= \psi(\bmy_{t}\}$ and $m=\sup\{t;t<\tau'_1\}$.
By construction we have that
\begin{align*}
& \quad v(\bmy)
\\&=\mathbb{E}\left[R(\bmy_{0},\bm{\eta}_{0})+\mathbb{E}\left[\ldots+\gamma^{l-1}\mathbb{E}\left[R(\bmy_{\tau_1-1},\bm{\eta}_{\tau_1-1})+\ldots+\gamma^{m-l-1}\mathbb{E}\left[ R(\bmy_{\tau'_1-1},\bm{\eta}_{\tau'_1-1})+\gamma\bm{\mathcal{M}}_iv(\bmy')\right]\right]\right]\right]
\\&<\mathbb{E}\left[R(\bmy_{0},\bm{\eta}_{0})+\mathbb{E}\left[\ldots+\gamma^{l-1}\mathbb{E}\left[ R(\bmy_{\tau_1-1},\bm{\eta}_{\tau_1-1})+\gamma\bm{\mathcal{M}}_iv(\bmy_{\tau_1})\right]\right]\right]
\end{align*}
We now use the following observation $\mathbb{E}\left[ R(\bmy_{\tau_1-1},\bm{\eta}_{\tau_1-1})+\gamma\bm{\mathcal{M}}_iv(\bmy_{\tau_1})\right]\\\ \text{\hspace{30 mm}}\geq \min\left\{\bm{\mathcal{M}}_iv(\bmy_{\tau_1}),\underset{\bm{\eta}_{\tau_1}\in\mathcal{A}}{\max}\;\left[ R(\bmy_{\tau_{1}},\bm{\eta}_{\tau_{1}})+\gamma\sum_{\bm{y'}\in\mathcal{S}}P(\bm{y'};\bm{\eta}_{\tau_1},\bmy_{\tau_1})v^{\boldsymbol{\pi},g}(\bmy')\right]\right\}$.

Using this we deduce that
\begin{align*}
&v(s>\mathbb{E}\Bigg[R(\bmy_{0},\bm{\eta}_{0})+\mathbb{E}\Bigg[\ldots
\\&+\gamma^{l-1}\mathbb{E}\left[ R(\bmy_{\tau_1-1},\bm{\eta}_{\tau_1-1})+\gamma\max\left\{\bm{\mathcal{M}}^v(\bmy_{\tau_1}),\underset{a_{\tau_1}\in\mathcal{A}}{\max}\;\left[ R(\bmy_{\tau_{k}},\bm{\eta}_{\tau_{k}})+\gamma\sum_{\bm{y'}\in\mathcal{S}}P(\bm{y'};\bm{\eta}_{\tau_1},\bmy_{\tau_1})v(\bmy')\right]\right\}\right]\Bigg]\Bigg]
\\&=\mathbb{E}\left[R(\bmy_{0},\bm{\eta}_{0})+\mathbb{E}\left[\ldots+\gamma^{l-1}\mathbb{E}\left[ R(\bmy_{\tau_1-1},\bm{\eta}_{\tau_1-1})+\gamma\left[T v^{\boldsymbol{\pi},\tilde{g}}\right](\bmy_{\tau_1})\right]\right]\right]=v(\bmy)
\end{align*}
where the first inequality is true by assumption on $\bm{\mathcal{M}}_i$. This is a contradiction since $g'$ is an optimal policy for the Switching agent. Using analogous reasoning, we deduce the same result for $\tau'_k<\tau_k$ after which deduce the result. Moreover, by invoking the same reasoning, we can conclude that it must be the case that $(\tau_0,\tau_1,\ldots,\tau_{k-1},\tau_k,\tau_{k+1},\ldots,)$ are the optimal activation times. 

\end{proof}

\section*{Proof of the Optimal Intervention-Policy Characterisation}
\begin{proof}
The proof of the Theorem is straightforward since by Theorem \ref{theorem:existence}, Switching agent's problem can be solved using a dynamic programming principle. The proof immediately by application of Theorem 2 in \citep{sootla2022saute}.

\end{proof}

\clearpage

\end{document}